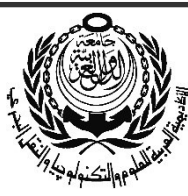

ARAB ACADEMY FOR SCIENCE, TECHNOLOGY
AND MARITIME TRANSPORT

College of Engineering and Technology

Department of Mechanical Engineering

Smart Control Systems for Energy Management

# ARTIFICIAL INTELLIGENCE APPROACHES FOR ENERGY-EFFICIENT LASER-CUTTING MACHINES

By

MOHAMED ABDALLAH MOHAMED HAMED SALEM

Egypt

A thesis submitted to AASTMT in partial

Fulfillment of the requirements for the award of the degree of

MASTER OF SCIENCE

in

SMART CONTROL SYSTEMS FOR ENERGY MANAGEMENT

## Supervisors

| Prof. Hamdy A. Ashour | Asst. Prof. Ahmed Elshenawy |
|---|---|
| Electrical & Control Engineering | Electrical & Control Engineering |
| Faculty of Engineering | Faculty of Engineering |
| AASTMT | AASTMT |
| Alexandria | Alexandria |

2023

# Declaration

I certify that all the material in this thesis that is not my own work has been identified, and that no material is included for which a degree has previously been conferred on me.

The contents of this thesis reflect my own personal views, and are not necessarily endorsed by the University.

Name:

Signed:

Date:

# إقرار الباحث

أقر بأن المادة العلمية الواردة في هذه الرسالة قد تم تحديد مصدرها العلمي وأن محتوى الرسالة غير مقدم للحصول علي أي درجة علمية أخرى، وأن مضمون هذه الرسالة يعكس أراء الباحث الخاصة وهى ليست بالضرورة الآراء التي تتبناها الجهة المانحة.

الاسم ......................

التوقيع ......................

التاريخ ......................

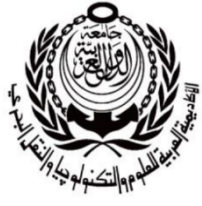

# إقرار لجنة التحكيم والمناقشة
## محمد عبدالله محمد حامد سالم

تم مناقشة هذه الرسالة وإجازتها بتاريخ: 12 /10 / 2023

### أعضاء اللجنة

**الأساتذة المشرفون على الرسالة:**

| الإسم | : | أ.د. حمدى أحمد عاشور | التوقيع: | ........................ |

الوظيفة: أستاذ الهندسة الكهربية ـ كلية الهندسة والتكنولوجيا ـ الاكاديمية العربية للعوم والتكنولوجيا والنقل البحري ـ الأسكندرية

| الإسم | : | د. أحمد خميس الشناوي | التوقيع: | ........................ |

الوظيفة: أستاذ مساعد الهندسة الكهربية ـ كلية الهندسة والتكنولوجيا ـ الاكاديمية العربية للعوم والتكنولوجيا والنقل البحري ـ الأسكندرية

**الأساتذة المحكمون:**

| الإسم | : | أ.د. سهير فتحى خميس رزيقه | التوقيع: | ........................ |

الوظيفة: أستاذ الهندسة الميكانيكيه ـ كلية الهندسة والتكنولوجيا ـ جامعة الأسكندرية الأسكندرية

| الإسم | : | أ.د. أمنية عمرو جمال الدين عطالله | التوقيع: | ........................ |

الوظيفة: أستاذ هندسة الاتصالات والالكترونيات ـ كلية الهندسة والتكنولوجيا ـ الاكاديمية العربية للعوم والتكنولوجيا والنقل البحري ـ الأسكندرية

---

| الإسم | : | أ.د. حمدى أحمد عاشور | التوقيع: | ........................ |

الوظيفة: أستاذ الهندسة الكهربية ـ كلية الهندسة والتكنولوجيا ـ الاكاديمية العربية للعوم والتكنولوجيا والنقل البحري ـ الأسكندرية

أودعت هذه الرسالة بالمكتبة بتاريخ    /    / 2023

III

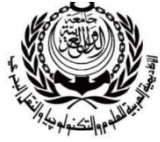

**We certify that we have read the present work and that in our opinion it is fully adequate in scope and quality as a thesis towards the partial fulfillment of the Master's Degree requirements in**

**Smart Control Systems for Energy Management**
**Mohamed Abdallah Mohamed Hamed Salem**
From
**College of Engineering and Technology**
**AASTMT**
/ / 2023

### Supervisors

Name: Prof. Hamdy Ahmed Ashour

Position: Professor of Electrical Engineering, College of Engineering and Technology, AASTMT, Alexandria

Signature: …………………………………….

Name: Dr. Ahmed Khamis El shenawy

Position: Associate Professor of Electrical Engineering, College of Engineering and Technology, AASTMT, Alexandria

Signature: …………………………………….

### Examiners

Name: Prof. Sohair Fathi Khamis Rezeka

Position: Professor of Mechanical Engineering, College of Engineering, Alexandria University, Alexandria

Signature: …………………………………….

Name: Prof. Omneya Amr Gamal El din Attallah

Position: Professor of Electronics and Communications Engineering, College of Engineering and, Technology, AASTMT, Alexandria

Signature: …………………………………….

Name: Prof. Hamdy Ahmed Ashour

Position: Professor of Electrical Engineering, College of Engineering and Technology, AASTMT, Alexandria

Signature: …………………………………….

# Acknowledgment

Thanks to Allah all mighty for this great blessing and grace. I am grateful for this great opportunity and would like to express my gratitude to my supervisors, Prof. Dr. Hamdy Ashour and Dr. Ahmed Elshenawy, for their invaluable guidance and support throughout my research. Their expertise has been instrumental in elevating the quality of my work. Additionally, I would like to thank Dr. Mohamed Aboud from Fablab-Ideaspace and the team in EME Bourg and E-just team for sharing their data and providing access to the Fablab and laser cutting machines.

Mohamed Abdallah

2023

# Abstract

Laser cutting process has a wide range of recent industrial applications, while its energy consumption and environmental impacts pose significant challenges. This study proposes deep learning-based methodologies to reduce energy usage in laser-cutting machines. Addressing the still lack of adaptive control and open-loop nature of $CO_2$ laser cutter's air suction pump, this research utilizes closed-loop configurations to enhance energy consumption. Since implementing an adaptive closed-loop system within laser cutters becomes problematic due to the diverse array of processable materials, this study introduces diverse material classification methods utilizing deep learning (DL) and speckle sensing techniques. Among the advanced classification strategies, the study investigates lens-less-based speckle sensing and a customized convolutional neural network (CNN) model. To achieve even greater energy efficiency, an additional classification approach employing a USB camera and transfer learning through the pre-trained VGG16 CNN model is proposed. Moreover, a deep learning model for smoke level detection during cutting, to simultaneously adapt the air suction pump's power according to smoke generation and material identification, is proposed and investigated. The combination of deep learning methodologies into laser cutting offers a perfect opportunity for substantial energy reduction and cost savings. This integration prompts the exhaust suction pump to automatically halt during inactive times, yielding a significant energy-efficient operation. The experimentally implemented closed-loop system dynamically adjusts pump power based on the classified material being cut and the detected ensuing smoke levels, for magnifying energy efficiency within laser-cutting machines. The experimental results emphasize a remarkable 20% to 50% reduction in energy consumption of the smoke suction pump (based on operating conditions). This research introduces an innovative and effective approach to mitigate energy consumption in laser-cutting machines based on recent artificial intelligence approaches, thus contributing significantly to the sustainable development requirements the manufacturing sector.

# Table of Contents

# List of Figures

## List of Figures (Cont'd)

# List of Tables

# List of Abbreviations

| | |
|---|---|
| ABS | Acrylonitrile Butadiene Styrene |
| ANN | Artificial Neural Network |
| AR | Augmented Reality |
| CAD | Computer-Aided Design |
| CAM | Computer-Aided Manufacturing |
| CNN | Convolutional Neural Networks |
| DL | Deep Learning |
| FCL | Fully Connected Layer |
| FN | False Negative |
| FP | False Positive |
| GAP | Global Average Pooling |
| GWP | Global Warming Potential |
| GHGE | Greenhouse Gas Emissions |
| HAZ | Heat Affected Zone |
| IOT | Internet Of Things |
| LCA | Life Cycle Assessment |
| LR | Learning Rate |
| LSTM | Long Short-Term Memory |
| MDF | Medium-Density Fibreboard |
| Mtco2 | $CO_2$ Emissions Per Million Tons - mPtn |
| PETG | Polyethylene Terephthalate Glycol |
| PL | Protective Layer |
| PVC | Polyvinyl Chloride Or Vinyl |
| Relu | Rectified Linear Unit |
| RGB | Red, Green, And Blue |
| TP | True Positive |
| TPU | Thermoplastic Polyurethane |
| TN | True Negative |
| UPLCI | Unit Process Life-Cycle Inventory |
| USB | Universal Serial Bus |
| WSN | Wireless Sensor Networks |

# List of Symbols

| Symbols | Nomenclatures | Units |
|---|---|---|
| $A'_{ij}$ | The ouput matrix from element wise multiplication between reciptive filed from the input featuer map and the used filter | - |
| $b_k$ | The bias term for feature map $k$ in layer $l$ | - |
| $E'_a$ | Total energy consumed after applying the proposed approach | KWh |
| $E'_b$ | Total energy consumed by traditional air suction pump | KWh |
| $E_s$ | Energy savings | KWh |
| $f$ | $f_h$ and $f_w$ represent the height and width of the receptive field | - |
| $f_{n'}$ | The number of feature maps | - |
| $l$ | The convolutional layer $l$ | - |
| $m$ | The hight of the input feature map | - |
| $m'$ | The hight of the output feature map | - |
| $n$ | The width of the input feature map | - |
| $n'$ | The width of the output feature map | - |
| $O_f$ | An output feature map | - |
| $P$ | Padding | - |
| $P_a$ | Adaptive control scenario's power demand | W |
| $P'_a$ | Adaptive control scenario total power demand | W |
| $P_b$ | The baseline power demand | W |
| $P'_b$ | Baseline scenario total power demand | W |
| $P_l$ | Laser power | mW |
| $P_s$ | Power savings | W |
| $s$ | Stride and $s_h$ and $s_w$ denote the vertical and horizontal strides | - |
| $w_{u,v,k',k}$ | The weight between any neuron in feature map $k$ of layer $l$ and its corresponding input situated at row $u$ and column $v$, and feature map $k'$ | - |
| $w_l$ | Laser Wavelength | nm |
| $x_{i',j',k'}$ | The output of a neuron located at layer $l-1$, row $i'$, column $j'$, and feature map $k'$ | - |
| $z_{i,j,k}$ | The output of a neuron situated at row $i$, column $j$, in feature map $k$ | - |
| $\sigma(z)_j$ | Softmax function maps the input vector $z_j$ to a probability distribution over the classes | - |

# Chapter 1: INTRODUCTION

## 1.1  Overview and Background

Laser cutting is a fundamental process that plays a crucial role in various industries, enabling precise and efficient fabrication in the era of advanced manufacturing [1], [2]. Its widespread adoption has revolutionized industrial operations, allowing for intricate designs, rapid production, and enhanced productivity [3]. However, while laser cutting offers numerous advantages, it is not without its drawbacks. This study focuses on addressing two significant challenges associated with laser cutting: energy consumption and environmental impact.

In today's global landscape, industries are increasingly prioritizing energy efficiency and sustainability as crucial considerations. The urgent focus on reducing carbon footprints and minimizing environmental impact [4] necessitates the exploration of methods to optimize energy consumption in manufacturing processes [5]. Laser cutting, despite its numerous advantages, often entails high energy requirements, resulting in elevated carbon dioxide emissions and a strain on natural resources. Consequently, it is imperative to develop strategies that mitigate energy consumption and promote environmental sustainability in laser cutting operations [6]. Consequently, relevant research by [7] enables a comparative assessment of total power demand in 2kW fiber laser cutting machine tool. as outlined in Figure 1.1.

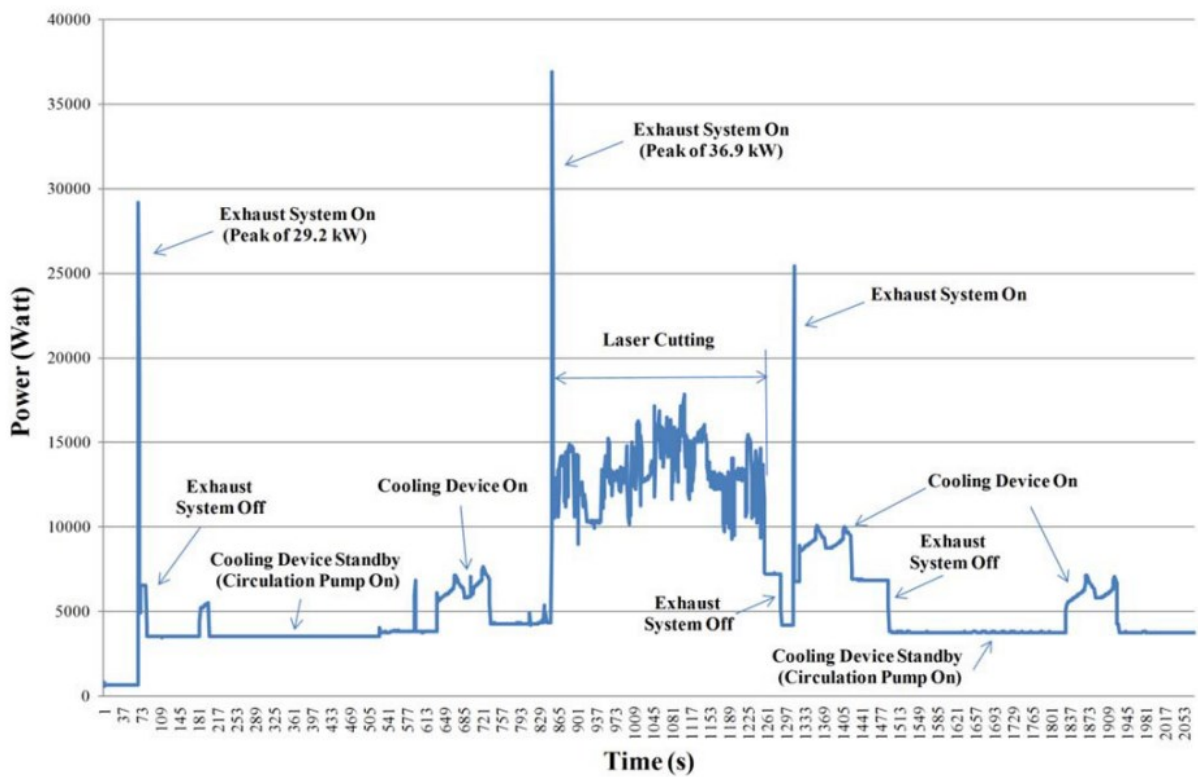

Figure 1.1 Power profile of a 2kW fiber laser cutting machine tool [7].

Moreover, the advent of Industry 4.0 has paved the way for advanced digital fabrication and manufacturing processes. This new paradigm integrates automation, data exchange, and artificial intelligence into industrial operations, leading to more efficiency and productivity [8]. In this research work, the application of deep learning techniques holds significant promise for optimizing laser cutting processes and reducing energy consumption. By leveraging the power of deep learning models and data analysis, it is possible to develop intelligent systems that adapt laser cutting operations based on real-time conditions, resulting in energy savings and enhanced sustainability [9].

The integration of deep learning and laser cutting technologies has the potential to revolutionize the industry by addressing energy consumption and environmental impact. This research aims to explore and develop innovative approaches to enhance energy usage in laser cutting machines, thereby contributing to the sustainability and efficiency of manufacturing processes.

## 1.2  Motivation

The motivation behind this research stems from the need to address the challenges associated with demand side management and environmental impact in industrial processes. Several key factors have driven the motivation for this study:

**Energy Efficiency and Cost Reduction:** With the rising importance of energy efficiency and cost reduction in today's industrial landscape, it is crucial to find ways to minimize energy consumption in manufacturing processes. Laser cutting, as a widely used fabrication technique, requires substantial energy inputs, which can significantly impact operational costs. By exploring methods to minimize energy consumption in laser cutting, businesses can achieve cost savings and enhance their competitiveness in the market.

**Environmental Sustainability:** The growing emphasis on environmental sustainability has prompted industries to assess and mitigate their environmental footprints. Laser cutting processes, due to their energy-intensive nature, contribute to carbon dioxide emissions and resource depletion. Developing strategies to reduce the environmental impact of laser cutting aligns with global efforts to combat climate change and promote sustainable manufacturing practices.

**Advancements in Deep Learning and Automation:** The advent of Industry 4.0 and the integration of advanced technologies, such as deep learning and automation, present new opportunities for optimizing manufacturing processes. Deep learning models can analyze vast amounts of data and make intelligent decisions, enabling adaptive and efficient operations. Applying these techniques to laser cutting can lead to significant energy savings and improved process control.

**Health and Safety Concerns:** Laser cutting processes can generate hazardous fumes, aerosols, and dust particles that pose risks to the health and safety of operators and workers. By reducing emissions and controlling the release of harmful substances, the well-being of personnel can be safeguarded, ensuring a healthier work environment.

The motivation behind this research is to contribute to the advancement of energy-efficient, environmentally sustainable, and safer laser cutting practices. By harnessing the potential of deep learning and automation, minimizing energy consumption, reducing emissions, enhancing process control, and improving the overall sustainability of laser cutting operations is possible. These efforts align with the broader goals of achieving sustainable development and responsible manufacturing in the face of global challenges.

## 1.3 Problem Statement

The problem addressed in this research revolves around the energy consumption and environmental impact of laser cutting processes. Despite the numerous advantages of laser cutting, there are significant challenges that need to be overcome:

**High Energy Consumption:** Laser cutting operations typically require substantial energy inputs, resulting in increased operational costs and carbon dioxide emissions. The continuous operation of energy-consuming components, such as the exhaust suction pump and air compressor, without effective control mechanisms further contributes to energy wastage.

**Environmental Impact:** Laser cutting processes can generate fumes, aerosols, and dust particles, posing environmental hazards and health risks. The carbon dioxide emissions associated with high energy consumption also contribute to climate change and resource depletion.

**Lake of Adaptive Control and Regulation:** Traditional laser cutting machines often utilize simple and not optimized control techniques that can adjust machine operations based on real-time conditions. This results to inefficient energy usage and limited responsiveness to cutting requirements and material characteristics variations.

Accordingly addressing these challenges is essential, hence by developing intelligent control strategies and leveraging deep learning techniques, it is possible to minimize energy consumption, enhance process control, and mitigate the environmental impact associated with laser cutting.

## 1.4  Proposed Solutions

This research aims to add significant contributions to the field of laser cutting and energy management in manufacturing processes. By addressing the challenges of energy consumption and environmental impact, the findings and outcomes of this study have both theoretical and practical implications.

Firstly, this work contributes to the body of knowledge by exploring the application of deep learning techniques in laser cutting machines. The development of intelligent control mechanisms based on deep learning models enhances the adaptability and efficiency of laser cutting operations. Which does not only impact energy consumption but also paves the way for advanced automation and optimization in the manufacturing industry.

Secondly, the proposed approach for detecting and classifying the material being cut during laser cutting processes offers a solution to optimize machine operations. By adjusting the machine parameters based on the material properties, energy waste can be minimized, resulting in improved energy efficiency and reduced environmental impact.

Thirdly, the integration of a smoke detection system and speckle sensing using a lens-less camera introduces innovative methods for controlling the exhaust suction pump and reducing the environmental impact of laser cutting. The detection of smoke and odor-emitting materials enables the system to regulate the power input of the exhaust suction pump, ensuring efficient extraction and minimizing the release of harmful substances into the environment.

The significance and impact of this research could be extended beyond the academic realm. The findings and proposed strategies have the potential to be implemented in other industrial settings, leading to tangible energy savings, reduced carbon emissions, and enhanced sustainability in laser cutting operations.

## 1.5 Thesis Outline

The thesis has been divided into six chapters as follows:

**Chapter One:** provides an overview of the research topic, highlighting the importance of laser cutting in diverse industries. The motivation behind the study is discussed, emphasizing the need to address energy consumption and environmental impact.

**Chapter Two:** focuses on conducting a comprehensive literature review. It explores the existing body of knowledge and research related to laser cutting, energy management, environmental impact, and deep learning techniques in addition to speckle sensing in manufacturing processes. The chapter synthesizes and analyzes relevant studies, identifying gaps and establishing the theoretical foundation for the research.

**Chapter Three:** delves into the application of deep learning techniques for material classification in laser cutting using speckle sensing. Various algorithms and models used for material classification are examined, and their strengths and limitations are discussed. The chapter presents the proposed methodology and experimental setup for training and evaluating the deep learning model used in the study. The results of the material classification model are presented and analyzed.

**Chapter Four:** focuses on the detection of material during laser cutting processes using deep learning techniques. It discusses the development of a deep learning model specifically designed to detect material and its integration into the laser cutting system. The chapter describes the methodology for training and testing the material classification model and presents the results of the classification accuracy. The findings are discussed in detail.

**Chapter Five:** explores energy reduction strategies in laser cutting. It investigates methods to minimize energy consumption by controlling components, such as air suction pumps and compressors. The chapter suggests the use of deep learning models to automatically adjust machine operations based on real-time conditions, resulting in energy savings. The methodology, experimental setup, and results of the proposed energy management strategies are presented and analyzed.

**Chapter Six:** concludes the thesis by summarizing the research findings, discussing the contributions made by the study, and their implications for the field. The chapter highlights the practical applications of the research and provides recommendations for further advancements in laser cutting energy reduction using deep learning techniques.

# Chapter 2: LITERATURE REVIEW

## 2.1 Overview and Background

This study presents an energy management approach based on deep learning techniques, since the utilization of deep learning techniques in the production and manufacturing processes is an industry 4.0 application. The literature review of this study relies on five main topics as shown in figure 2.1. The literature review chapter starts with an introduction to the study and the related technologies then follows by an overview about the industry 4.0. Then the related deep learning techniques with a detailed discussion about the material classification techniques and in-depth description of the laser speckle sensing technology together with its applications in the laser cutting industry.

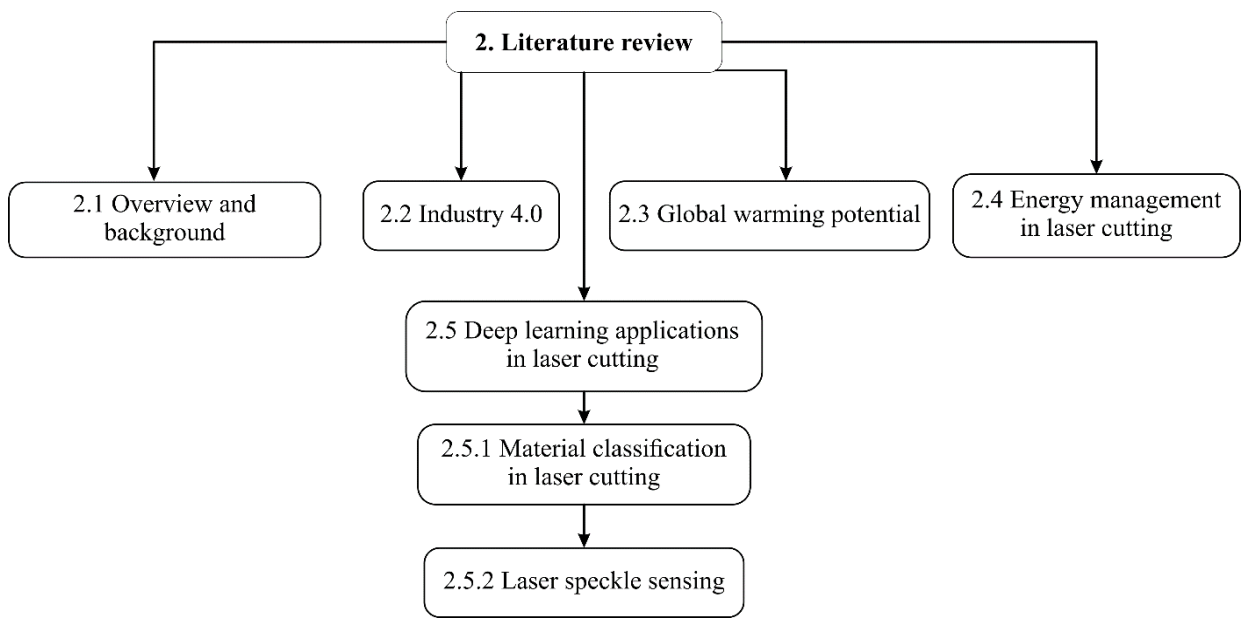

Figure 2.1 Outline of the literature review.

## 2.2 Industry 4.0

The Fourth Industrial Revolution, also known as Industry 4.0, represents a paradigm shift in production and manufacturing processes, leveraging advanced technologies such as Wireless Sensor Networks (WSN), Internet of Things (IoT), Augmented Reality (AR), Additive Manufacturing, Big Data, Data Analysis, and Deep Learning (DL). Figure 2.2 illustrates the related technologies, which are at the forefront of the fourth industrial revolution [10].

The adoption of Industry 4.0 has led to the emergence of smart factories, representing a new generation of manufacturing facilities. These smart factories incorporate various Industry 4.0 technologies in innovative ways to enhance productivity and improve the life cycle assessment (LCA) of products. By integrating intelligent systems, real-time data analytics, and automation, smart factories aim to optimize production processes, reduce downtime, and enhance overall operational efficiency.

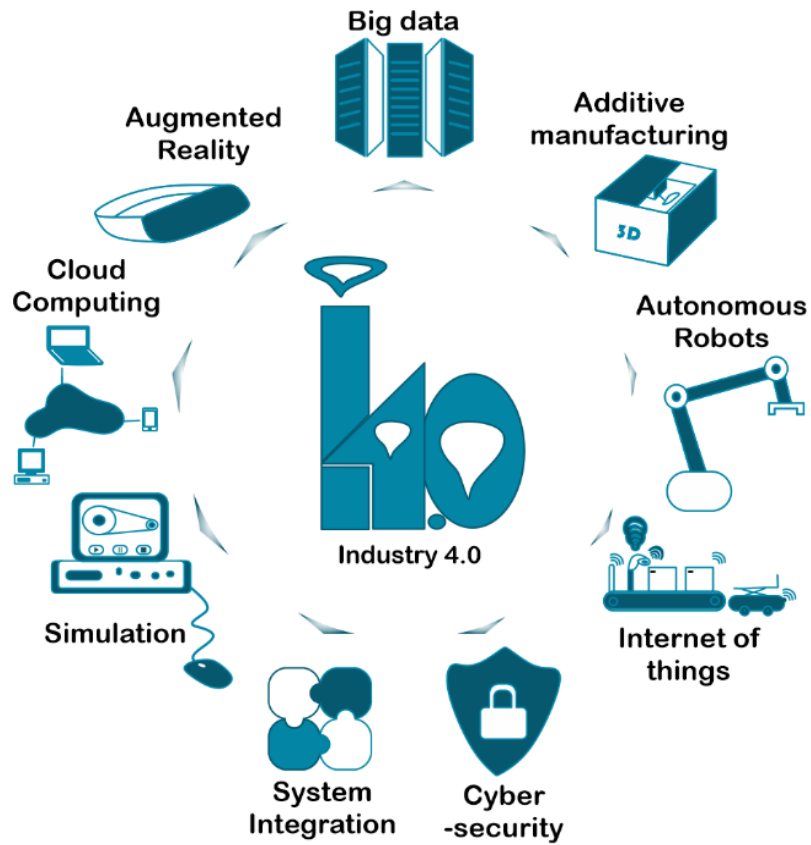

Figure 2.2 The technologies related to industry 4.0.

Industry 4.0 represents a paradigm where integrated factories are formed by connecting the product lifecycle with supply chain management [11]. It involves the utilization of digital technologies to gather and analyze real-time data, generating valuable insights for the system [12]. This information plays a crucial role in enhancing productivity, improving quality, and optimizing the product lifecycle assessment (LCA). Furthermore, the analysis of data and the generation of instantaneous information contribute to minimizing total energy consumption across the manufacturing process [13].

## 2.3 Global Warming Potential

The reduction of energy consumption and effective energy management have gained significant attention in both academic and professional domains due to their impact on Greenhouse Gas

Emissions (GHGE) and Global Warming Potential (GWP). Companies such as Daimler in Germany and Canadian Forest Products, which have implemented Industry 4.0 practices, have reported energy efficiency improvements of 30% and 15% respectively [14]. As fossil fuel combustion contributes approximately 67% of global $CO_2$ emissions, the continuous increase in concentration directly affects climate change and global warming [15-16]. In order to achieve sustainable outcomes, it is crucial to reduce the total energy consumed per unit of production [17-18].

The manufacturing sector is one of the industries with a direct impact on the environment, significantly contributing to the overall global impact. Considering the projected growth and rapid expansion of this industry, it is expected that the environmental impact will further increase [19].

In a critical review conducted in 2020 on forecasting global energy demand, the total amount of $CO_2$ emissions per million tons (MtCO2) resulting from industrial processes and fossil fuel combustion was calculated. The forecast covers the period from 2020 to 2040.

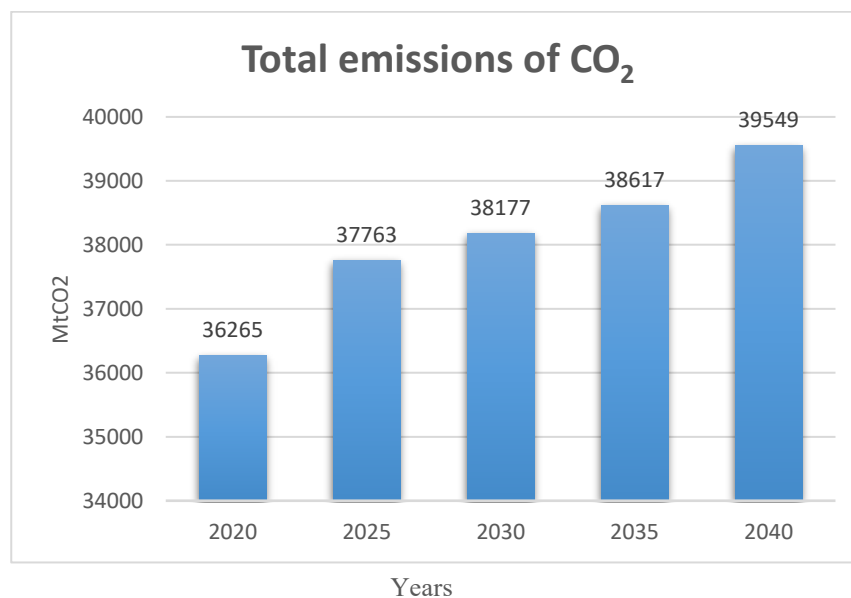

Figure 2.3 A forecasting for the total amount of $CO_2$ in the power generation and industrial processes according to results from [20].

The analysis of $CO_2$ emissions over the years indicates a consistent increase, and this trend is projected to continue in the coming years, as illustrated in Figure 2.3 [20].

In 2012, a new methodology was introduced to control the environmental impact and carbon dioxide emissions resulting from manufacturing processes. This methodology is based on the Cooperative effort on Process Emissions in manufacturing initiative (CO2PE!) and Unit Process

Life-Cycle Inventory (UPLCI), proposed a measurement approach to ensure consistency among datasets [21].

Building upon the CO2PE! the initiative, a model was developed in 2018 to calculate the powder reuse cycle in laser beam melting, as well as to assess energy consumption and resource utilization. The model also investigated the impact of material losses and powder recycling on the Global Warming Potential (GWP) [22].

Another study conducted in 2014 focused on environmental performance and CO2PE! methodology. This study analyzed three different types of laser cutters to enhance performance by analyzing the energy consumption and power profile of the machine tools during various production modes. The study depicted the variations in power demand from the laser source, exhaust system, total machine tool, and laser cooler, as shown in Figure 2.4 [7].

These research endeavors demonstrate the ongoing efforts to assess and improve the environmental performance of manufacturing processes by employing methodologies such as CO2PE! and UPLCI. By understanding the factors contributing to carbon dioxide emissions and energy consumption, it becomes possible to devise strategies for mitigating their impact and achieving more sustainable manufacturing practices.

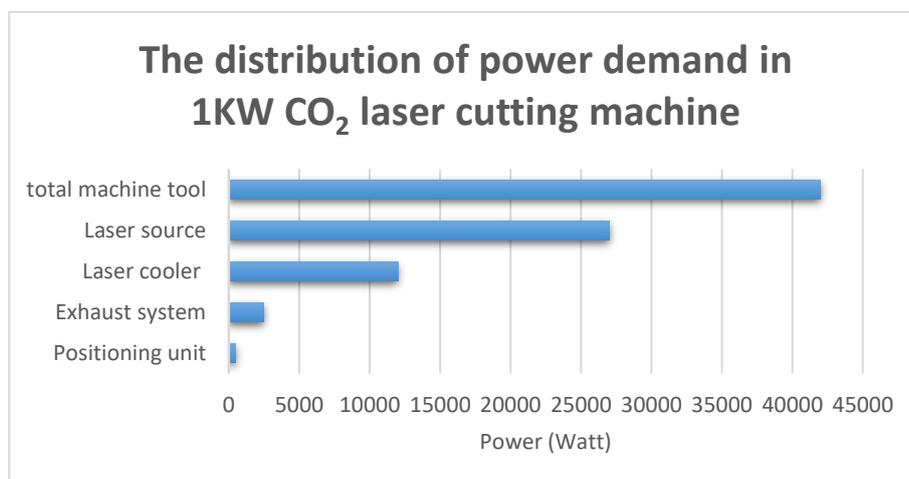

Figure 2.4 Power demand distribution from each component of the $CO_2$ laser cutter in the working mode at 1KW [7].

Therefore, additive manufacturing and laser cutting have a direct impact on environmental performance. In order to improve machine efficiency and energy performance, it is crucial to measure the power demand of each component of the machine. This allows for a strategic assessment of the environmental performance of laser cutting machines [7].

## 2.4 Energy Management in Laser Cutting Machine

In the field of engineering, the fabrication of sheet materials for use in the production of various components and products is a fundamental process. This fabrication is primarily facilitated by Computer Numerical Control (CNC) sheet cutting machines. These machines encompass a range of technologies, including thermal methods such as plasma, laser, and oxy-fuel cutting, as well as water jet cutting.

Among these technologies, CNC laser cutting has emerged as a widely adopted method owing to its exceptional precision and high efficiency. CNC laser cutting machines are frequently employed for their ability to deliver accurate results and impressive productivity levels. The operation of these CNC cutting machines relies on Numerical Control (NC) programs, which are meticulously crafted through the utilization of Computer-Aided Manufacturing (CAM) systems [22].

This integration of CNC technology, laser cutting precision, and CAM-driven program development plays a pivotal role in the modern engineering industry. It enables the fabrication of intricate components with a high degree of accuracy and repeatability, ultimately contributing to the advancement of engineering processes and the quality of end products [7].

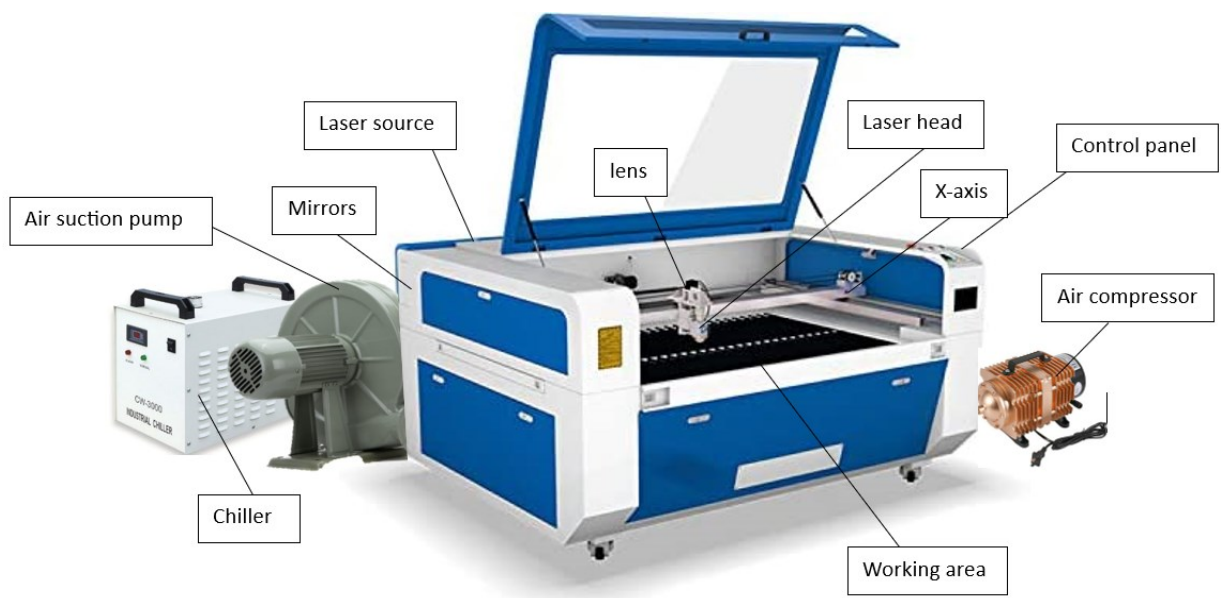

Figure 2.5 Components of $CO_2$ laser cutting machine.

Figure 2.5 shows the main components of a $CO_2$ laser cutting machine while the power distribution of each component is depicted in Figure 2.6, which is the power distribution profile of a 4 KW $CO_2$ laser cutting machine, as observed in the aforementioned study [7]. The figure highlights that the total power demand of the machine is not solely attributed to the laser source power. Instead, it comprises the laser source power, power consumed by the laser cooler, power of the exhaust system, and power of the positioning unit. Notably, the exhaust system operates at a constant power level without any control during the cutting process. Additionally, the exhaust system only initiates when the operator manually opens it before commencing cutting operations. During the cutting process, the exhaust system's power remains constant, and there is no variation in its power demand during the standby periods between different cutting operations.

It's important to mention that the laser power utilized in these results is 4 KW, which is the highest power level. However, when a low-power $CO_2$ laser tube (e.g., 80W or 100W $CO_2$ laser tube) is used in laser cutting machine, the power demand of the exhaust and cooling systems may surpass the power output of the laser source.

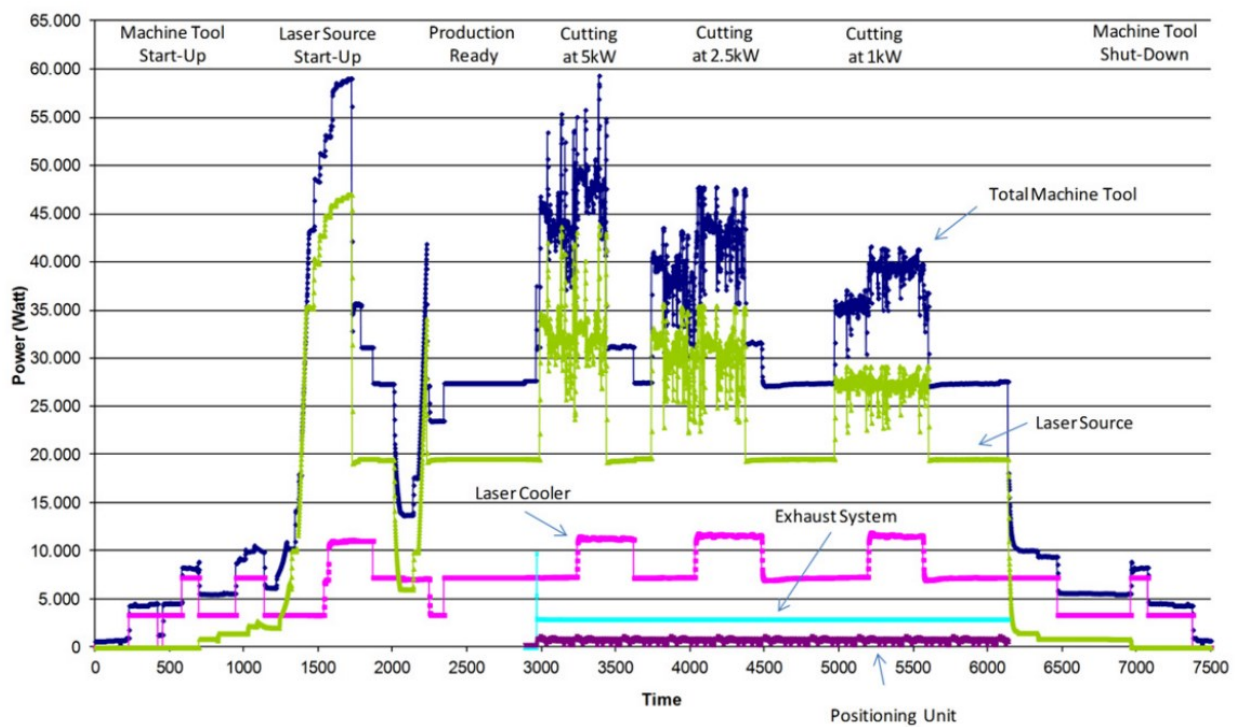

Figure 2.6 Power distribution pattern of a 4kW $CO_2$ laser cutting machine according to [7].

Furthermore, it is evident that the power demand of the exhaust system remains constant throughout the cutting process and even during the standby time of the laser source. However, it is crucial to highlight that the power demand of the exhaust system can be effectively managed through the implementation of a dedicated control system. By integrating a control mechanism that considers the specific material being cut, its properties, as well as the type and volume of dust, fumes, aerosols, and exhaust generated during the cutting process, it becomes possible to adjust the power demand of the exhaust system accordingly.

In this study, a 100W laser cutting machine was employed to implement and evaluate the proposed energy management approach. This dynamic control approach ensures the controlled operation of the exhaust system, aligning its power usage with the specific cutting requirements. Consequently, this approach not only enhances overall energy efficiency but also minimizes unnecessary power wastage, promoting a more sustainable and environmentally friendly laser cutting process.

## 2.5  Related work

### 2.5.1  Material Classification in Laser Cutting

Laser cutting is a widely used technology in many industrial sectors due to its high precision and efficiency [23]. However, the process generates harmful pollutants such as dust, smoke, and aerosols, which pose a risk to the environment and workers' health [24]. To ensure safe and efficient laser cutting, it is crucial to monitor and control the process in real time. Speckle sensing has emerged as a promising method for monitoring laser cutting and classifying materials by analyzing the speckle pattern produced by the laser on the surface structure of the used material [25], it is possible to extract valuable information about the material and the cutting process. Deep learning techniques have displayed immense capabilities in the past few years for analyzing speckle patterns and categorizing different materials [26,27].

The utilization of laser cutters in workshops is a prevalent practice, however, it comes with its own set of challenges. Although there are various support tools available to assist operators in laser cutting tasks, such as PacCAM [28] a tool for packing parts according to the placed sheet inside the laser cutter, Fabricaide [29] also proposed another tool that integrates the creation and preparation of designs for fabrication, despite the increasing availability of diverse materials for laser cutting, there remains a shortage of systems that assist operators in effectively navigating and selecting appropriate cutting parameters for each material type. Users face difficulties in finding unmarked sheets from material inventories or spare parts in a laser cutting workshop since many materials share a similar visual appearance, like transparent materials (e.g., Acrylic, PETG, and Acetate) [30].

Consequently, users may mistakenly choose the wrong material from the stack and apply the incorrect power and speed settings, which can lead to material wastage or pose a risk to the environment and workers' health because there are numerous materials that are not safe for laser cutting due to the released toxic fumes [31,32]. Unfortunately, the similarity in appearance between safe and hazardous materials can lead to hazardous materials being mistaken for safe ones.

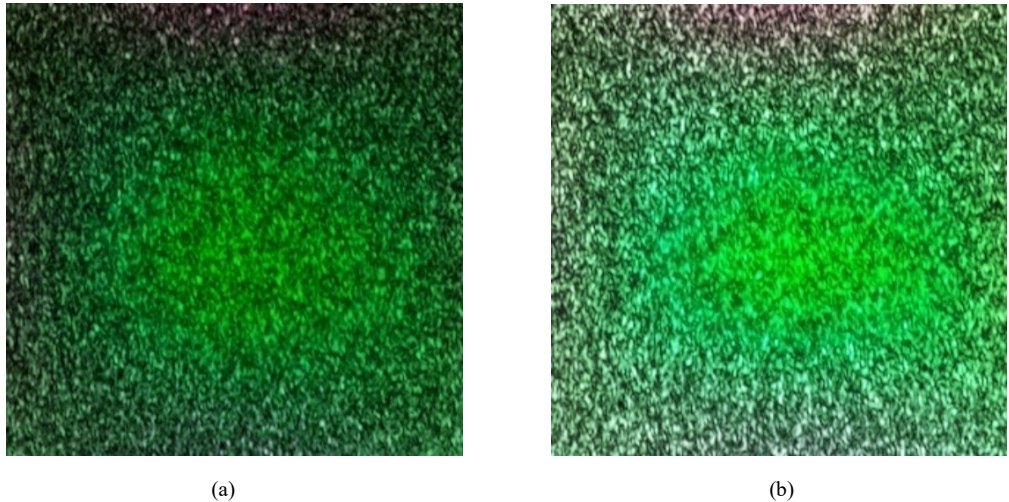

(a)                  (b)

Figure 2.7 Two different samples from the SensiCut dataset. (a) Speckle patterns of Oak Hardwood. (b) Speckle patterns of MDF [25].

To address this issue, a camera can be added to the laser cutter to recognize materials, In the SensiCut study by Dogan et al. [25], a lens-less camera and deep learning techniques were utilized to classify laser cutting materials based on their surface structure using speckle patterns. They created a dataset of 30 different classes of materials, some samples from the dataset are shown in Figure 2.7.

### 2.5.2 Laser Speckle Sensing

Laser speckle sensing is a non-contact optical technique that works by shining a laser beam onto a rough surface. When the laser light hits the surface, it is scattered in all directions due to the surface's roughness. The scattered light waves interfere with each other, creating a pattern of bright and dark spots known as a speckle pattern [33]. The speckle patterns contain valuable information about the surface, such as its roughness, texture, and movement. By analyzing the statistics of the speckle patterns, laser speckle sensing can provide high-resolution measurements of the surface properties in real-time [34].

This technique can be used in a wide range of applications, including surface inspection, material characterization, biological imaging, and water content [35], due to its ability to provide detailed information about surface properties without making physical contact with the surface itself [25], as shown in figure 2.8, where a laser beam is directed at a material's surface structure, and the reflected beams are captured by an image sensor. The interference between the reflected rays in different phases produces a speckle image. An overview of the main components of laser speckle sensing is shown in Figure 2.8.

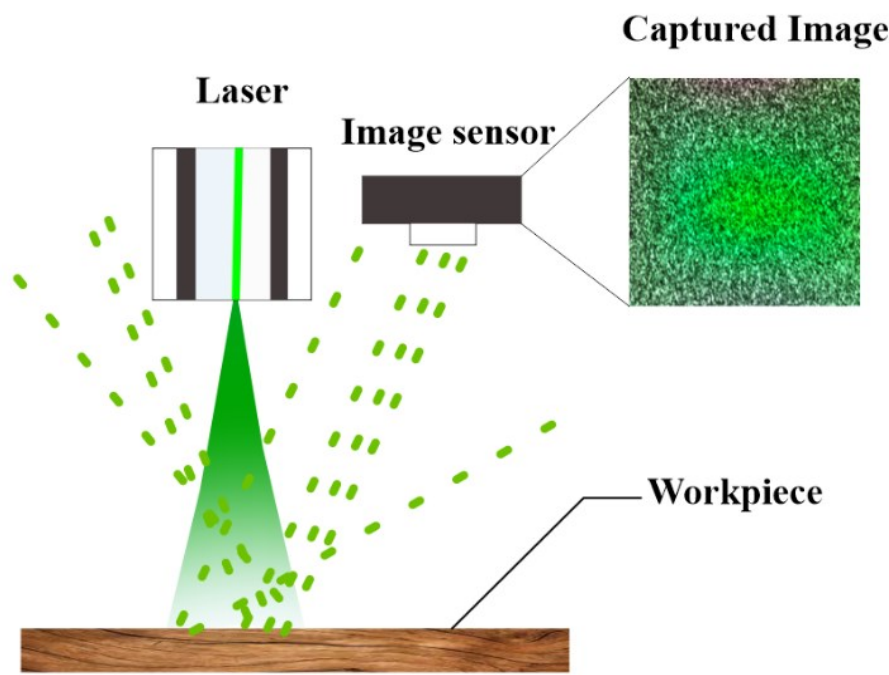

Figure 2.8 Components of laser speckle sensing.

### 2.5.3 Convolutional Neural Networks

Convolutional Neural Networks (CNNs), often referred to as convnets, have showcased remarkable capabilities in addressing complex visual classification tasks. Notably, they have excelled in tasks such as image recognition, object recognition and handwritten digit classification [36,37]. Several factors underlie their outstanding performance, including their intrinsic high-capacity architecture, harnessing parallel computing environments, and the availability of extensive training datasets.

A fundamental mechanism employed by a convnet involves the application of nonlinear filter stages to input images, enabling the transformation of pixel information into a distinctive feature space. Each filter stage combines an image convolution operation with a filter, followed by the introduction of a nonlinearity function, like tanh(·). Subsequently, the outcome undergoes down-sampling via max-pooling or average-pooling, and the result is then normalized. During the training phase, the convnet systematically optimizes its parameters based on the provided input data [37].

The features learned through this process are subsequently channelled into a classifier, such as a Multi-Layer Perceptron (MLP) or a Support Vector Machine (SVM). These classifiers are responsible for categorizing the content within an image, thereby demonstrating the versatility and effectiveness of CNNs in a broad spectrum of visual recognition tasks [37].

Contrary to the common belief that training and test data must share the same feature space and distribution, this assumption does not hold in numerous real-world scenarios. Similarly, the human visual system adeptly manages diverse visual tasks using a unified approach, suggesting the utilization of a common method for various visual recognition tasks. This knowledge transfer is recognized as transfer learning. From that end a pretrained models such as AlexNet, ResNet50, and VGG16 [38] are proposed and showed a promising performance in transfer learning applications, these models are trained on ImageNet dataset [39].

## 2.6  Research Gap and Thesis Main Contribution Objectives

As mentioned in Section 2.4, the total power demand of laser cutting machines includes the laser source power, power consumed by the laser cooler, power of the exhaust system, and power of the positioning unit. In the case of low-power $CO_2$ laser cutting machines designed for processing non-metallic materials such as wood, plastic, textile, and leather, the power of the exhaust system can exceed that of the laser source. Typically, the laser source in these machines operates in the power range of 50 to 300 W, while the exhaust system's power demand ranges from 0.35 to 1.5 kW. An inefficiency is further compounded by the absence of a control strategy for the exhaust system, resulting in constant power demand, even when no exhaust by-products, dust, fumes, smoke, or gases are generated during the cutting process.

Addressing this energy inefficiency is the primary objective of this study. The study proposes a control strategy for the exhaust suction pump, which utilizes a deep learning model in conjunction with speckle sensing to classify materials before the cutting process. This classification aids in determining the optimal power requirement for the exhaust system. Additionally, the study introduces an innovative approach to generate speckle patterns using an affordable USB camera paired with a red laser pointer. Furthermore, it contributes to the field by creating a brand-new dataset containing speckle pattern images captured using a red laser and a USB camera, as well as a dataset of smoke images from laser cutting machines. Most significantly, this study introduces a custom CNN model specifically designed for speckle sensing. Overall, it employs three different deep-learning models to manage and minimize the energy consumption in laser cutting machines. The flow of the proposed models throughout this study is presented in Figure 2.10. The use of multiple detection models boosts classification accuracy, and solely detecting the produced smoke is not sufficient since certain materials may not produce visible smoke but can still emit odors and invisible exhaust, such as when cutting an acrylic sheet. The objectives of this research aim to bridge the gap in energy-efficient laser cutting and contribute to more sustainable manufacturing processes.

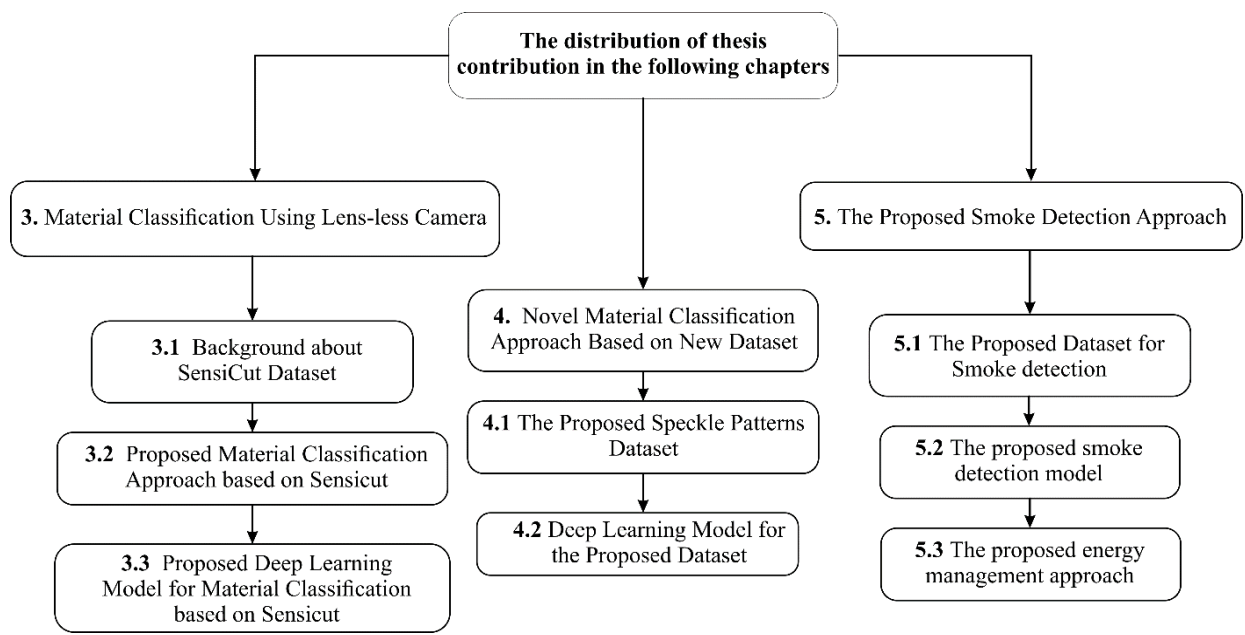

Figure 2.9 The distribution of thesis contribution in the following chapters.

# Chapter 3: MATERIAL CLASSIFICATION USING LENSS-LESS CAMERA

## 3.1 Overview and Background

Since the required power and speed of the laser source vary depending on the material being cut, and the amount of smoke and exhaust produced during the cutting process also changes with the material being processed, an important study conducted by MIT's SensiCut team [25] has introduced an innovative method for classifying laser cutting materials. This approach combines speckle sensing and deep learning to accurately classify materials based on their speckle patterns. Figure 3.1 visually illustrates this innovative approach, which enables precise material classification, even when materials have visual similarities. Consequently, it enhances the overall efficiency and effectiveness of material classification in laser cutting operations.

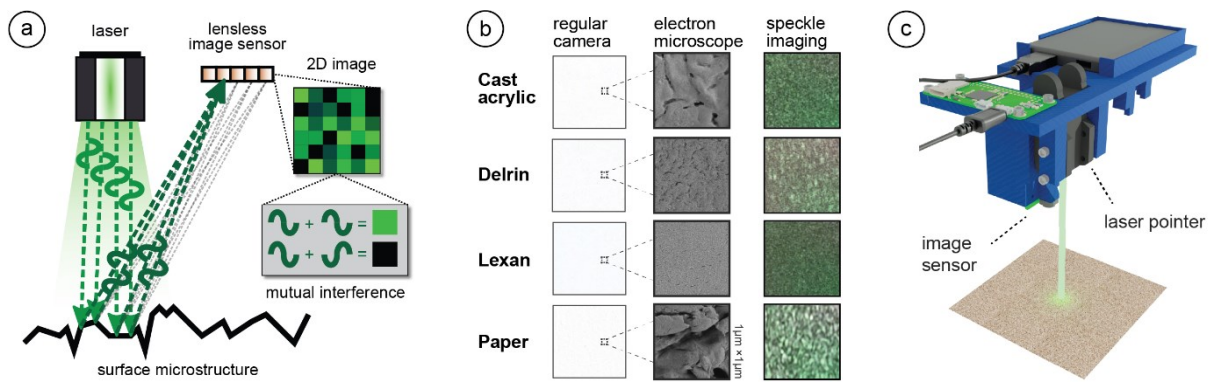

Figure 3.1 Speckle sensing explanation according to [25], (a) an illustration for the speckle sensing technology, (b) a comparison between the regular, microscopic, speckle pattern images for visually similar materials, (c) the proposed an-in by Sensicut study [25].

Before delving into the implementation of a CNN model, it is valuable to list the main components and steps to be considered when training and implementing a CNN model. These steps involve performing the following tasks: -

1. Data Collection: Gather a dataset that is relevant to your problem, with labeled examples.
2. Preprocessing the Collected Data: Preprocess the data, including resizing, normalization, and data augmentation to ensure it's suitable for training.
3. Model Architecture Design and Feature Extraction: Decide on the overall architecture of the CNN, including the number of layers, type of layers, and their connectivity.
4. Classifier: to learn high-level features and perform classification or regression.
5. Training: Train the model using the preprocessed dataset, and monitor training metrics (e.g., accuracy, loss) to assess the model's performance.

6. Fine Tuning: Choosing an appropriate loss function, optimization, activation functions, and hyperparameters based on the task, such as mean squared error for regression or cross-entropy for classification.
7. Evaluation: Evaluate the model on a separate validation dataset to assess its generalization ability.
8. Testing: Use the trained model to make predictions on new, unseen data.
9. Model Deployment: Deploy the model in an application or a system to make real-time predictions.

## 3.2 Material Classification using Sensicut Dataset.

### 3.2.1 Background about Sensicut dataset

The SensiCut dataset, which is available on Kaggle platform [25], contains 39,609 images classified into 59 different categories. These categories belong to 30 distinct material types, some of which, such as acrylic, are available in multiple colors. Therefore, the dataset contains more than 30 directories, although the classification is only made for 30 material types, the distribution of the materials in the Sensicut dataset is demonstrated in Figure 3.2.

The dataset consists of 649 speckle pattern images for each class, these speckle patterns are green due to the use of a laser with a wavelength of 515nm, as shown in Figure 3.1, with image resolution of 800 pixels by 800 pixels. The color of the generated speckle patterns depends on the wavelength of the used laser, with the most critical channel among the three RGB layers corresponding to the color of the laser source. This study conducted an investigation to demonstrate that the color of the generated speckle patterns consistently depends on the wavelength and color of the laser source used in their generation. As a result, Figure 3.3 displays three distinct speckle pattern images captured by a lens-less Raspberry Pi camera (see appendix A), each using a different laser pointer.

In the first image, a laser with a wavelength of 515 nm was used, resulting in a green image. The second image was generated using a 532 nm laser, producing green speckles. Conversely, when a laser with a wavelength of 650 nm was employed, the speckles appeared in red in the captured image. In summary, the color of the speckle patterns in the generated images corresponds to the color or wavelength of the laser source being used.

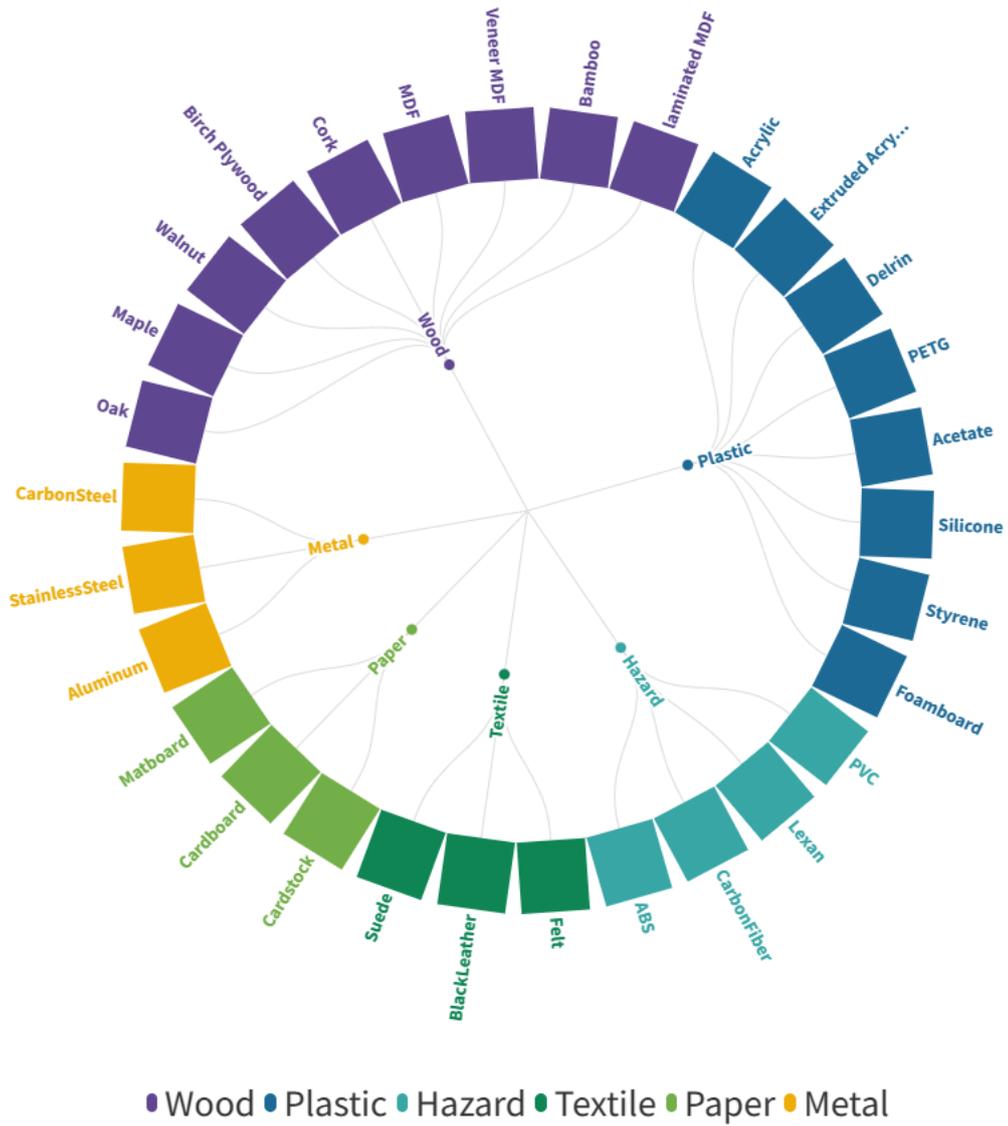

Figure 3.2 Distribution of the materials used in SensiCut dataset.

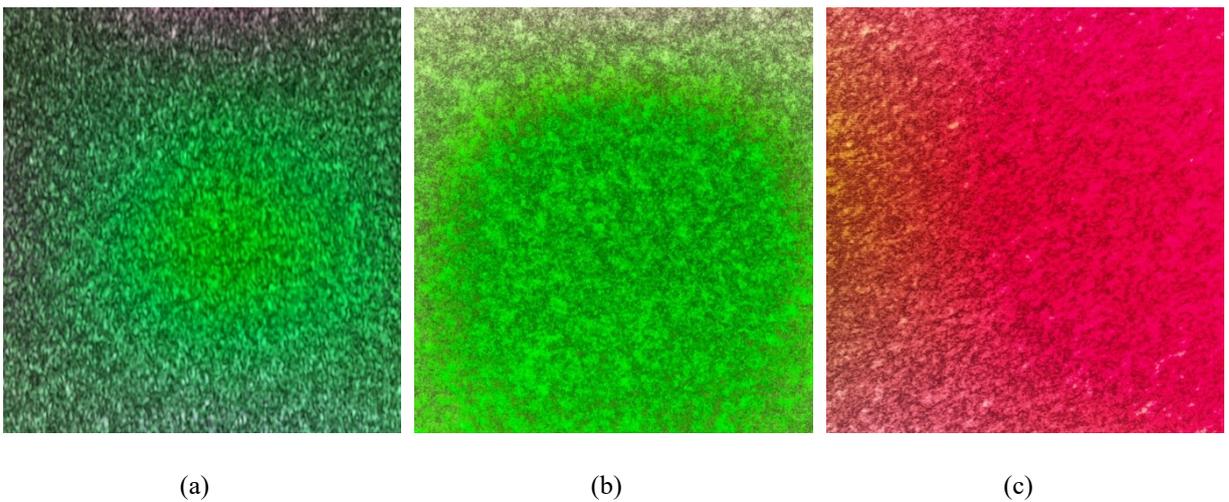

(a)            (b)            (c)

Figure 3.3 Three different laser sources are used to produce speckles. (a), (b), and (c) correspond to lasers with wavelengths of 515nm, 532nm, and 650nm, respectively.

This study proposes a new simple technique for material classification using speckle sensing. Unlike traditional methods that use all three RGB channels of the speckle pattern images for training a convolutional neural network, the proposed approach only utilizes the layer corresponding to the laser color. This approach not only reduces the training time of the model but also makes the method more efficient by reducing the inference time. The suggested methodology also addresses the challenge of using different laser sources for speckle pattern generation. By utilizing the layer associated with the laser's color, features can be extracted from a single color channel, resulting in a reduction of dimensions in the input layer of the deep learning model. This approach enables the exclusive utilization of speckle patterns from the laser's color channel, disregarding other color channels. Consequently, a more practical solution is provided.

### 3.2.2 Proposed Approach for Material Classification

The previous section explained that the main concept behind this study is to utilize solely the layer that corresponds to the used laser during the speckle pattern-capturing process.

In Figure 3.4, a sample image of Maple Hardwood from the SensiCut dataset is shown, with the original RGB image and the individual red, green, and blue layers displayed separately for comparison.

The visual comparison presented in Figure 3.4 clearly shows that the green channel of the sample image of Maple hardwood appears to be the most informative layer for material classification. The green layer provides a similar pattern to the original image and appears to contain the most significant features for distinguishing between different materials. In contrast, the other color layers seem to be either noisy or could be neglected for the classification task. Therefore, the proposed approach that utilizes solely one layer for material classification is expected to provide better accuracy and faster inference time compared to the traditional method that utilizes all three RGB channels.

To validate the effectiveness of the proposed approach, experiments were utilized SensiCut dataset, which comprises 39,609 speckle pattern images, each corresponding to 30 different material types. In order to assess the proposed method, it was compared against the Sensicut model as a baseline model [25]. Sensicut utilized all three RGB layers for material classification by means of transfer learning from a pre-trained ResNet-50 model [36]. The experiments showed that the proposed approach achieved higher accuracy and faster inference time compared to the baseline model. Specifically, the proposed approach achieved an accuracy of 98.3%, while the Sensicut model achieved an accuracy of 98.01%. Moreover, the proposed approach required only 13.5% of the time required by the Sensicut model for inference.

The results shown in section 3.3 demonstrate the effectiveness of the proposed approach for material classification using speckle sensing, which provides a viable solution for reducing the required time for model training and inference while maintaining high accuracy levels. Furthermore, it offers a solution to the issue of changing the used laser in generating speckle patterns, allowing for more flexibility and ease of use in practical applications.

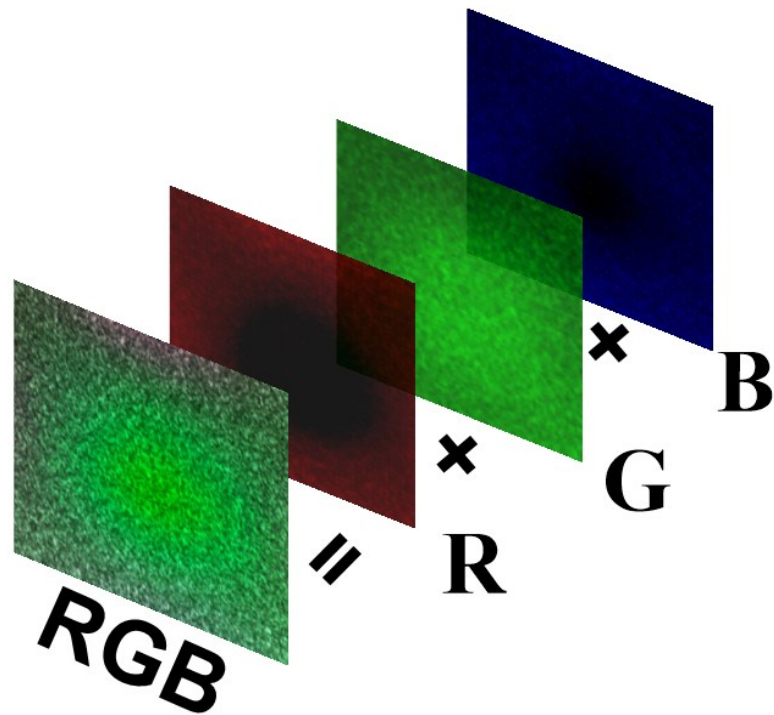

Figure 3.4 The original RGB image of Maple Hardwood from the SensiCut dataset. And its blue, green, and red channels.

### 3.2.3 Proposed Deep Learning Model for Material Classification

The material classification approach proposed in this study utilizes a CNN to learn the distinctive features of speckle pattern images from the Sensicut dataset. The architecture of the CNN model employed in this research is depicted in Figure 3.16, and a detailed description of its components is presented in the following context. However, before delving into the details of the deep learning model, it is essential to provide a comprehensive explanation of the image representation and the pre-processing steps employed, as illustrated in Figure 3.5.

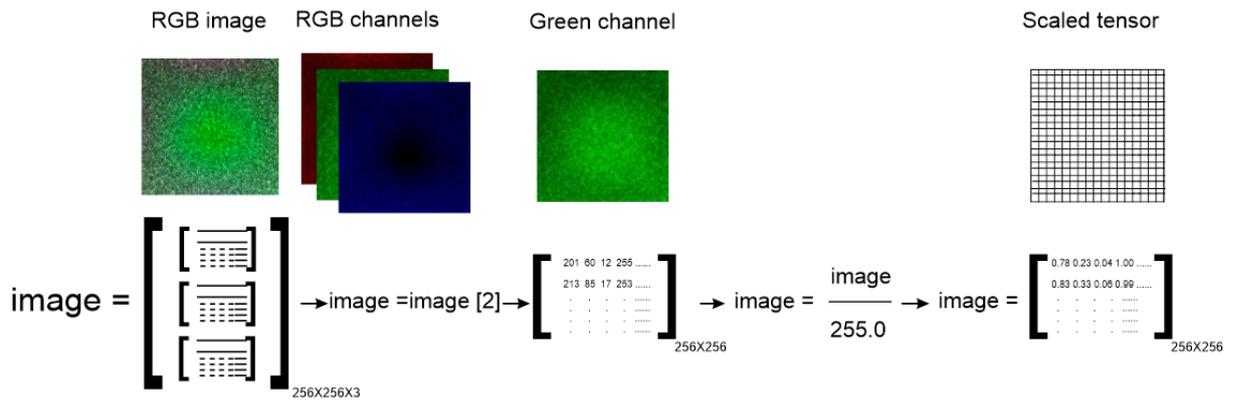

Figure 3.5 The proposed pre-processing steps on the input images.

The architecture is comprised of four convolutional layers followed by pooling layers, followed by two fully connected (dense) layers. The primary purpose of using convolutional layers is to extract features from the speckle pattern images through the application of filters or kernels. These filters perform convolution operations by conducting element-wise multiplications on the input image, as illustrated in Figure 3.6.

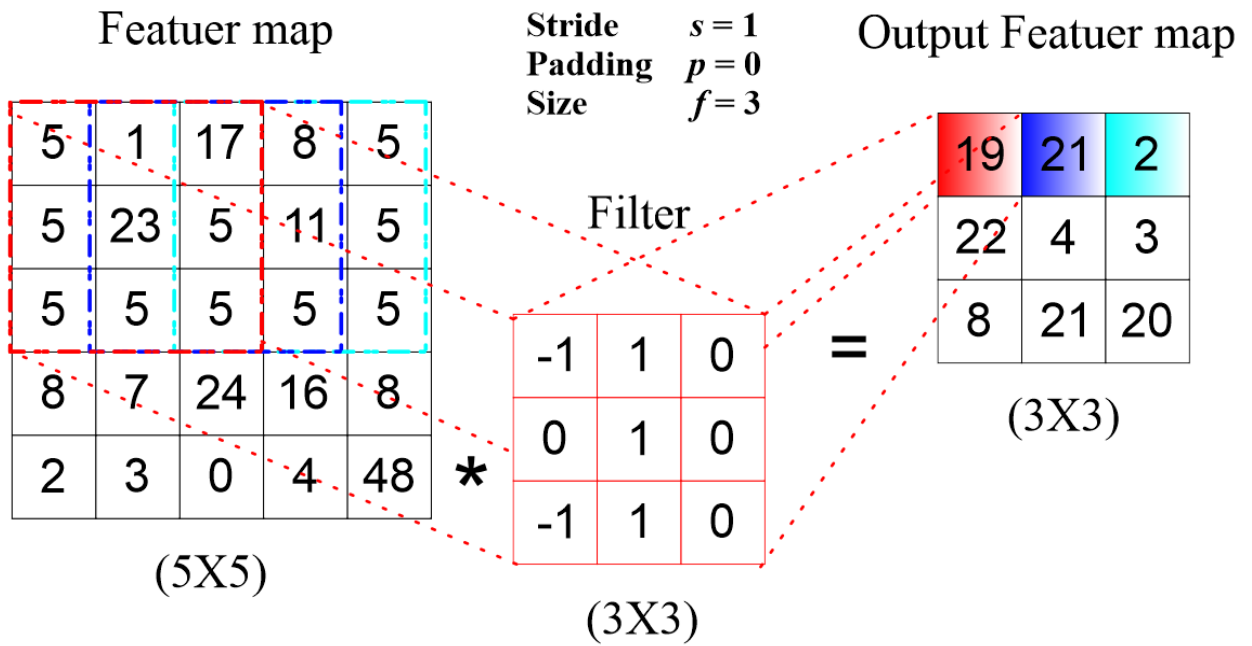

Figure 3.6 An explanation of the convolution process for a 5 X 5 feature map and a 3 X 3 filter.

The convolution process generates feature maps based on the number of filters or kernels employed in the convolutional layer. The resulting feature map of a convolutional process can be calculated using Equation (3.1) [37].

$$z_{i,j,k} = b_k + \sum_{u=0}^{f_h-1} \sum_{v=0}^{f_w-1} \sum_{k'=0}^{f_{n'}-1} x_{i',j',k'} \times w_{u,v,k',k} \quad \text{with} \quad \begin{cases} i' = i \times s_h + u \\ j' = j \times s_w + v \end{cases} \tag{3.1}$$

here $z_{i,j,k}$ represents the output of a neuron situated at row $i$, column $j$, in feature map $k$ of the convolutional layer $l$. $s_h$ and $s_w$ denote the vertical and horizontal strides, while $f_h$ and $f_w$ represent the height and width of the receptive field, respectively. Additionally, $f_{n'}$ corresponds to the number of feature maps.

The variable $x_{i',j',k'}$ signifies the output of a neuron located at layer $l - 1$, row $i'$, column $j'$, and feature map $k'$. For the first layer channel $k'$ instead of feature map $k'$.

$b_k$ which acts as a bias term for feature map $k$ in layer $l$, as a control knob that fine-tunes the overall brightness of feature map $k$.

Lastly, $w_{u,v,k',k}$ denotes the weight between any neuron in feature map $k$ of layer $l$ and its corresponding input situated at row $u$ and column $v$, and feature map $k'$[37].

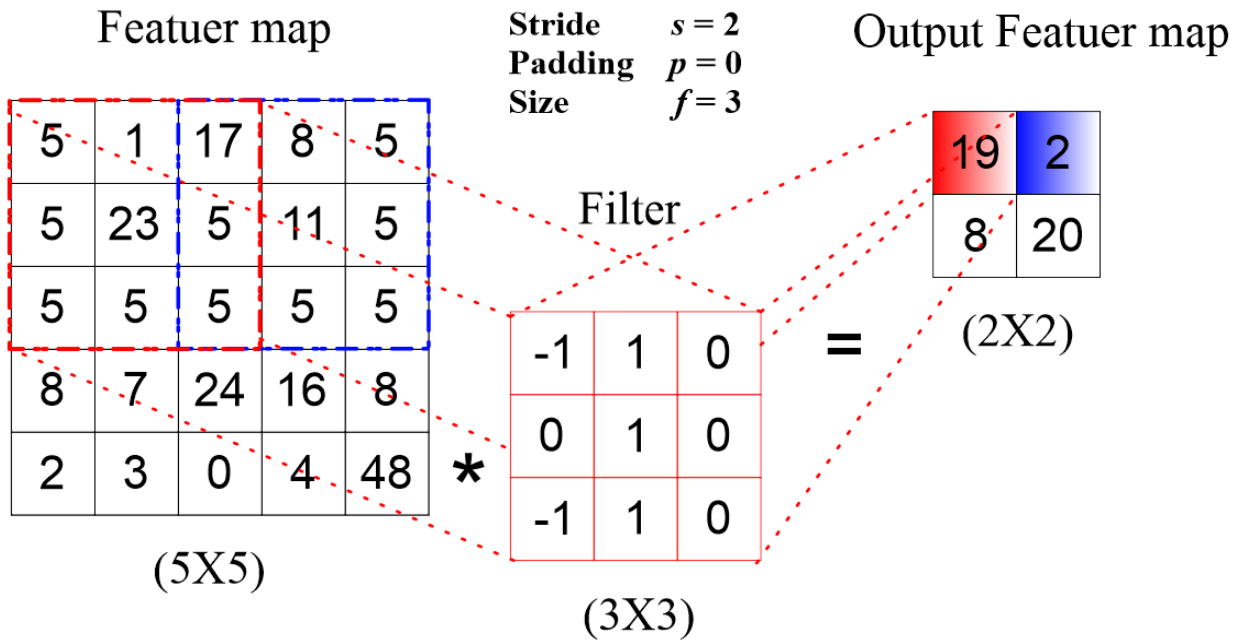

Figure 3.7 Further explanation for the convolution process while using a stride of s = 2.

The output feature map resulting from the convolutional process is observed to undergo a 2 X 2 reduction, as depicted in Figure 3.7. This reduction occurs when a stride value of $s = 2$ is employed, compared to the otput feature map illustrated in Figure 3.6 where the stride $s = 1$, the dimention of the output featuer map is observed to undergo a 3 X 3 reduction according to equation 3.2.

$$m' = \frac{m - f_h + 2p}{s} + 1 \qquad (3.2)$$

Where $m'$ is the hight of the output feature map, $f_h$ is the hight of the receptive filed, $p$ is the padding used in the hight, $s$ is the stride used in the hight, and $m$ is the hight of the input feature map.

Since equation 3.2 is used to calculate the hight of an output feature map, accordingly the following equation 3.3 used to calculate the width of an output feature map.

$$n' = \frac{n - f_w + 2p}{s} + 1 \qquad (3.3)$$

Where $n'$ is the width of the output feature map, $f_w$ is the width of the receptive filed, $p$ is the padding used in the width, $s$ is the stride used in the width, and $n$ is the width of the input feature map.

The convoluation process not only relies on the padding and stride between the input featuer map and the reciptive field, however the value each element inside the output featuer map comes from a dot product used between the reciptive field and the input featuer map. Since the $n', m'$ are the dimensions of the output feature map $O_f$, then the output feature map can be written as follows.

$$O_f = O_i * f = \begin{bmatrix} O_1 & O_2 & O_3 \\ O_4 & O_5 & O_6 \\ O_7 & O_8 & O_9 \end{bmatrix}_{n' X m'} \qquad (3.4)$$

If the input featuer map $A_f$ is 5 X 5 as shown in figure 3.6 and figure 3.7.

$$A_i = \begin{bmatrix} A_1 & A_2 & A_3 & A_4 & A_5 \\ A_6 & A_7 & A_8 & A_9 & A_{10} \\ A_{11} & A_{12} & A_{13} & A_{14} & A_{15} \\ A_{16} & A_{17} & A_{18} & A_{19} & A_{20} \\ A_{21} & A_{22} & A_{23} & A_{24} & A_{25} \end{bmatrix} = \begin{bmatrix} 5 & 1 & 17 & 8 & 5 \\ 5 & 23 & 5 & 11 & 5 \\ 5 & 5 & 5 & 5 & 5 \\ 8 & 7 & 24 & 16 & 8 \\ 2 & 3 & 0 & 4 & 48 \end{bmatrix}$$

And the filter or kernel $f$ is 3X3 and have the following values:

$$f = \begin{bmatrix} f_1 & f_2 & f_3 \\ f_4 & f_5 & f_6 \\ f_7 & f_8 & f_9 \end{bmatrix} = \begin{bmatrix} -1 & 1 & 0 \\ 0 & 1 & 0 \\ -1 & 1 & 0 \end{bmatrix}$$

If the stride $s = 1$ and no padding used $p = 0$, then the second element of the output feature map $O_2$ can be calculated according to the following equation 3.5.

$$A' = \begin{bmatrix} f_1 & f_2 & f_3 \\ f_4 & f_5 & f_6 \\ f_7 & f_8 & f_9 \end{bmatrix} \odot \begin{bmatrix} A_2 & A_3 & A_4 \\ A_5 & A_6 & A_7 \\ A_8 & A_9 & A_{10} \end{bmatrix} \qquad (3.5)$$

$$O_2 = \sum_{i=1}^{f_w} \sum_{j=1}^{f_h} A'_{ij} \qquad (3.6)$$

Where $O_2$ is the second element in output featuer map and $A'_{ij}$ is the ouput matrix from element wise multiplication between reciptive filed from the input featuer map and the used filter as shown in equation 3.5, also $i$ and $j$ are the rows and columns of the matrix $A'$.

According to equation 3.6 the second elemnt of the output featuer map $O_2$ while the stride $s = 1$ and no padding used $p = 0$, is obtained by equation 3.7.

$$O_2 = A_2 f_1 + A_3 f_2 + A_4 f_3 + A_7 f_4 + A_8 f_5 + A_9 f_6 + A_{12} f_7 + A_{13} f_8 + A_{14} f_9 \qquad (3.7)$$

In contrast, If the stride $s = 2$ and no padding used $p = 0$, then the second element of the output feature map $O'_2$ can be calculated according to the following equation 3.8.

$$O'_2 = A_3 f_1 + A_4 f_2 + A_5 f_3 + A_8 f_4 + A_9 f_5 + A_{10} f_6 + A_{13} f_7 + A_{14} f_8 + A_{15} f_9 \qquad (3.8)$$

The dimensions of the input image as shown in figure 3.5 are 256 X 256 X 1, where only one-color channel is used from the input image, corresponding to the laser color used during the speckle pattern-capturing process.

The first layer of the model performs a convolution process using 32 filters each with 3 X 3 dimension, with no padding $p = 0$ and strid $s = 1,$ the output feature map dimensions can be calculated be equations 3.2 and 3.3.

$$\frac{256 - 3 + (2 * 0)}{1} + 1 = 254$$

Since the size of the input image in Sensicut model was 256, the same size is used in the proposed model. Then the previous equations 3.2 and 3.3 were applied to calculate the size of the output feature maps from the first layer, which results 254 X 254 feature map as shown in figure 3.8.

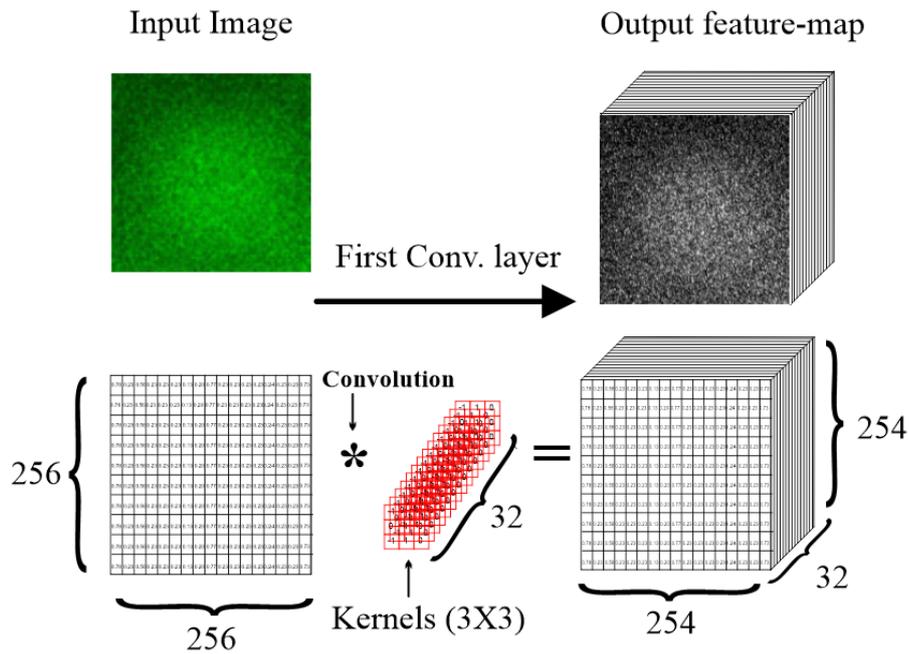

Figure 3.8 Explanation for the first convolutional layer.

After calculating the weights (elements of the output feature maps), an activation is applied to them by the neurones in the next layer, the rectified linear unit (ReLU) activation function has been used to activate the input weights in this layer.

Despite the nun saturation behaviour of the ReLU activation function as shown in equation 3.9 and figure 3.9, it performs faster than the sigmoid (figure 3.10) and hyperbolic tangent (tanh) function (figure 3.11).

$$f(x) = \begin{cases} x & x \geq 0 \\ 0 & x < 0 \end{cases} \qquad (3.9)$$

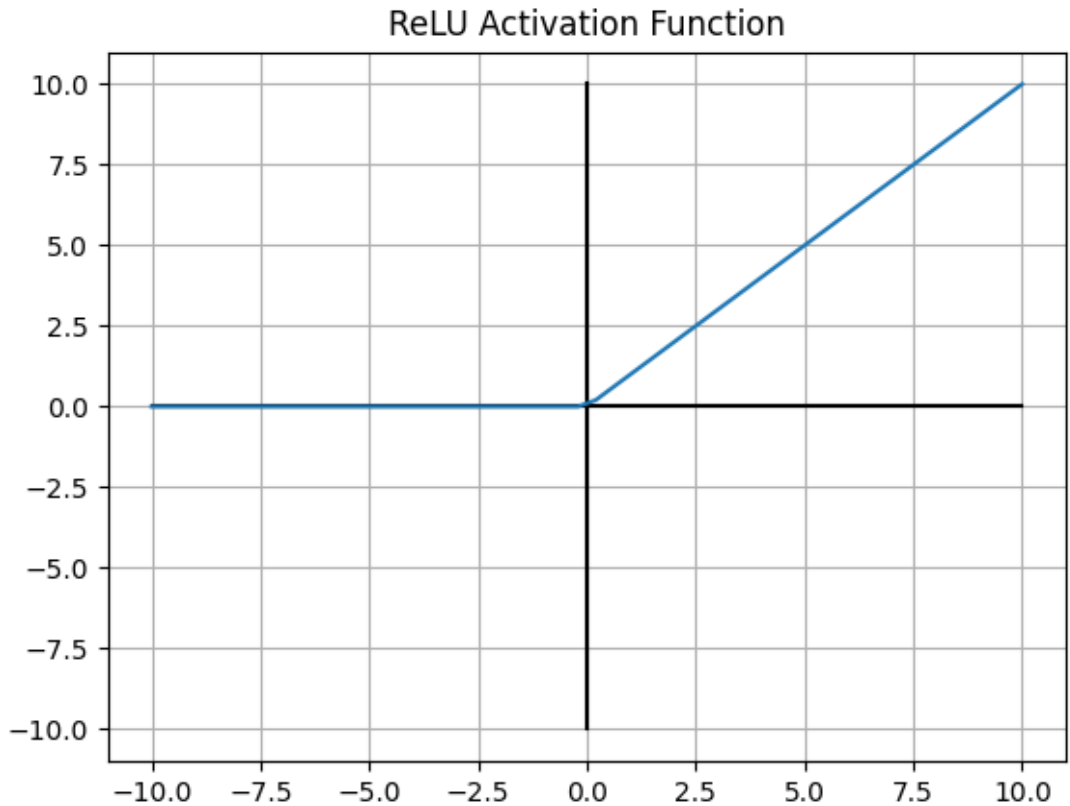

Figure 3.9 Relu activation function.

In contrast and as shown in Figure 3.9 and Equation 3.11, the sigmoid activation function saturates and this makes sigmoid activation function a good choice to activate the output layer of the binary classification models. The main drawback behind using sigmoid activation function with the hidden layers of the deep learning models is the occurrence of vanishing gradients as the bigger values of the input weights make the gradient of the sigmoid function vanishes.

$$f(x) = \frac{1}{1 + e^{-x}} \qquad (3.10)$$

It is also worth noteing that there are another activation functions like the hyperbolic tangent function and SoftMax activation function which will be mentioned later in the output layer. The hyperbolic tansh function Equation 3.11 used as an activation function specially with the recurrent nearual networks (RNN) and long short term memory (LSTM) cells.

$$f(x) = tanh(x) \qquad (3.11)$$

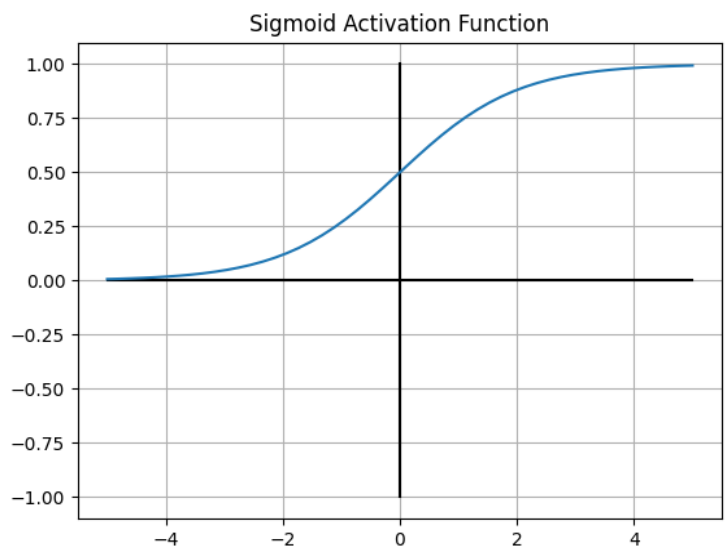

Figure 3.10 Sigmoid activation function.

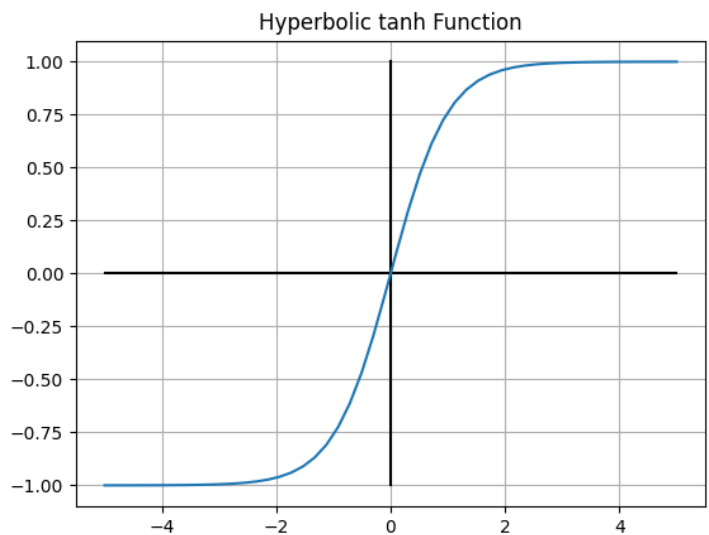

Figure 3.11 Hyperbolic tangent activation function (tanh).

After the first convolutional layer, a max-pooling layer has been used to reduce the spatial dimension by half. Max pooling layer utilises a window (kernel) of size 2X2 to slides across the feature maps with the specified stride then outputs the maximum number of the area covered by the kernel. The description of Max-pooling is illustrated in figure 3.12.

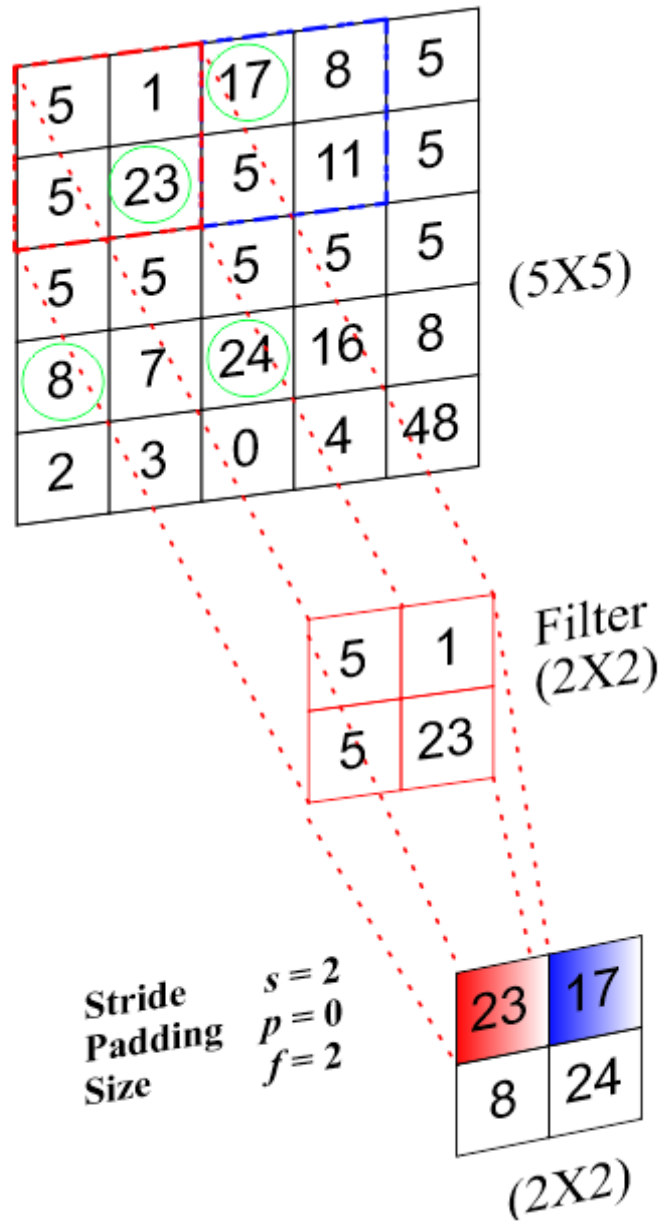

Figure 3.12 Max pooling by 2 x 2 kernel with no padding and stride s = 2.

The max pooling layers in this study uses a stride of 2, also no padding has used in the max pooling layers, so the output of each max pooling layers in the used deep learning model is reduces the size of feature maps by half according to Equations 3.2 and 3.3, accordingly, the first max pooling layer reduces the size of the 32 feature maps from 254 X 254 to 127 X 127 as depicted in figure 3.13. Since the max pooling layers are non-trainable layers there is no need to activate these layers by an activation function, in contrast with the convolutional layers where there are trainable weights used in.

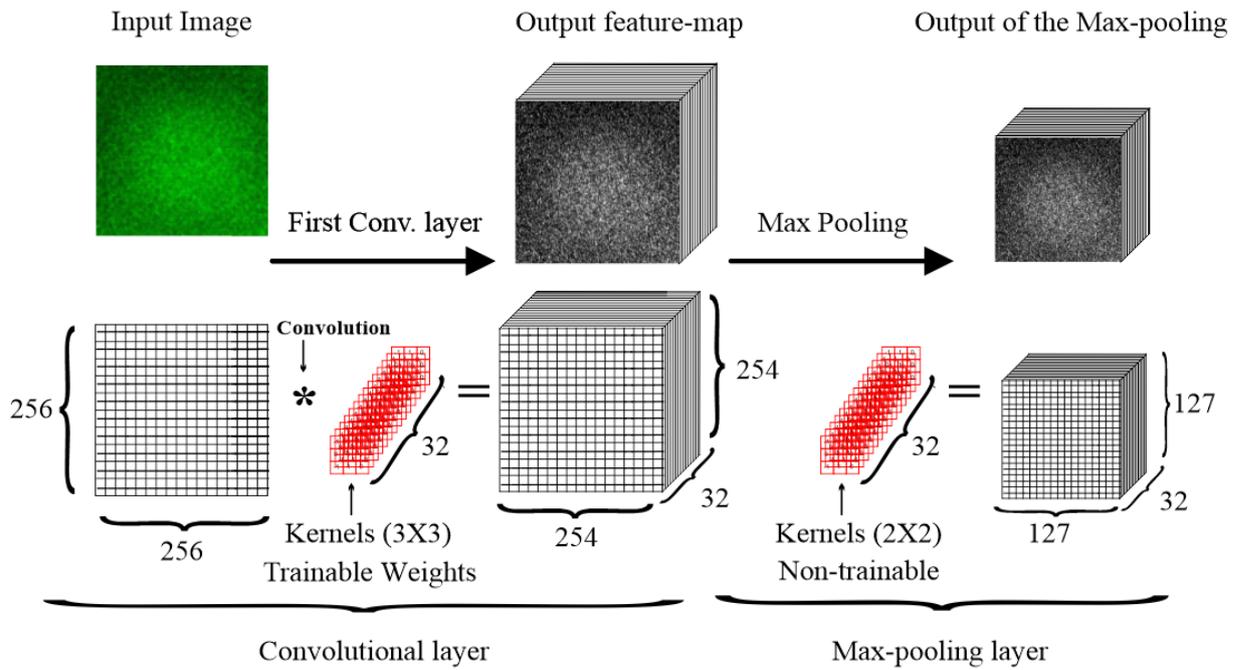

Figure 3.13 The first convolutional and max pooling layers.

The second convolutional layer has 64 filters with a kernel size of 3x3 and a ReLU activation function, followed by another max-pooling layer. As shown in figure 3.11 the size of the output feature maps from the first max pooling layer is 127 X 127, accordingly the size of the output feature maps from the second convolutional layer is 125 X 125 X 64, as result to the corresponding parameters to the table 3.1. The second max pooling layer likewise the previous max pooling layer. However, its parameters also depicted in the table 3.2.

Furthermore, the third convolutional layer has 128 filters with a kernel size of 3x3 and a ReLU activation function, followed by another max-pooling layer, accordingly the size of the output feature maps from the second convolutional layer is 60X60X128, as result to the parameters used in the layer as shown in the table 3.3.

Table 3.1 The parameters used in the second convolutional layer.

| Parameter | Value |
| --- | --- |
| Stride *(s)* | 1 |
| Padding *(p)* | 0 |
| Kernel size *(f)* | 3X3 |
| Filters | 64 |

Table 3.2 The properties of the second max pooling layer.

| Parameter | Value |
|---|---|
| Stride (s) | 2 |
| Padding (p) | 0 |
| Kernel size (f) | 2X2 |
| Output size | 62X62X64 |

Table 3.3 The properties of the third convolutional and max pooling layers.

| Parameter | Convolution | Max pooling |
|---|---|---|
| Stride (s) | 1 | 2 |
| Padding (p) | 0 | 0 |
| Kernel size (f) | 3X3 | 2X2 |
| Output size | 60X60X128 | 30X30X128 |

Table 3.4 The properties of the fourth convolutional and max pooling layers.

| Parameter | Convolution | Max pooling |
|---|---|---|
| Stride (s) | 1 | 2 |
| Padding (p) | 0 | 0 |
| Kernel size (f) | 3X3 | 2X2 |
| Output size | 28X28X128 | 14X14X128 |

The fourth convolutional layer has 128 filters with a kernel size of 3x3 and a ReLU activation function, which minimizes the size of the input feature maps from 30 to 28 in 128 new features, then followed by another max-pooling layer where the size of the feature maps reduced to 14 from 28 in the same number of features (128). The parameters used in the fourth convolutional and max pooling layers are summarised in table 3.4.

Since the classification part in the proposed model involves utilizing a network of fully connected dense layers of neurons, this artificial neural network (ANN) takes a 1D flattened input, so it is important to flatten the extracted features from the convolutional layers before making any classifications on the input image. Consequentially, the output from the last convolutional layer is connected to two dense layers by a flattening layer.

Another conversion method could be done by a Global Average Pooling (GAP) layer which involves outputting only one number from each feature by taking the average of the entire feature, this means that the output from the global average layer in case of an input feature maps of 14 X 14 X 128 dimensions would be 128, it seems like a big destroying of the output feature maps but it may performers better than the traditional flattening, especially if the size of feature maps is fairly small, Figure 3.14 shows GAP the process.

There are two methods of converting the extracted features to a 1D vector, the first conversion technique simply stacking (Flattening) the tensors to a 1D vector as shown in the figure 3.15, so if the dimensions of the output feature maps from the last convolutional layers are 14 X 14 X 128 as in the proposed model the output flattened layer is 14 * 14 * 128 a 1D vector of 25,088 elements.

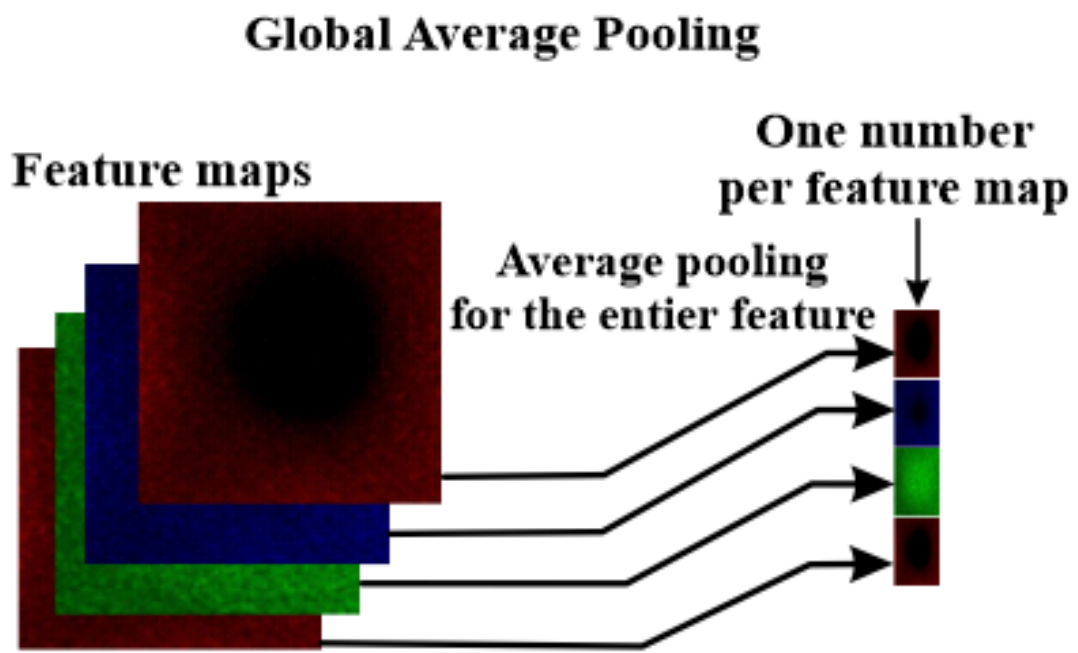

Figure 3.14 The global average pooling process.

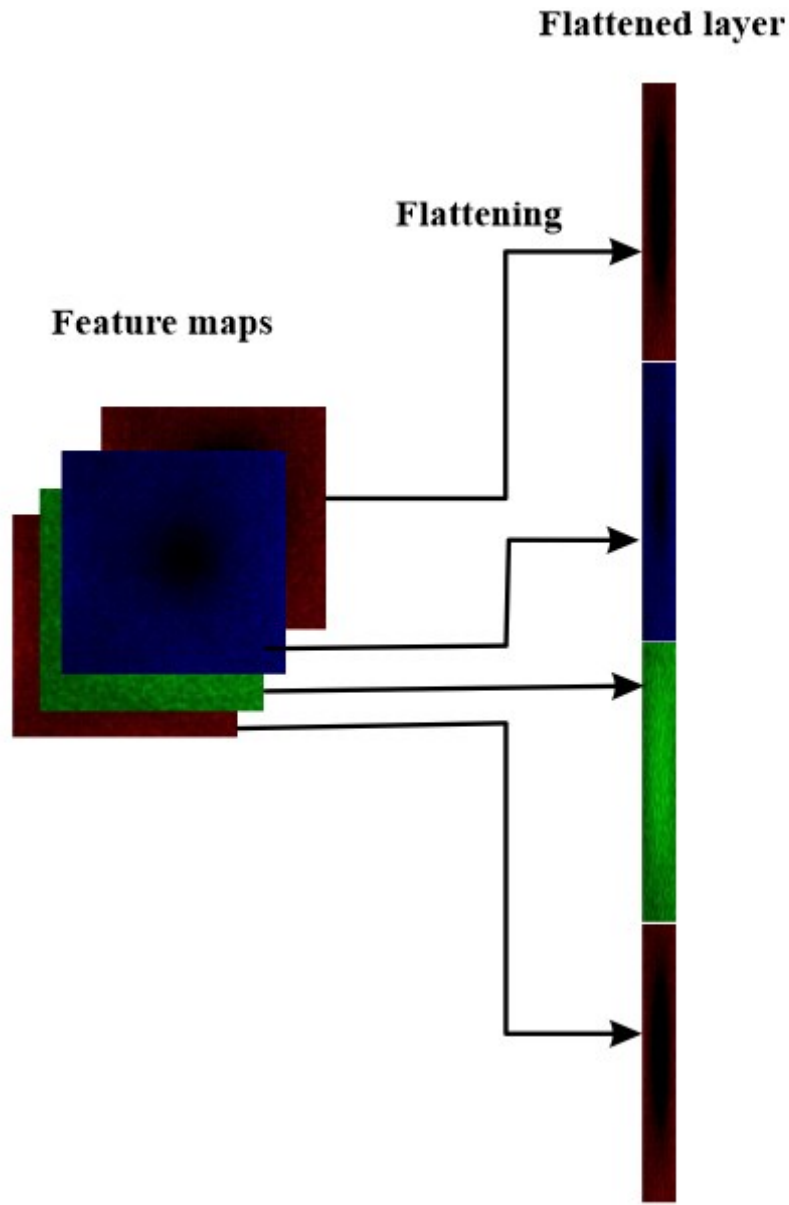

Figure 3.15 The flattening process.

After flatting the output feature maps from the last convolutional and max pooling layers, two fully connected layers are employed to classify the input image according to the flattened feature maps, the first layer consists of 512 neurons and activated by ReLU activation function, and since the second layer is the final output layer it must utilises neuron for each class, as result the final dense layer consists of 30 neurons. In order to make the final layer outputs a probability distribution for each class in the end of classification process a SoftMax activation function has used to activate the output layer, the output probability distribution vector is obtained by equation 3.12.

$$\sigma(z)_j = \frac{e^{z_j}}{\sum_{j=1}^{k} e^{z_j}} \; for \; j = 1, \ldots, k \qquad (3.12)$$

where $\sum_{j=1}^{30} e^{z_j}$ denotes the *j* element of the output vector, $z_j$ is the input vector, *k* is the number of classes, and *e* is the base of the natural logarithm. The softmax function maps the input vector $z_j$ to a probability distribution over the classes, where each element in the output vector represents the probability of the corresponding class label $\sigma(z)_j$. However, a further explanation for the proposed model is depicted in Figure 3.16.

**Optimizer**: In the context of the proposed CNN model, the optimizer used during training was Adam. The learning rate, which determines how quickly the model learns from the training data, was set to 0.001 as a starting point.

**Loss Function**: The categorical cross-entropy loss function was used to calculate the difference between the predicted probabilities and the actual class labels during training.

**Metrics**: The output accuracy of the model was used to monitor its performance during training.

To prevent the model from overfitting on the training images and to enhance its ability to generalize with different speckle images, an image augmentation technique was employed, wherein the input images were subjected to zooming in and out within a range of ±20%. This helped the model better generalize with materials of varying thicknesses. Finally, a batch size of 256 was used during the training process.

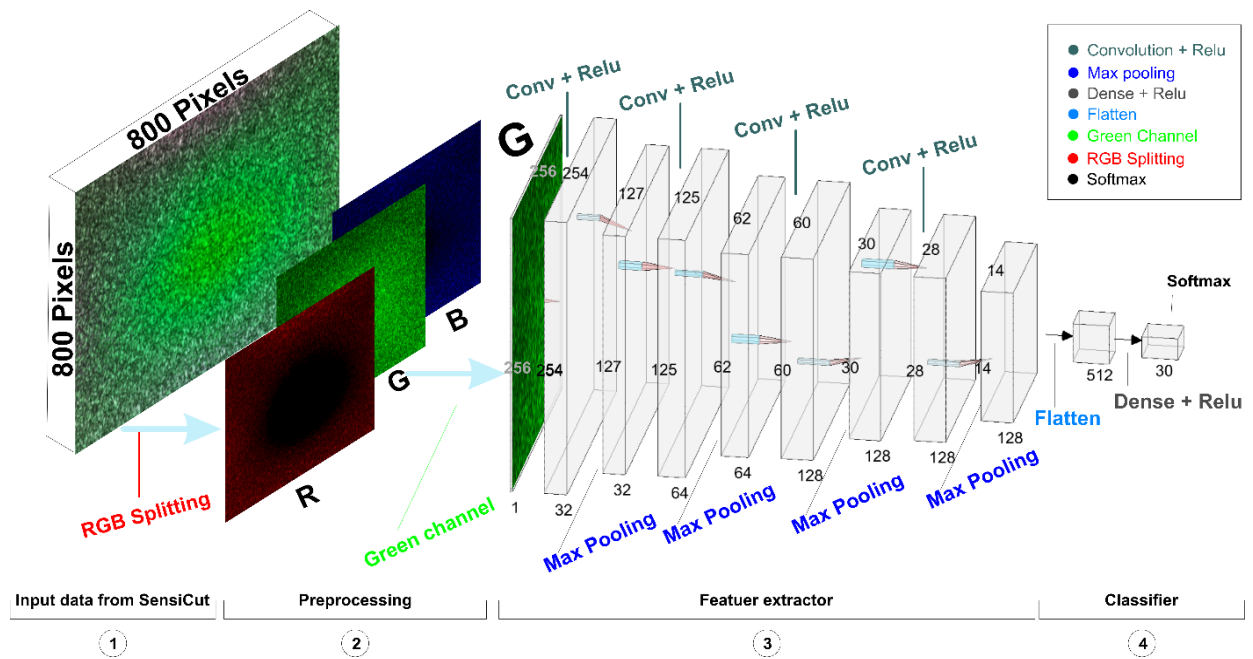

Figure 3.16. Proposed deep learning architecture.

The proposed model was compared with the model presented in the Sensicut study [25], it utilized all three RGB layers for material classification and used transfer learning from a pre-trained ResNet-50 model with around 25.6 million parameters. In contrast, the proposed model has only 13.1 million parameters. The results demonstrated that the proposed model achieved an impressive accuracy of 98.3% on a test set of 3000 images of various materials. Moreover, the proposed model is significantly faster than the Sensicut model, requiring only 3.5% of the inference time of the baseline model. These results indicate that the proposed model is efficient and has the potential for real-time material classification and hazardous material detection applications.

## 3.3 Evaluation of the Proposed Model Using Lens-less Camera

In order to validate the proposed approach of using solely the green layer, the model has trained on three different techniques the normal RGB images, the grayscale images, the grayscale images using a weighted average that emphasizes the green layer, and the green layer images, the results show clearly that the model generalized better with the green images among the others as shown in Figure 3.17.

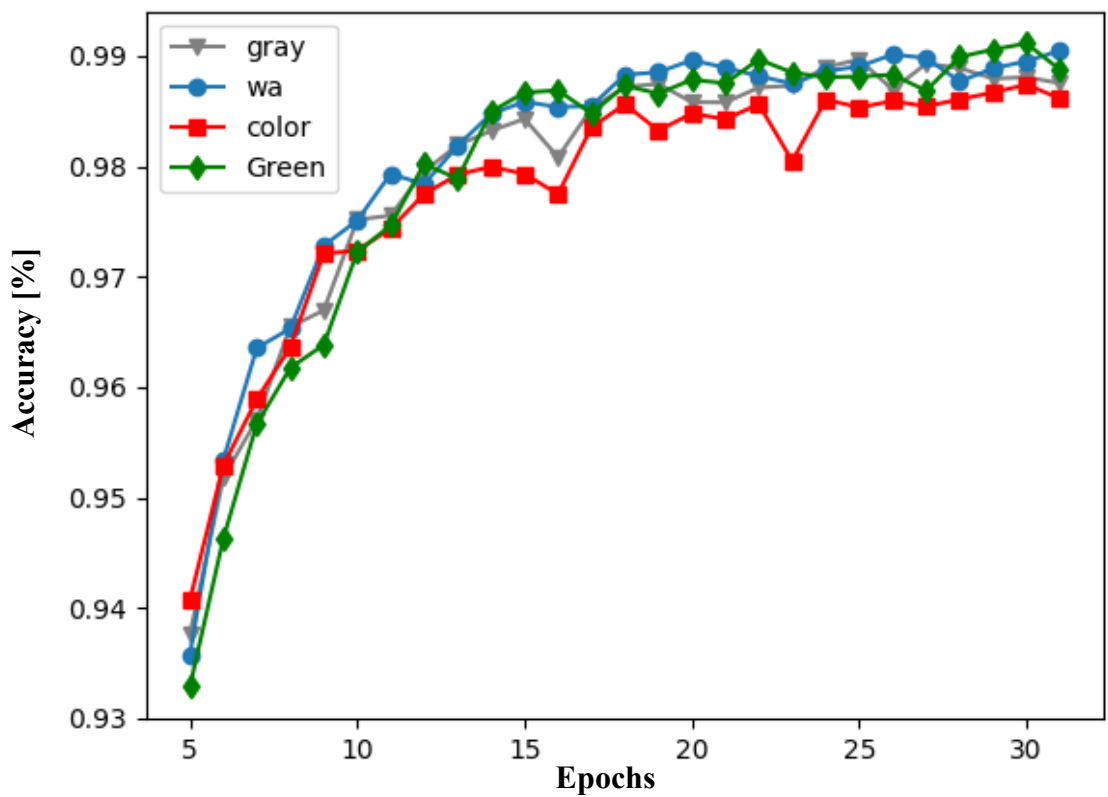

Figure 3.17 Model performance across different image layers.

Furthermore, the proposed model was compared with VGG16, ResNet50, and Xception models and result clearly shows that the proposed approach achieved the highest accuracy across the tested models as shown in Figure 3.18.

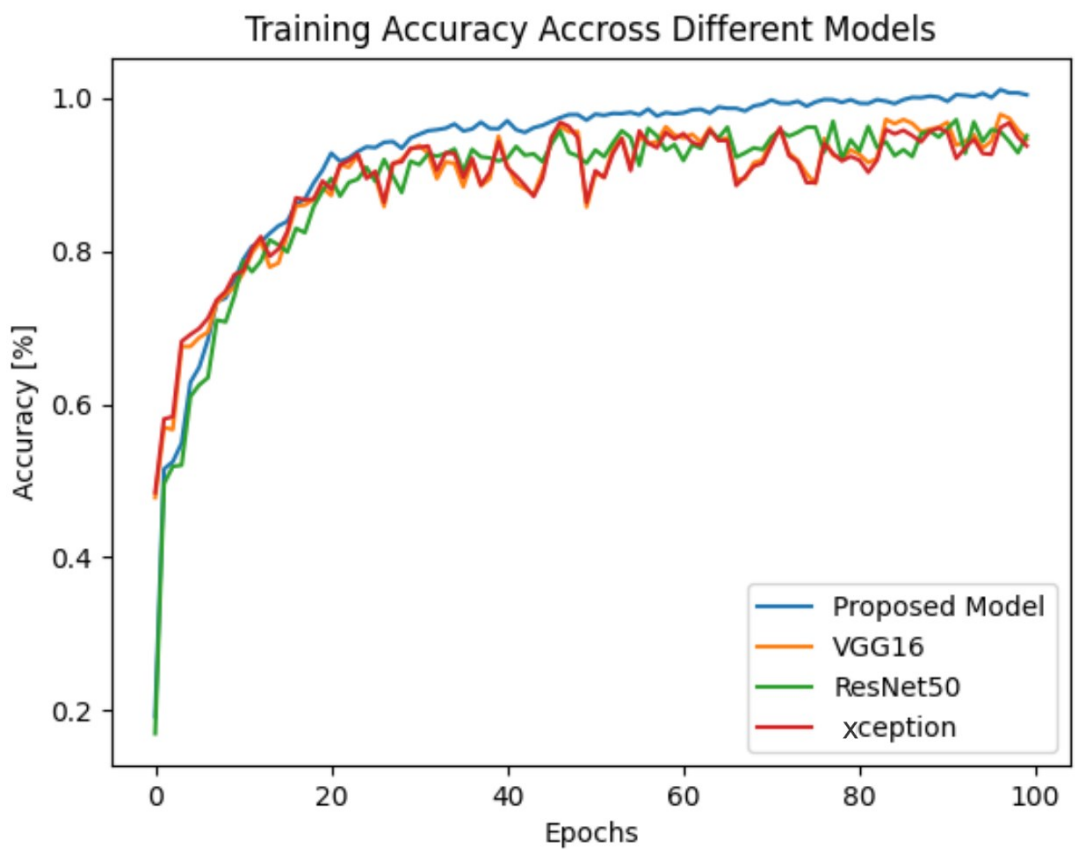

Figure 3.18 Different models comparison.

By minimizing the training and inference time required for material classification, the proposed technique can potentially enhance the efficiency of various industrial processes. Additionally, the adaptability of the technique to different types of lasers can broaden its applicability and make it a versatile tool for material characterization. This can have more significant impact in industries such as automotive, aerospace, and electronics, where rapid and accurate material identification is critical for ensuring product quality and performance.

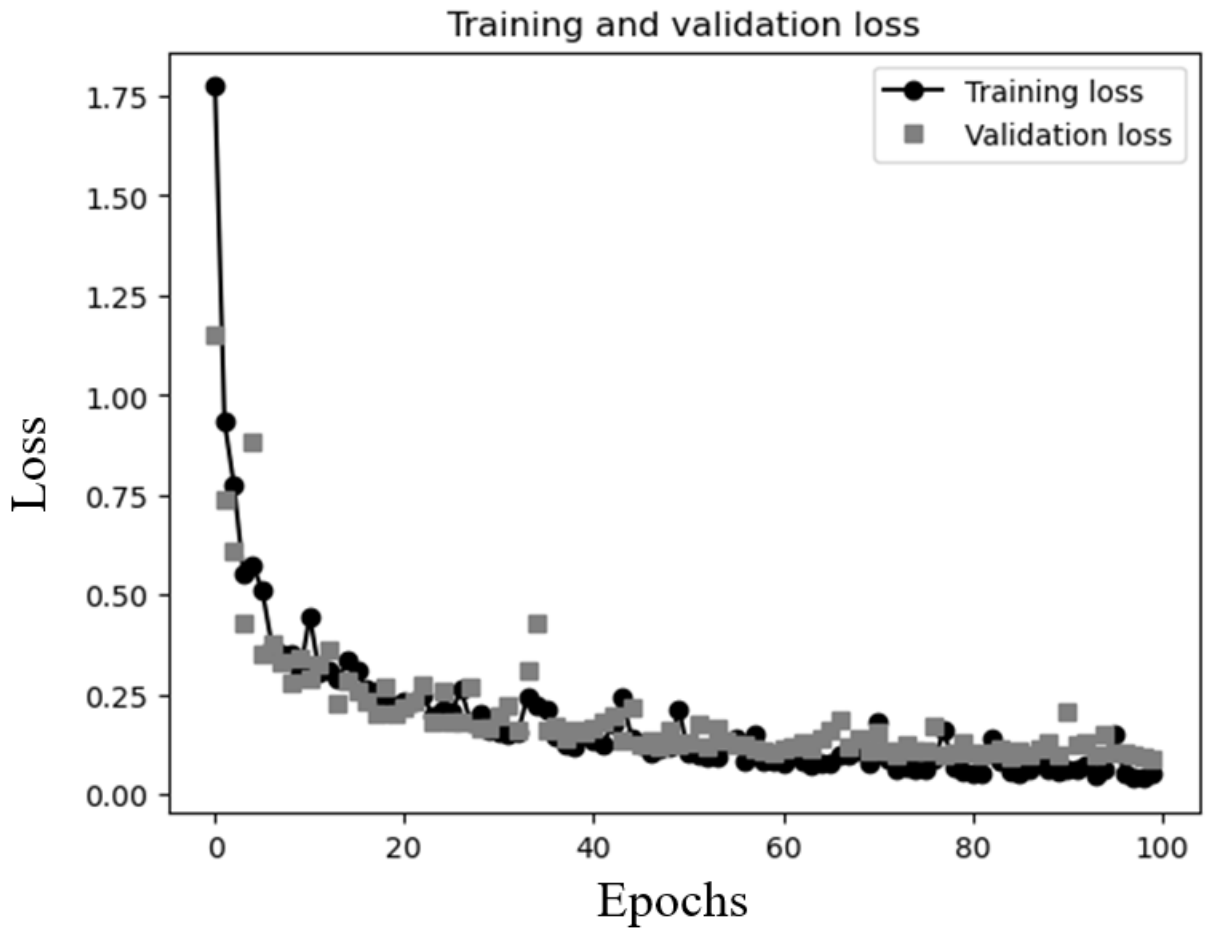

Figure 3.19 Training and validation loss.

The training and validation loss graph in Figure 3.19 illustrates the performance of the proposed model over 100 epochs. The graph shows that the model achieved stable accuracy and loss throughout the training process. The loss consistently decreased during training, while the accuracy increased rapidly in the initial epochs and then converged to a high value. The accuracy is depicted in more detail in Figure 3.20.

The proposed model's performance was evaluated by conducting experiments on a dataset consisting of 3000 images. A confusion matrix was generated to validate its accuracy in distinguishing between 30 distinct materials.

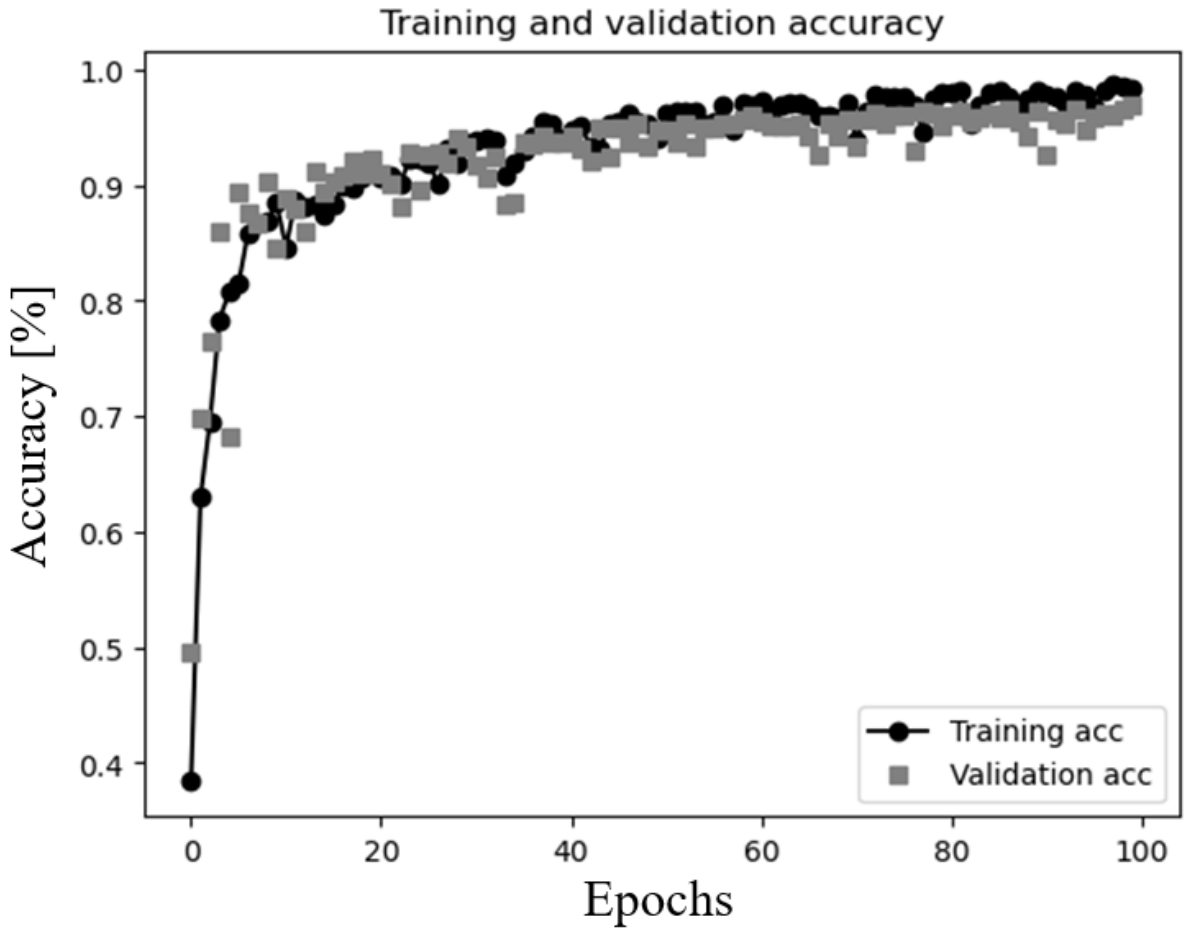

Figure 3.20 Training and validation accuracy.

Figure 3.21 illustrates the confusion matrix for the test set using the proposed model. The confusion matrix shows that the proposed model achieved a high accuracy in identifying the materials, with an overall accuracy of 97.8%. Furthermore, the model was subjected to testing its capability in recognizing hazardous materials that possess the potential to adversely impact the environment. Acrylonitrile butadiene styrene (ABS), Polyvinyl chloride (PVC), Lexan, and Carbon Fibre were included in the materials selected for this purpose. Notably, ABS, PVC, Lexan, and Carbon Fibre have the potential to generate hazardous fumes and particles when exposed to laser-cutting processes. Therefore, testing the proposed model's ability to accurately identify these materials is of critical importance, as it has the potential to aid in environmental protection efforts.

| True \ Predicted | Oak | Maple | Walnut | Birch Plywood | Cork | MDF | Veneer MDF | Bamboo | laminated MDF | Acrylic | Extruded Acrylic | Delrin | PETG | Acetate | Silicone | Styrene | Foamboard | PVC | Lexan | CarbonFiber | ABS | Felt | Black Leather | Suede | Cardstock | Cardboard | Matboard | Aluminum | StainlessSteel | CarbonSteel |
|---|---|---|---|---|---|---|---|---|---|---|---|---|---|---|---|---|---|---|---|---|---|---|---|---|---|---|---|---|---|---|
| Oak | 81 | | | | | 19 | | | | | | | | | | | | | | | | | | | | | | | | |
| Maple | | 100 | | | | | | | | | | | | | | | | | | | | | | | | | | | | |
| Walnut | | | 100 | | | | | | | | | | | | | | | | | | | | | | | | | | | |
| Birch Plywood | | | | 100 | | | | | | | | | | | | | | | | | | | | | | | | | | |
| Cork | | | | | 100 | | | | | | | | | | | | | | | | | | | | | | | | | |
| MDF | 3 | | | | | 97 | | | | | | | | | | | | | | | | | | | | | | | | |
| Veneer MDF | | | | | | | 100 | | | | | | | | | | | | | | | | | | | | | | | |
| Bamboo | | | | | | | | 100 | | | | | | | | | | | | | | | | | | | | | | |
| laminated MDF | | | | | | | | | 100 | | | | | | | | | | | | | | | | | | | | | |
| Acrylic | | | | | | | | | | 100 | | | | | | | | | | | | | | | | | | | | |
| Extruded Acrylic | | | | | | | | | | | 99 | | 1 | | | | | | | | | | | | | | | | | |
| Delrin | | | | | | | | | | | | 100 | | | | | | | | | | | | | | | | | | |
| PETG | | | | | | | | | | | 1 | | 98 | | | | | 1 | | | | | | | | | | | | |
| Acetate | | | | | | | | | | | | | | 100 | | | | | | | | | | | | | | | | |
| Silicone | | | | | | | | | | | | | | | 92 | | | | | | | | 8 | | | | | | | |
| Styrene | | | | | | | | | | | | | | | | 100 | | | | | | | | | | | | | | |
| Foamboard | | | | | | | | | | | | | | | | | 91 | | | | | | | 9 | | | | | | |
| PVC | | | | | | | | | | | | 9 | | | | | | 91 | | | | | | | | | | | | |
| Lexan | | | | | | | | | | | | | | | | | | | 100 | | | | | | | | | | | |
| CarbonFiber | | | | | | | | | | | | | | | | | | | | 100 | | | | | | | | | | |
| ABS | | | | | | | | | | | | | | | | | | | | | 100 | | | | | | | | | |
| Felt | | | | | | | | | | | | | | | | | | | | | | 100 | | | | | | | | |
| Leather | | | | | | | | | | | | | | 10 | | | | | | | | | 90 | | | | | | | |
| Suede | | | | | | | | | 4 | | | | | | | | | | | | | | | 96 | | | | | | |
| Cardstock | | 2 | | | | | | | | | | | | | | | | | | | | | | | 91 | | 7 | | | |
| Cardboard | | | | | | | | | | | | | | 3 | | | | | | | | | | | | 97 | | | | |
| Matboard | | 1 | | | | | | | | | | | | | | | | | | | | | | 5 | | | 94 | | | |
| Aluminum | | | | | | | | | | | | | | | | | | | | | | | | | | | | 85 | | 15 |
| StainlessSteel | | | | | | | | | | | | | | | | | | | | | | | | | | | | | 100 | |
| CarbonSteel | | | | | | | | | | | | | | | | | | | | | | | | | | | | 8 | | 92 |

Figure 3.21 Confusion matrix for the 30 different materials.

Tests were conducted using 100 images per material, encompassing all 30 materials, including the four hazardous ones. The proposed model demonstrated high accuracy in identifying hazardous materials, highlighting its potential to contribute to environmental protection efforts.

The generated confusion matrix uncovered instances of confusion between Polyvinyl chloride or vinyl (PVC), a hazardous material, and Delrin, attributable to the similarities in their surface structures. The experiment involved samples with identical colour properties resembling a white transparent sheet, thereby leading to ambiguity in the results. To overcome this challenge in future studies, the adopted methodology incorporates diverse colours for these materials, enhancing the model's capacity to generalize and accurately discriminate the distinct speckle patterns associated with each material as show in Figure 3.22.

To further evaluate the proposed model, the model's sensitivity to identify the materials was tested using precision, which involves calculating the fraction or proportion of the number of hazardous materials that are actually predicted as positive. This is the number of truly predicted hazardous samples over the tested ones. Additionally, the recall metric was used to calculate the fraction or proportion of the number of materials that are predicted to be hazardous over the total number of truly hazardous samples in the test set. Finally, the F1 score metric was also used to evaluate the model's ability to identify the materials. The precision, recall, and F1-score metrics are calculated according to Equations 3.13, 3.14, and 3.15 respectively.

$$Precision = \frac{TP}{TP+FP}, \tag{3.13}$$

$$Recall = \frac{TP}{TP+FN}, \tag{3.14}$$

$$F1 = 2 \cdot \frac{Precision \cdot Recall}{Precision + Recall}, \tag{3.15}$$

Where      TP = Number of true positive samples

FN = Number of false negative samples

FP = Number of false positive samples

True Positive (TP) refers to the cases where the model correctly identifies a hazardous material as hazardous, while the False Negative (FN) refers to the cases where the model incorrectly identifies a hazardous material as non-hazardous, and False Positive (FP) refers to the cases where the model incorrectly identifies a non-hazardous material as hazardous.

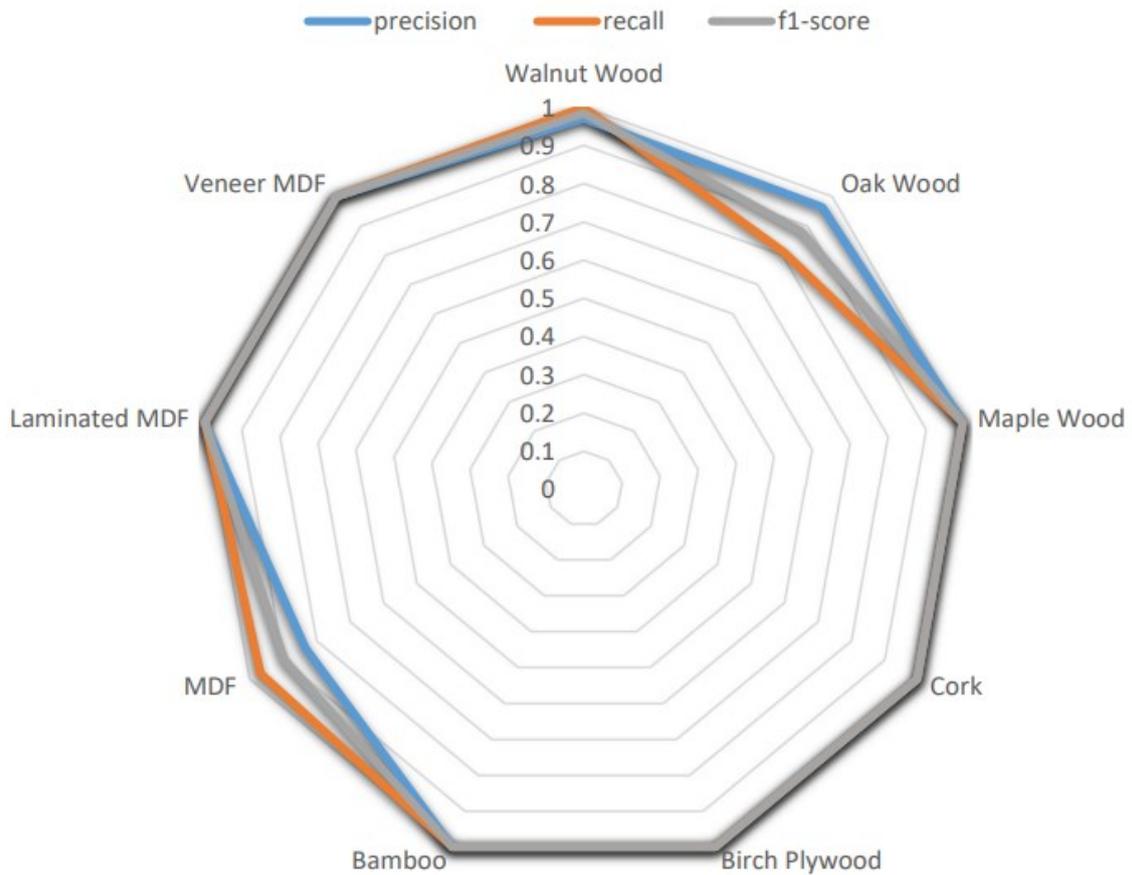

Figure 3.22 Classification report for different 9 wooden materials.

To further evaluate the proposed model, the F1 score, precision, and recall were computed for each material category. The results indicate that the proposed model was generally able to distinguish between the majority of wooden materials, such as Maple, Walnut, Birch Plywood, Cork, Veneer MDF, Bamboo, and laminated MDF. However, the randomness in the surface structure of Oak and MDF wood led to some confusion between these materials. Although the confusion between Oak and MDF was limited, it was still present when compared to the results of the base model that utilized full-color images. To address this, additional images of these two types of wood could be acquired from different positions and orientations to improve the model's ability to distinguish between them. Figure 3.20 presents the evaluation of the proposed model's performance in terms of the F1-score, precision, and recall for the wooden materials.

The precision, recall, and F1 score evaluation of the proposed model for plastic materials is depicted in Figure 3.23. The model exhibited higher classification accuracy for plastics, despite the confusion observed between silicon and felt, which was due to the black color of the samples that absorbed most of the laser rays and produced only a few speckle patterns in the images. The confusion may be alleviated by using samples with more distinguishable surface textures, thereby improving the model's performance.

Also, the evaluation of textile materials is summarized in Table 3.5. The F1 score for each category ranged from 0.90 to 1.00, indicating high precision and recall. The precision ranged from 0.91 to 1.00, and the recall ranged from 0.90 to 1.00.

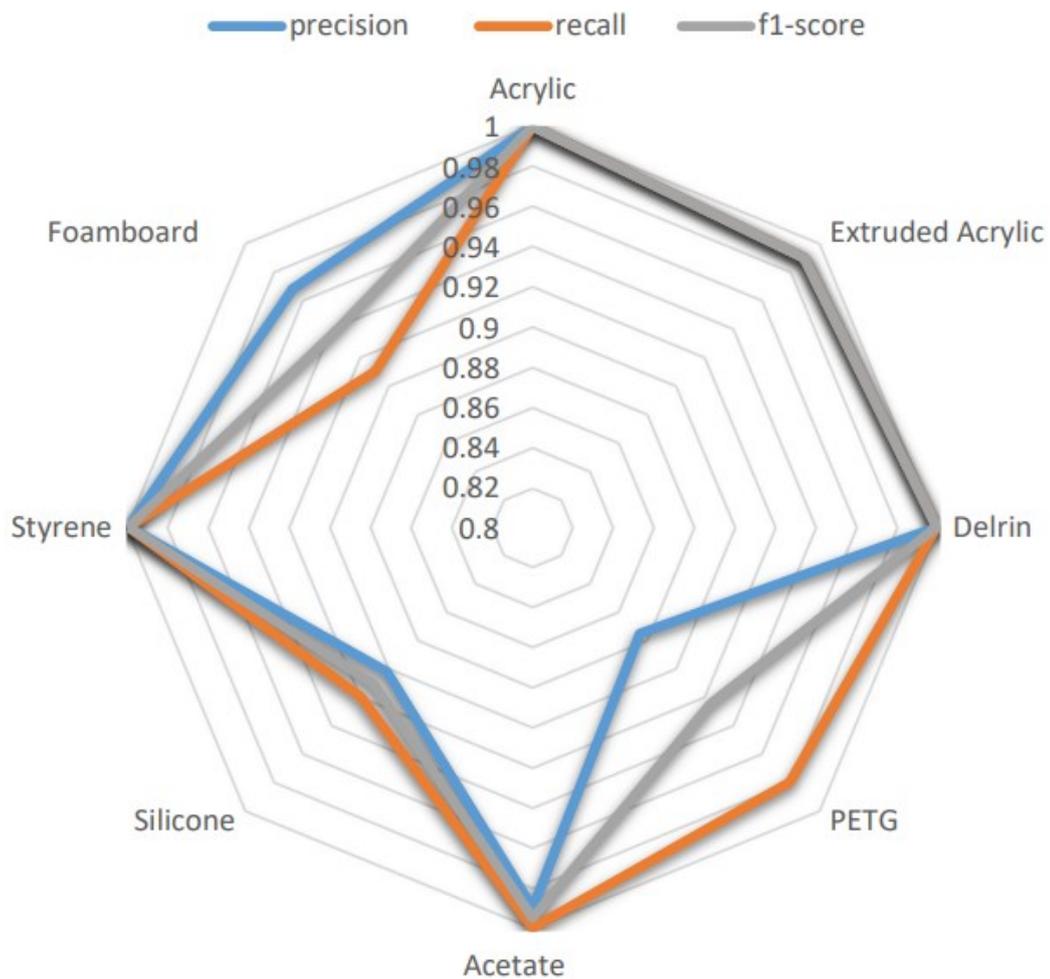

Figure 3.23 Classification report for different plastic materials.

Table 3.6 presents the precision, recall, and F1-score paper materials: Cardstock, Cardboard, and Matboard. The precision values range from 0.9151 to 0.9479, indicating a high level of accuracy in identifying the materials. The recall values range from 0.91 to 0.97, indicating the proportion of actual positives that were correctly identified by the model. The F1-score, which is a harmonic means of precision and recall, ranges from 0.9286 to 0.9417, indicating the overall good performance of the model for the paper materials.

Table 3.5 Classification report for different textile materials.

| Material | Precision | Recall | F1-Score |
|---|---|---|---|
| Felt | 1.00 | 1.00 | 1.00 |
| Leather | 0.9184 | 0.9 | 0.9091 |
| Suede | 1.00 | 0.96 | 0.9796 |

Table 3.6 Precision, recall, and F1-score for the paper materials.

| Material | Precision | Recall | F1-Score |
|---|---|---|---|
| Cardstock | 0.9479 | 0.91 | 0.9286 |
| Cardboard | 0.9151 | 0.97 | 0.9417 |
| Matboard | 0.9307 | 0.94 | 0.9353 |

Table 3.7 Precision, recall, and F1 score for metallic materials.

| Material | Precision | Recall | F1-Score |
|---|---|---|---|
| Aluminum | 0.914 | 0.85 | 0.8808 |
| Stainless-steel | 1.00 | 1.00 | 1.00 |
| Carbon Steel | 0.8598 | 0.92 | 0.8889 |

Table 3.8 Precision, recall, and F1-score for hazardous materials.

| Material | Precision | Recall | F1-Score |
|---|---|---|---|
| PVC | 0.9890 | 0.9 | 0.9424 |
| ABS | 1.00 | 1.00 | 1.00 |
| Lexan | 1.00 | 1.00 | 1.00 |
| Carbon Fibre | 1.00 | 1.00 | 1.00 |

Based on the results presented in Table 3.7, the proposed model achieved high precision, recall, and F1 score in classifying metallic materials. However, some confusion was observed between Aluminum and Carbon Steel. This confusion was also observed in the base model, which used full-colour images. These results suggest that acquiring more images of these materials from different positions and orientations may improve the model's ability to distinguish between them and therefore enhance its classification performance.

Most importantly, precision, recall, and F1-score were calculated for the hazardous materials, achieving remarkable results for all four hazardous materials. Despite the confusion between PVC and Delrin due to their similarity in surface structure and color, the proposed approach achieved 100% accuracy in classifying ABS, Lexan, and Carbon Fiber.

Overall, the proposed approach achieved high accuracy in classifying 30 different laser cutting materials, with most classes having a perfect score of 1.0. Lower scores for some materials, such as Oakwood, MDF, and Leather, may be due to physical property variation. Despite this, the approach is satisfactory, reducing material classification time and adaptable to different materials, providing a flexible solution for a safe and environmentally friendly laser cutting industry.

## 3.4  Summary of The Chapter

This chapter proposed an approach that achieved high accuracy rates in classifying a wide range of laser-cutting materials and detecting hazardous materials for safe and efficient cutting, including wood, plastics, metals, and others, with low computational time. Using one colour channel from the input image corresponding to the colour of the laser in conjunction with deep learning models, such as convolutional neural networks, significantly decreased the needed inference time compared to the lens-less camera-based laser speckle sensing with transfer learning. The approach employed in this chapter demonstrates its versatility by extending beyond laser cutting materials. It offers a practical solution for material classification across various domains, showcasing its adaptability. To develop and evaluate this approach, the model was trained on the SensiCut dataset, which leverages a lens-less camera for speckle pattern capture. One noteworthy achievement of this chapter was the development of a technique to extract the color channel corresponding to the laser used. This innovation led to a significant enhancement in classification accuracy compared to the SensiCut model's lens-less speckle sensing method. In essence, the proposed approach not only exhibits broad applicability but also surpasses the accuracy of existing techniques in the realm of lens-less based speckle sensing for material classification.

The proposed technique in this chapter can also pave the way for further research in the field of material classification based on laser speckle sensing and inspire the development of more efficient and adaptable methods for material characterization based on a lens-less camera.

# Chapter 4: USB-CAMERA-BASED CLASSIFICATION APPROACH

## 4.1 Overview and Background

The approach presented in the previous chapter exhibited a limitation that necessitates further exploration. Specifically, it relied on a green laser source for generating speckle patterns, whereas red laser pointers are more commonly employed in laser cutting machines. Thus, there is a compelling need to develop an alternative approach that leverages a red laser pointer for this purpose. Moreover, it's essential to consider cost-effectiveness in deploying speckle sensing for material classification. In this regard, this chapter introduces an innovative method using an affordable USB camera for capturing speckle patterns, eliminating the need for a more expensive lens-less camera. To facilitate the development of a deep learning model for material classification based on a USB camera, a dataset comprising 6000 images across eight primary material categories was employed. These categories encompassed materials such as Acrylic, medium-density fiberboard (MDF), plywood, foam board, ceramic, marble, leather, and Thermoplastic Polyurethane (TPU). To assess the performance of this novel approach, a series of experiments involved four acrylic sheets in varying colors, including white, gray, yellow, and clear. These particular colors were selected due to their widespread use across diverse applications. It's worth noting that the Sensicut dataset [25] did not encompass the yellow and gray color categories, further highlighting the need for this research.

## 4.2 Setup and Image Acquisition Process

The process of image acquisition involved mounting a Kisonli PC-3 HD 480P/30FPS USB camera (see appendix A) with the laser head. The lens of the camera was defocused to capture only the speckle patterns reflected from the material's surface. To maximize the scattered speckle patterns, the laser beam was also deliberately defocused. This was achieved by rotating the lens of the laser to produce a circle with a diameter of $7\ mm$, which covered a reasonable area from the surface of the material.

A camera with a focused lens is not capable of capturing speckles as it takes an image of the entire area, and therefore, the laser beam appears as a small dot in the captured image, requiring significant zoom to view the laser beam area. In contrast, when the lens is defocused or removed, the camera captures only the speckle patterns present in the entire image. To address this issue in the Sensicut dataset, a lens-less camera was used [25] (see appendix A).

This chapter proposes another different methodology for capturing speckle patterns by defocusing the camera lens instead of removing it. This approach could capture the speckle patterns by adjusting the camera focal lens outward. Fig 4.1 shows the captured scattered speckle patterns from the red laser pointer (650 $nm$ wavelength, <5 $mW$) on a white acrylic sheet to validate this approach using the defocused camera. The resulting image showed a detailed speckle pattern of the material surface.

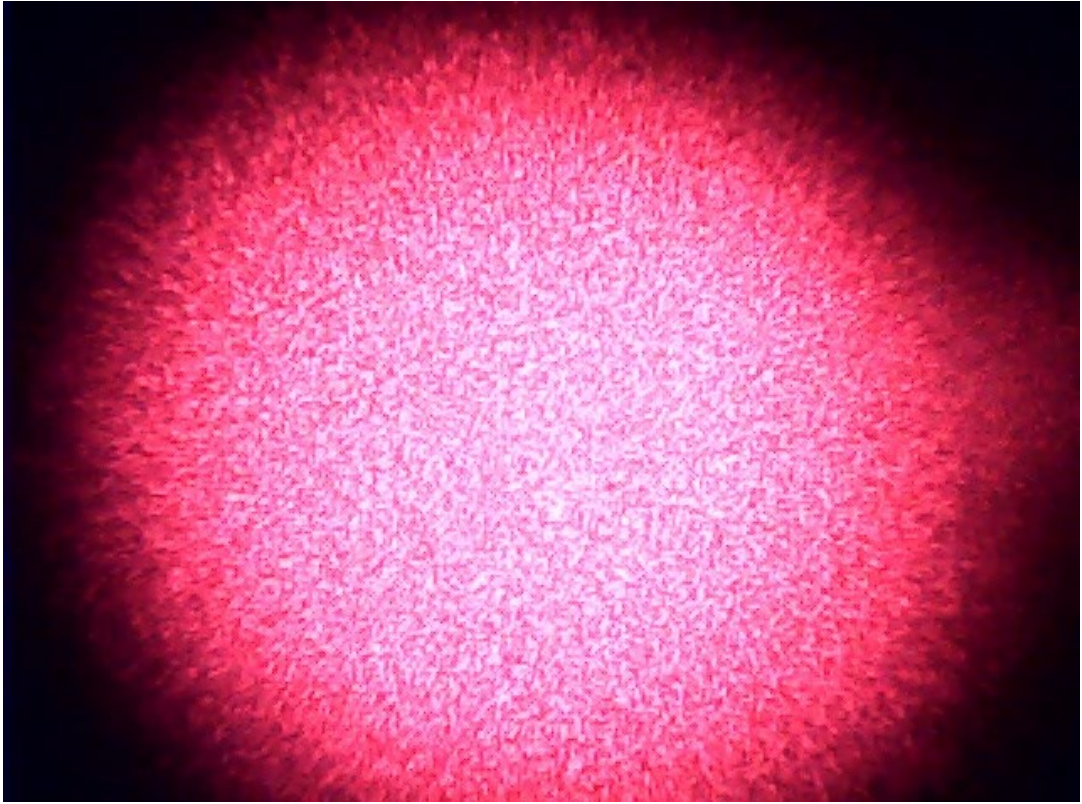

Figure 4.1 The generated Speckle patterns from a white acrylic sheet.

Then a Neje DIY 5.5 W laser cutter was programmed to move along with the camera to capture 300 speckle pattern images for each sample by navigating in both the X and Y directions across the material's surface. Additionally, the material was manually rotated to provide different orientations of its surface.

The classification of different materials is crucial, but it is also important to detect the presence of a protective layer on the material's surface as it impacts the cutting process. Furthermore, it is necessary to differentiate materials based on their colors. Therefore, this study used four acrylic samples and divided them into eight different classes based on the acrylic color and the presence or absence of the protective layer.

Table 4.1 Acrylic samples and the type of protective layer.

| Color | White PL | Clear PL | Without PL | Images |
|---|---|---|---|---|
| White | √ | - | √ | 600 |
| Yellow | - | √ | √ | 600 |
| Gray | √ | - | √ | 600 |
| Clear | √ | - | √ | 600 |

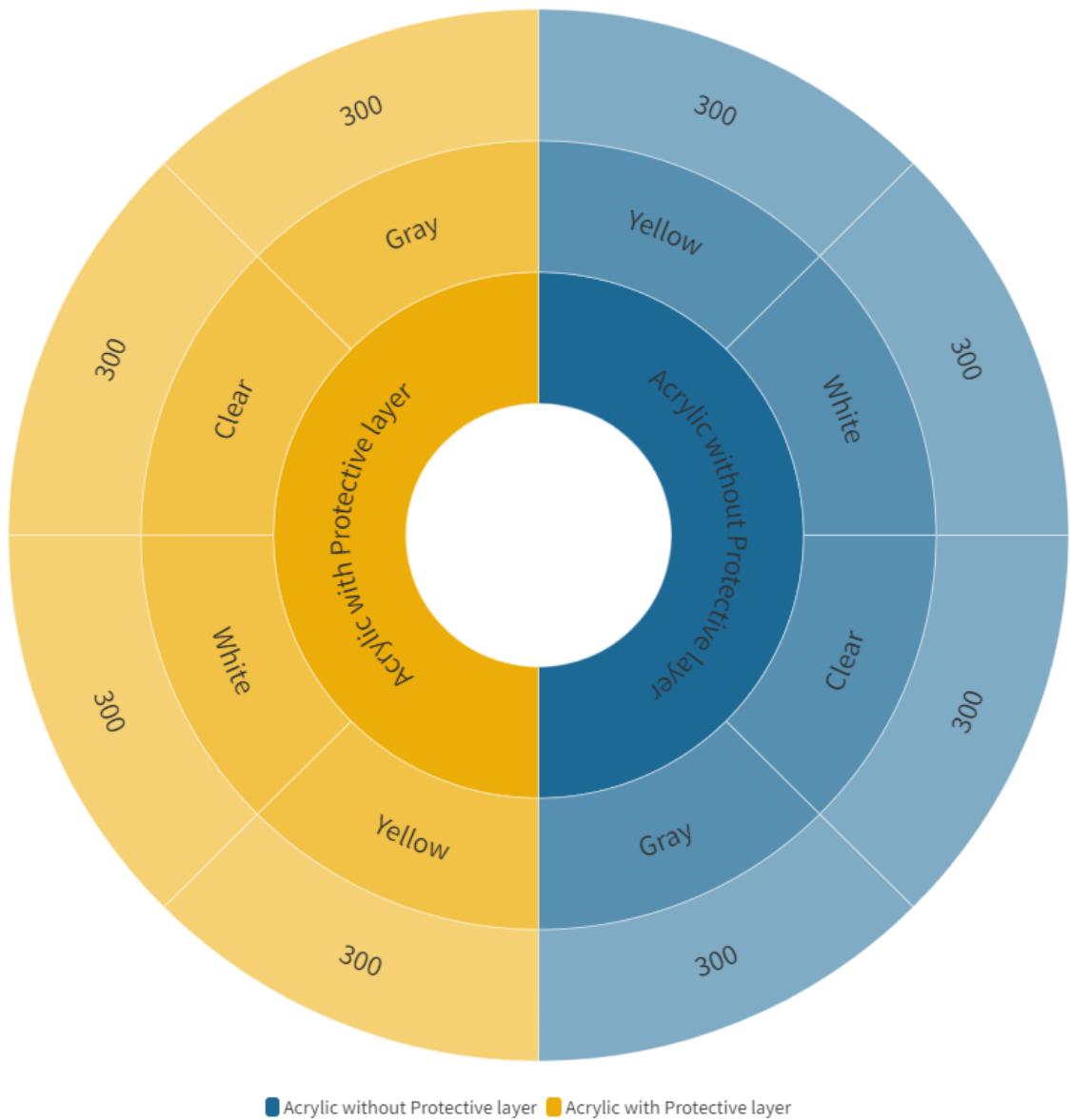

Figure 4.2 The distribution of acrylic samples with the presence of protective layer used in the dataset.

Figure 4.2 illustrates that, for each color of acrylic sheets used in this study, two distinct classes were created based on the presence of a protective layer (PL) on the sheet: one class for sheets with a white protective layer, and another class for sheets with a transparent or clear PL. A total of 300 images were generated for each category out of the eight different classes as presented also in Table 4.1. The inclusion of transparent protective layers was important for assessing the model's ability to detect sheets with a protective layer, even if the layer was transparent.

While Table 4.2 displays the number of images generated for Ceramic, Marble, and Plywood. These three materials were selected for their contrasting textures on the top and bottom surfaces. Specifically, each material has an engineered surface and a non-engineered surface. To capture this difference, two different classes were created for each surface. Eventually, Table 4.3 presents the number of images generated for TPU, Foam, Leather, and MDF, with 300 images created for each class, similar to the other classes in the dataset.

Table 4.2 The number of generated images for Ceramic, Marble, and Plywood.

| Material | Top layer | Bottom layer | Total |
|----------|-----------|--------------|-------|
| Ceramic  | 300       | 300          | 600   |
| Marble   | 300       | 300          | 600   |
| Plywood  | 300       | 300          | 600   |

Table 4.3 The number of generated images for TPU, Foam, Leather, and MDF.

| Material      | Number of Images |
|---------------|------------------|
| TPU           | 300              |
| Foam          | 300              |
| White leather | 300              |
| Black leather | 300              |
| Black MDF     | 300              |
| Wooden MDF    | 300              |

Figure 4.3 illustrates the distribution of materials beyond acrylic and the respective classes within the newly introduced dataset. It depicts the material distribution in the extended dataset, including diverse classes. The dataset comprises 20 classes, each with 300 images, totaling 6,000 images.

To facilitate model evaluation, 1,200 images (60 per class) were allocated to the validation set, while the remaining 4,800 images constituted the training set. Furthermore, 200 additional images were generated for testing the model's generalization capabilities, ensuring a robust assessment of its performance.

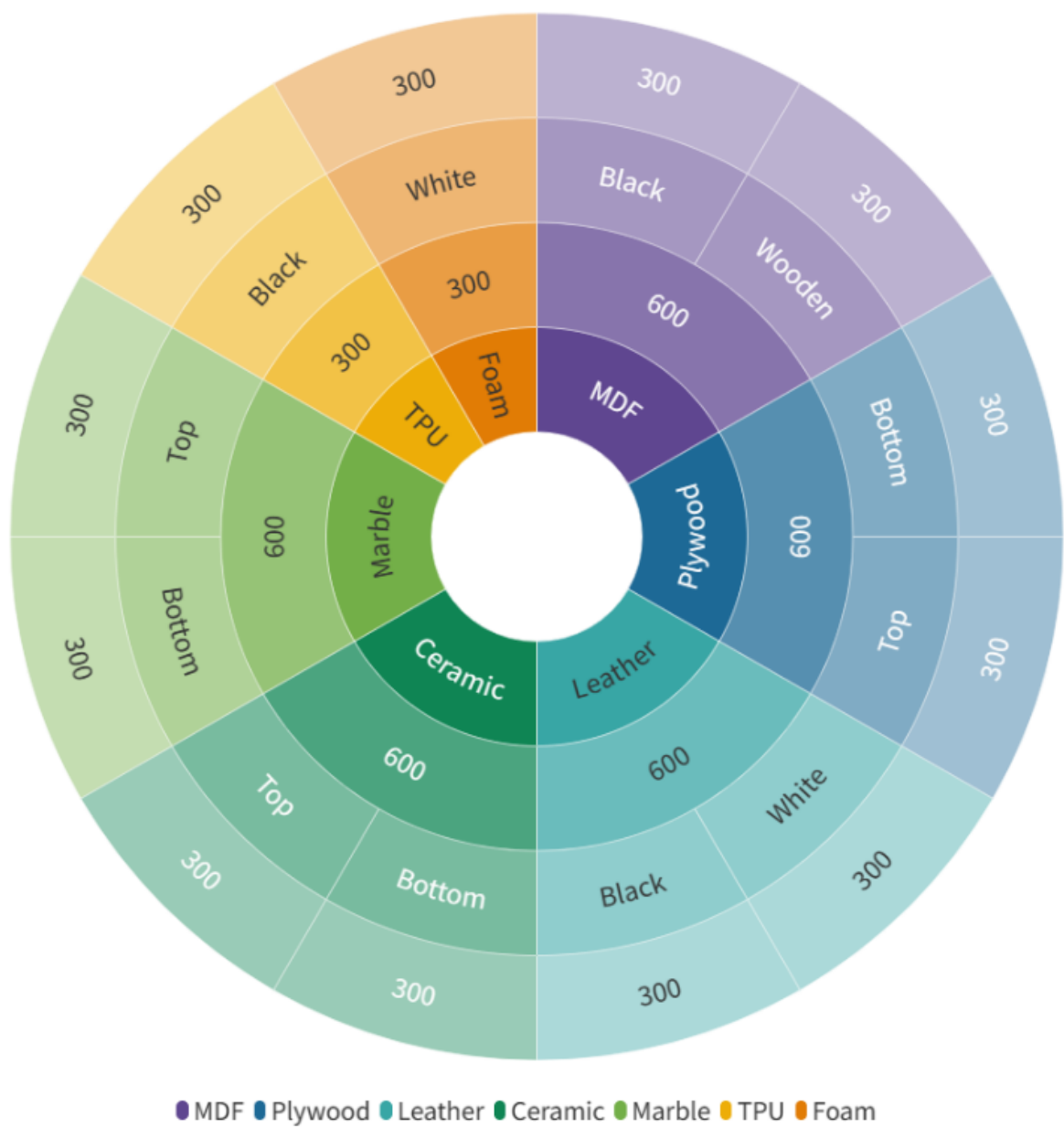

Figure 4.3 The distribution of non-acrylic samples and the type of protective layer used in the dataset.

## 4.3 The Proposed Deep Learning Model

This section presents the proposed deep-learning model for image classification using the VGG16 pre-trained model as a convolutional base to extract features from images. Then fully connected layers (Dense) were used to classify the input image as shown in Fig. 4.4.

The initial step involved pre-training the VGG16 model [38] on the ImageNet dataset [39]. The input images were preprocessed by an image data generator and resized to be entered into the model with a batch size of 64 and an input shape of (224, 224, 3). The VGG16 model architecture comprises five blocks of convolutional and pooling layers [38]. To construct the model, the VGG16 convolutional base was added as the first layer, followed by a Flatten layer, a fully connected layer with 256 neurons and a rectified linear unit (ReLU) activation function. Finally, a fully connected layer with 20 units, corresponding to the number of classes in our dataset, was added as the output layer, which was activated by a Softmax activation function.

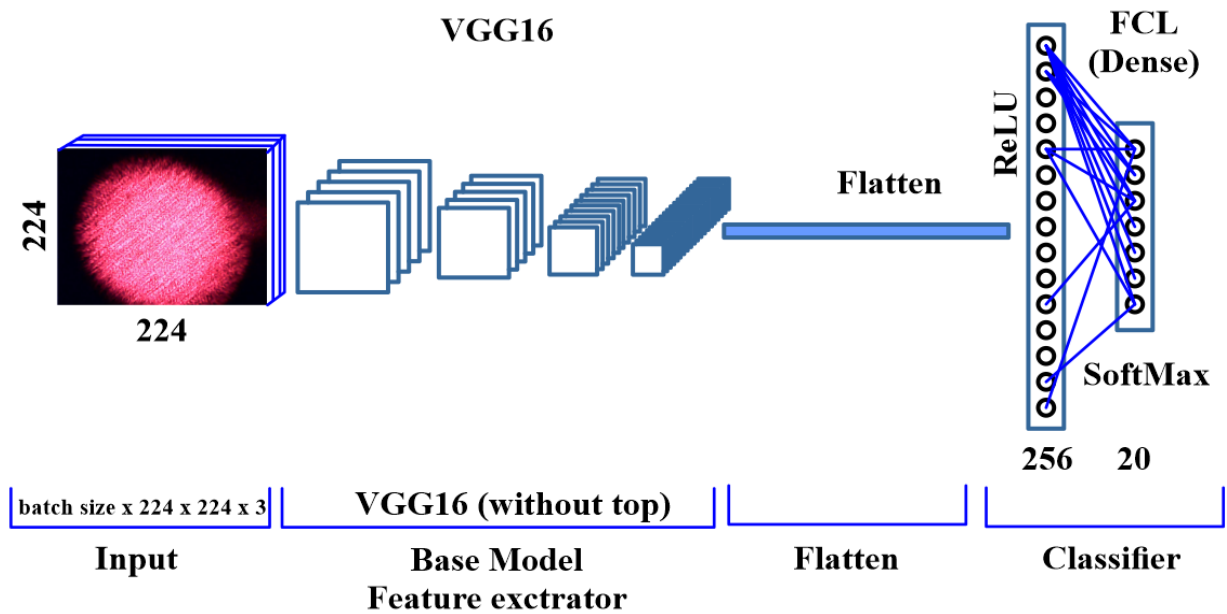

Figure 4.4 The proposed DL model architecture based on the pretrained VGG16 model.

To prevent the weights of the VGG16 convolutional base from being updated during training, they were frozen. Only the fully connected layers were trainable. Since there are 20 different classes in the dataset, the categorical cross-entropy loss function and Adam optimizer with a learning rate of 0.001 were used. The model's performance was evaluated using the accuracy metric as shown in Table 4.4.

Table 4.4 Compiling Parameters.

| Parameter | Type |
|---|---|
| Optimizer | Adam (learning_rate = 0.001) |
| Loss | Categorical cross-entropy |
| Metrics | Accuracy |

Two callbacks were employed during training. The first callback was implemented to save the best weights achieved during training, while the second callback was utilized to terminate the training process early if there was no significant improvement in the model's performance over a period of 10 epochs. The model was developed utilizing Jupyter Notebook and Google Colab.

The training dataset was augmented using various image techniques outlined in Table 4.5 to improve the model's variability and robustness. As speckle patterns are scattered and formed based on the material's surface structure, their sizes vary according to the material's thickness and the distance between the laser and the material. To avoid overfitting with the thicknesses used, a zoom range of ±20% was applied. Additionally, a rotation range of 40 was included to enable the model to generalize even if the workpiece was rotated and avoid overfitting with the used rotation angles. A width shift and height shift were also applied since the camera was mounted, and a workpiece with a large thickness could shift the location of speckle patterns in captured images. Furthermore, a horizontal flip was utilized to further increase the diversity of the training data and the model's sensitivity to even minor changes in the workpiece.

Table 4.5 Data Augmentation.

| Augmentation | Value |
|---|---|
| Zoom range | ±20% |
| Rotation range | 40 |
| Width shift range | 20% |
| Hight shift range | 20% |
| Horizontal flip | √ |
| Size | (224, 224) |

The comparison between the model performance across different pretrained models shows that the ResNet50 pretrained model performs well with the proposed dataset as shown in Table 4.6, so that the ResNet50 model has selected to extract features from to train the CNN model.

Table 4.6 Model comparison.

| Model | Accuracy | F1-score | Recall | Precision |
|---|---|---|---|---|
| ResNet50 | 0.98 | 0.97 | 0.98 | 0.97 |
| VGG16 | 0.96 | 0.95 | 0.94 | 0.96 |
| Xception | 0.96 | 0.94 | 0.96 | 0.93 |

## 4.4 Validation Results for The Proposed Classification Model

The accuracy of the model showed a significant improvement in the initial epochs of training as depicted in Fig. 4.5 The model achieved a high accuracy of 98% after 100 epochs of training, The model's ability to learn the underlying patterns in the data and generalize well is demonstrated by its high accuracy. This suggests that the model can be potentially used in practical applications to distinguish between various materials.

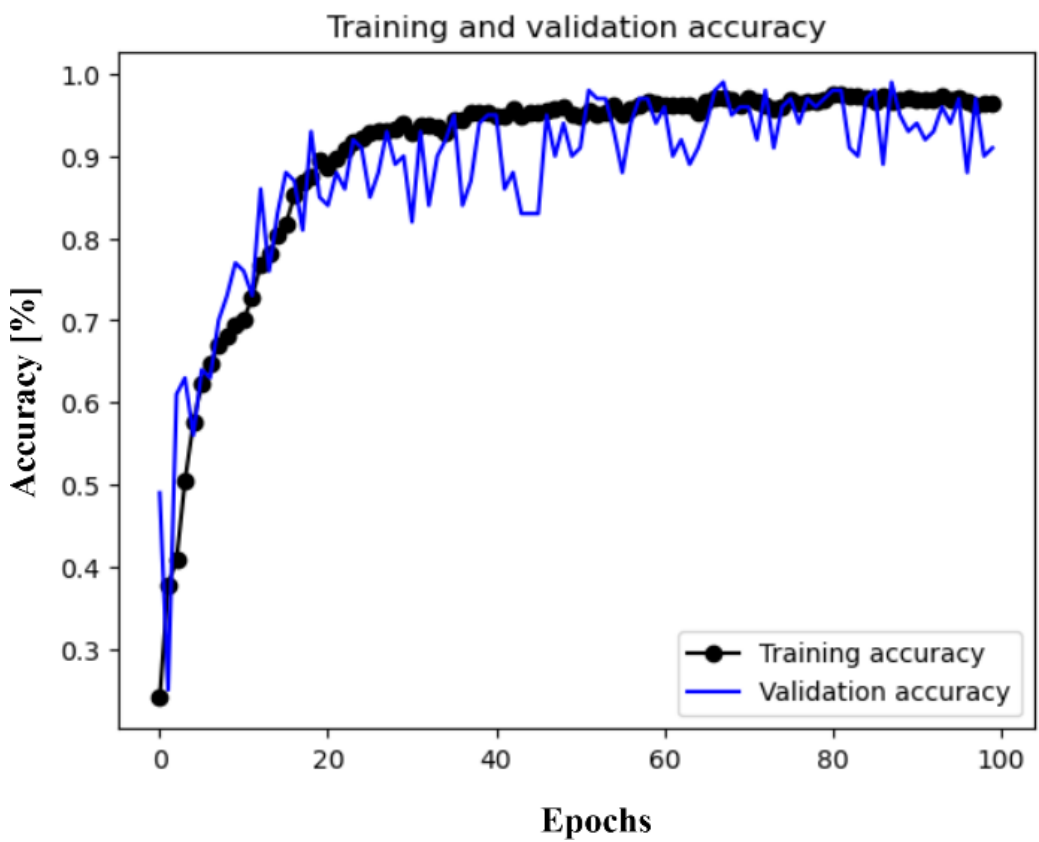

Figure 4.5 Training and validation accuracy.

The model's loss function, depicted in Fig. 4.6, showed a steep decline in the initial epochs of training, indicating that the model was quickly learning the patterns in the data. As training continued, the loss continued to decrease, but at a slower rate, until it reached a plateau at around 0.1, indicating that the model had converged to a good solution. The low loss value achieved by the model demonstrates its ability to effectively capture the important features of the different materials, and suggests that it can generalize well to new, unseen data. Overall, the combined high accuracy and low loss achieved by the model make it a promising tool for practical applications in material recognition and differentiation.

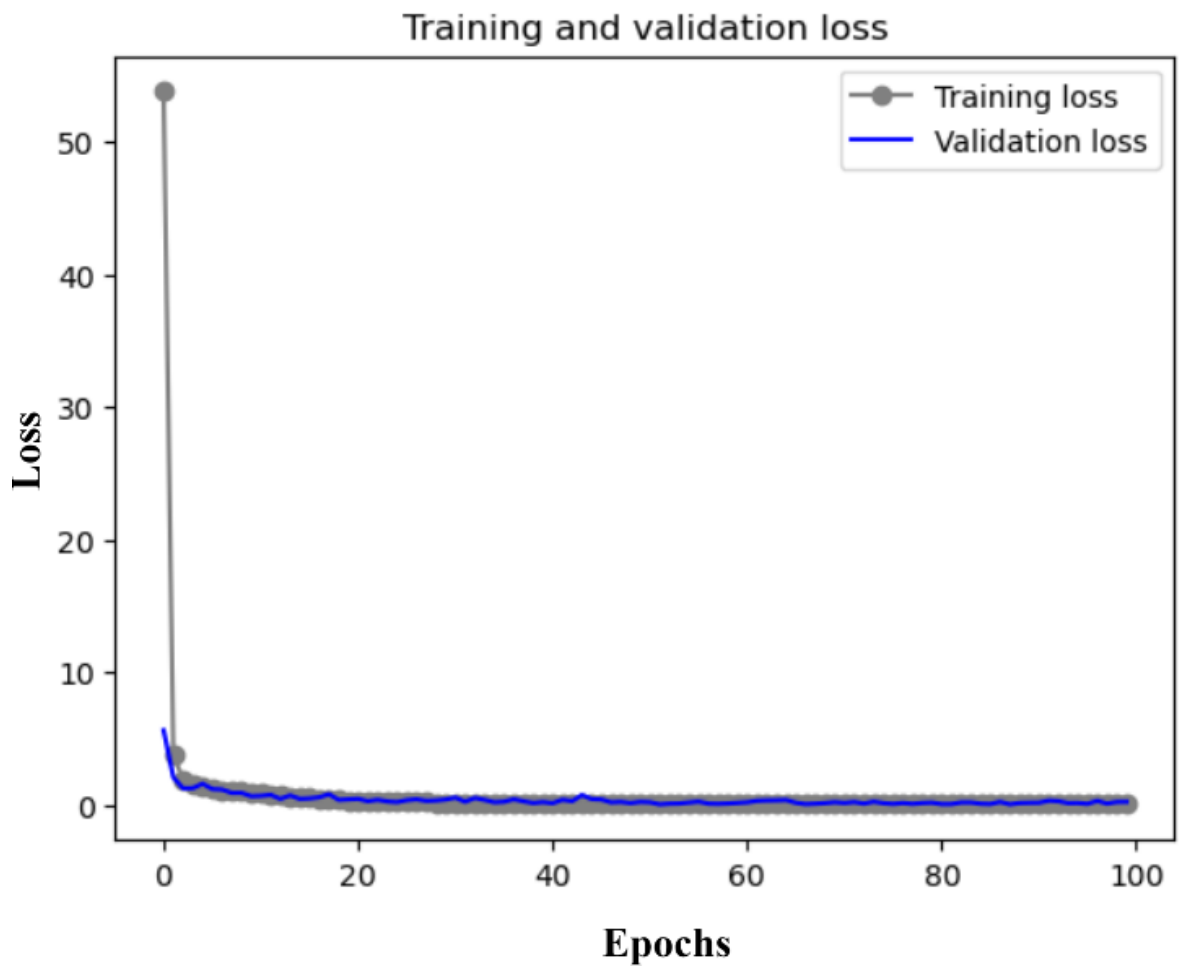

Figure 4.6 Training and validation Loss.

### 4.4.1 Classification Results

The model's accuracy and effectiveness in classifying and differentiating between 20 different surfaces were validated through a new test set of 200 images, demonstrating the model's reliability. The model's performance was then evaluated using a confusion matrix to analyze the truly tested labels and predicted ones, indicating the number of correct and incorrect classifications made by the model. This evaluation process helps to assess the accuracy of the model and its ability to generalize well to new data, ensuring its effectiveness in performing its intended task.

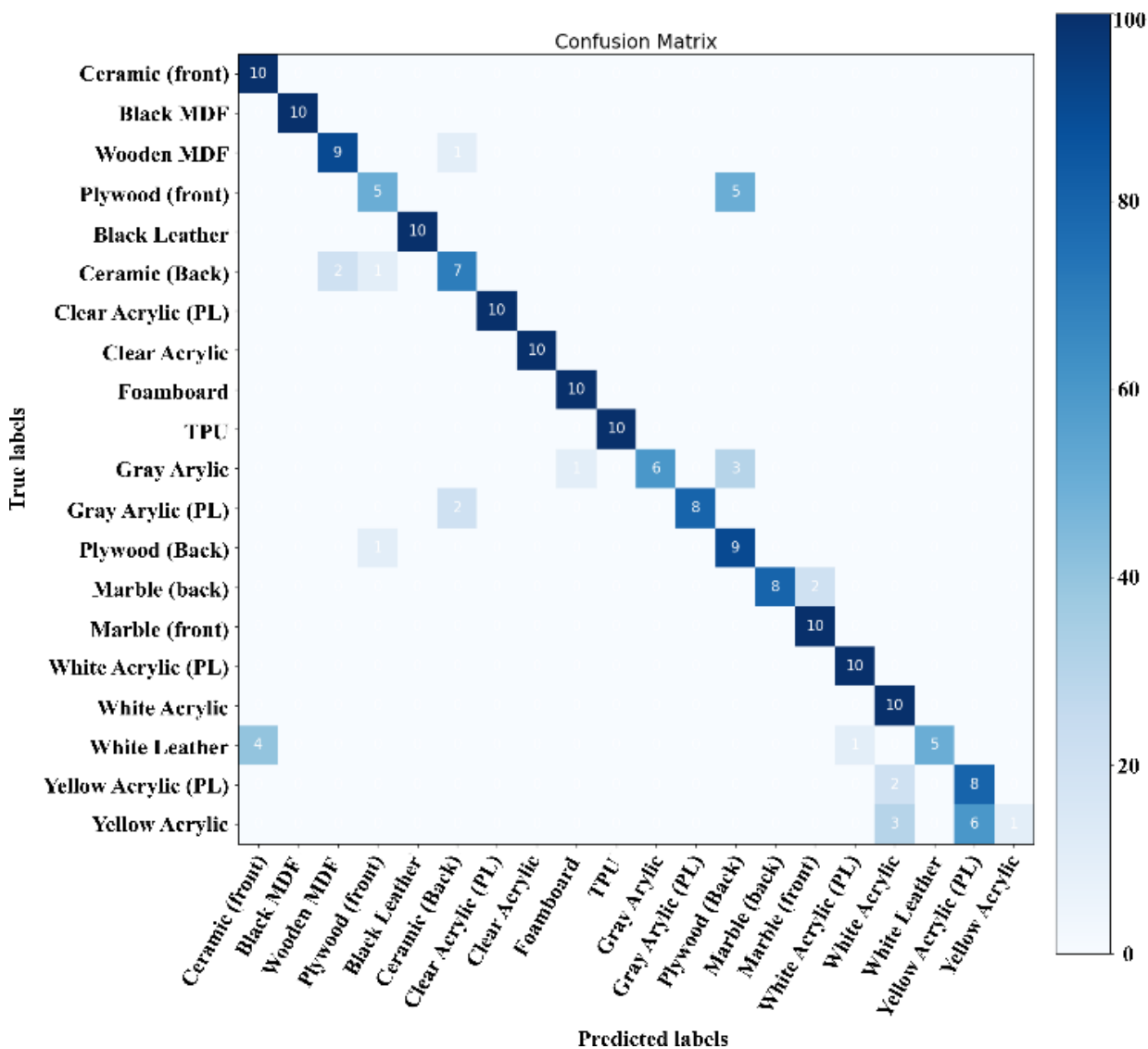

Figure 4.7 The resulting confusion matrix after testing the model.

As shown in Fig. 4.7, there is significant confusion between the yellow acrylic and the yellow acrylic with a transparent protective layer. It was found that the transparent layers do not make a significant change in the speckle patterns, which makes it challenging for the model to distinguish between materials with transparent layers.

Similarly, it is important to note that sanded plywood and its sanded top layer share the same surface structure, which makes it difficult to distinguish between them. As a result, the model did not perform well in generalizing these categories.

To address the significant confusion observed, the class of sanded plywood has been merged with the bottom layer as they share similar speckle patterns in the generated images. Additionally, they require the same laser-cut power and speed parameters when cutting using a laser cutter. Also, the yellow acrylic with the protective layer has merged into one class to form a single class for yellow acrylic with the protective layer. A new confusion matrix for the updated dataset forming 18 classes instead of 20 was generated and shown in Fig. 4.8.

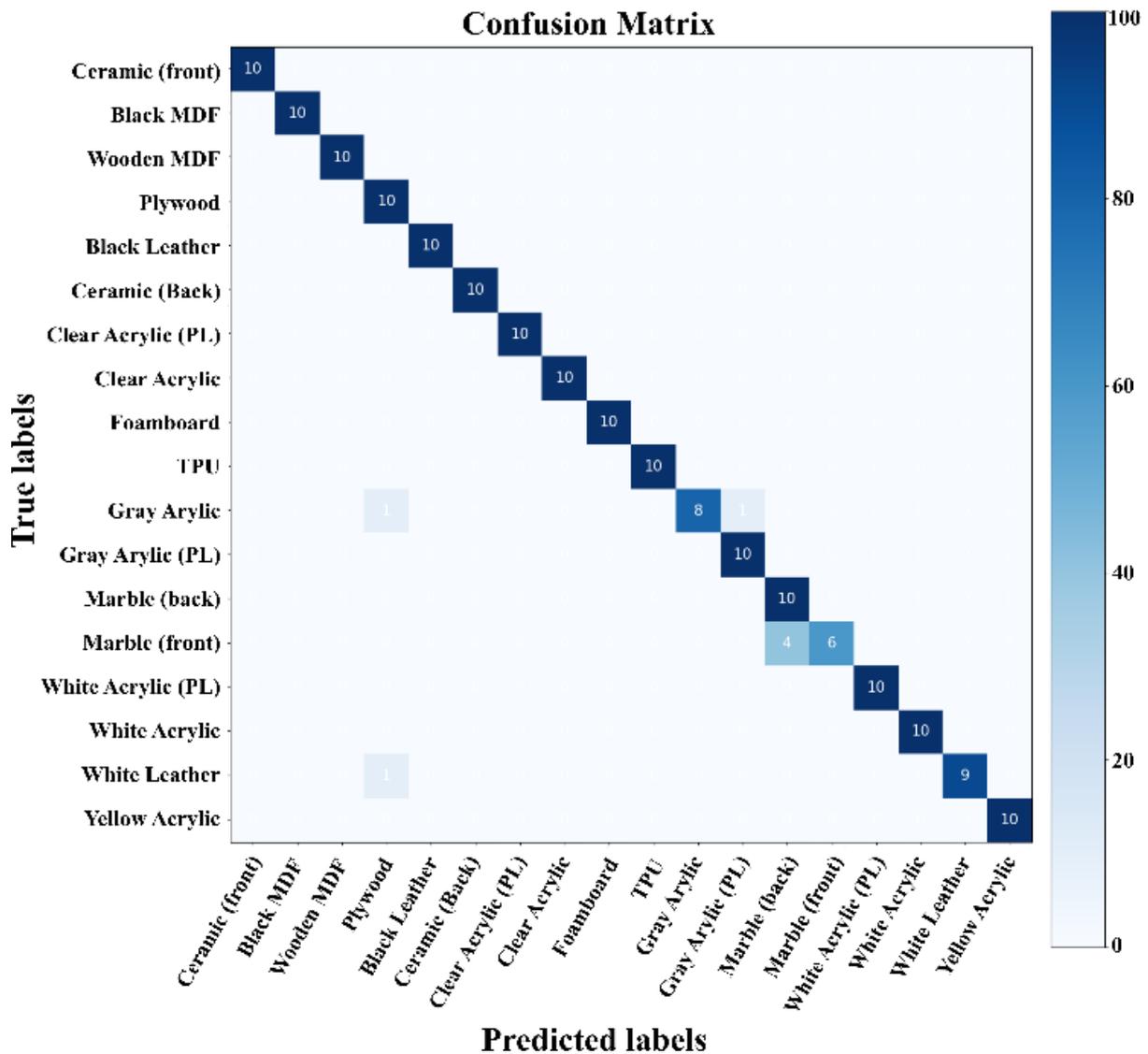

Figure 4.8 The resulting confusion matrix for the new 18 classes.

After merging the yellow with protective layer (PL) class and the sanded plywood class in the dataset, the model was retrained on the resulting 18 classes. Despite the confusion between the bottom and top surfaces of the marble, the model showed rapid improvement in accuracy, achieving a high level of accuracy in just a few epochs. This suggests that the merged dataset provided the model with better learning opportunities, enabling it to learn the underlying patterns in the data more effectively.

Furthermore, an evaluation of the model's performance in distinguishing between the 18 different surfaces was carried out through a classification report. Figure 4.9 shows the F1-score, precision, and recall for 12 different classes out of the 18 classes used in the test set, which achieved a relatively high classification accuracy of 100%.

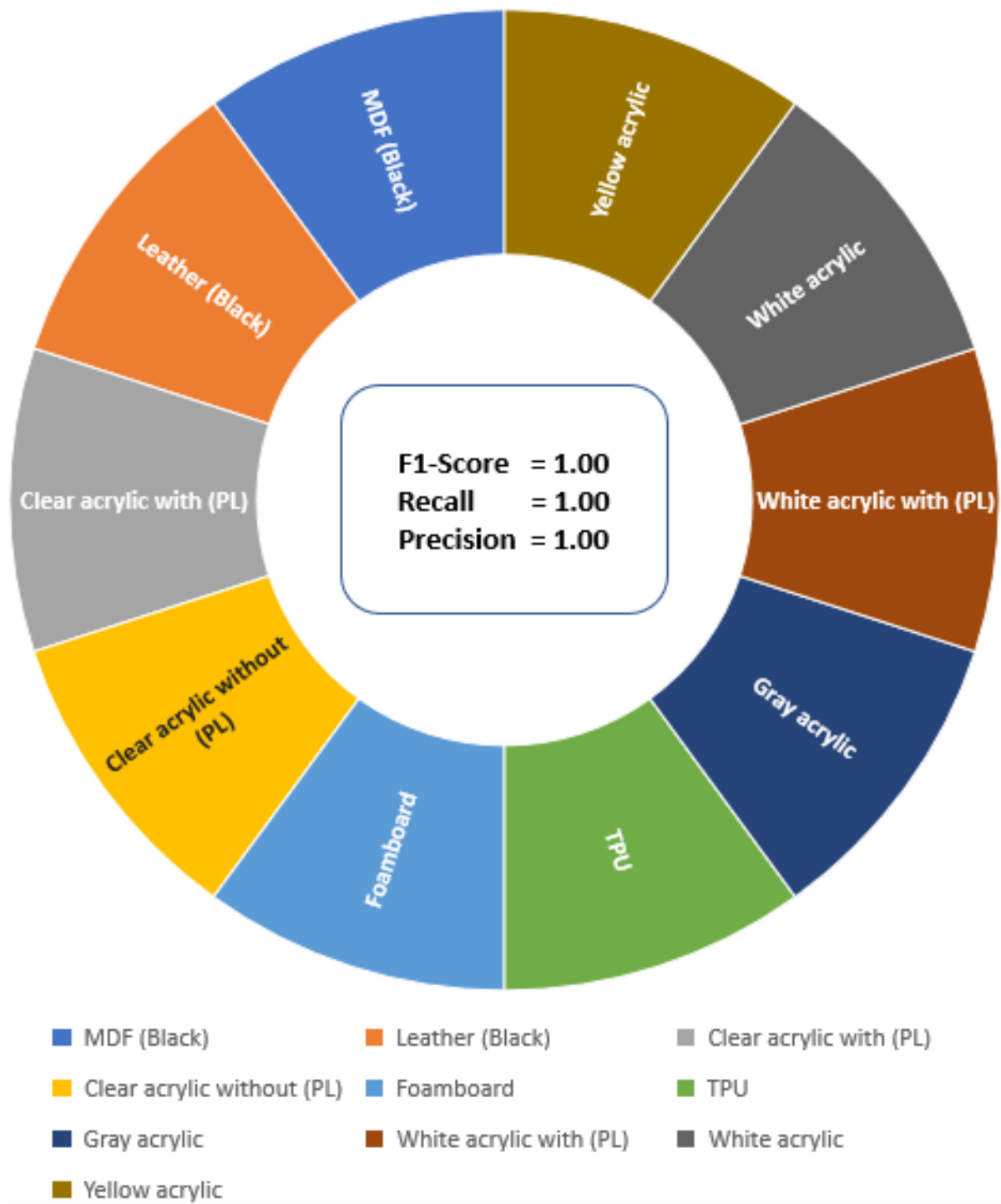

Figure 4.9 Performance Metrics for Materials with 100% test Accuracy.

Figure 4.10 demonstrates the model's ability to differentiate between acrylic sheets of various colors and between sheets with and without a protective layer. Although there was some confusion between yellow and transparent protective layers, the high F1-score, precision, and recall suggest that this method has the potential for detecting colored protective layers. The classification report also includes F1-score, precision, and recall for Ceramic and Marble samples. However, the results indicate that the model struggled to distinguish between the front and back layers of Marble samples due to their similar surface structure, leading to lower classification accuracy for these samples.

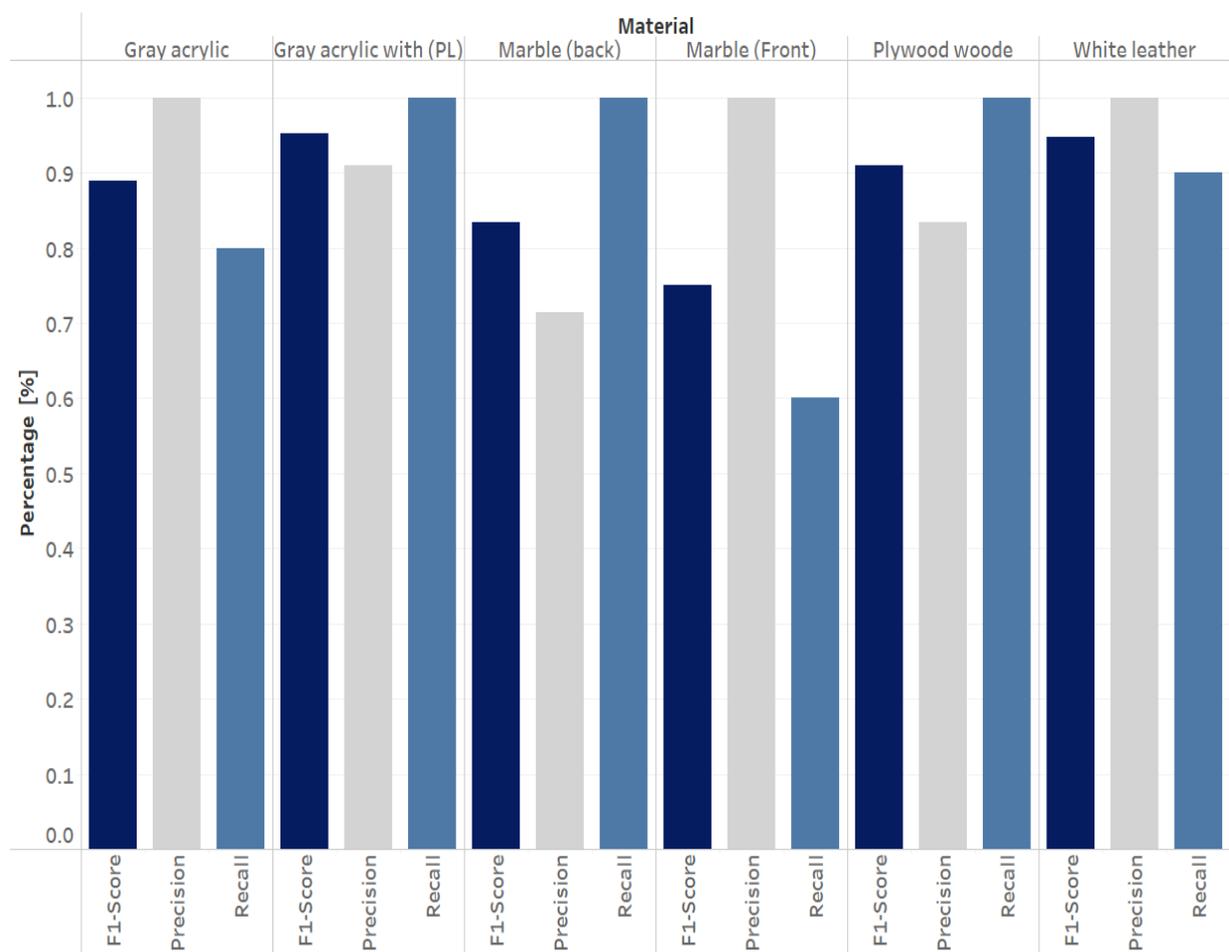

Figure 4.10 F1 score, precision, and recall for ceramic and marble.

The model was able to generalize well with TPU, foam, black leather, and black MDF, despite the confusion between plywood and other materials, especially thicker plywood samples. The training set only included plywood with a thickness of 1.5 mm, which affects the model's robustness in classifying thicker samples.

Furthermore, the proposed approach suffers from a significant drawback in that darker and low-reflective surfaces, such as black MDF and black leather, produce smaller speckle patterns, as seen in Fig. 4.11(a) and 4.11(b) respectively. In contrast, brighter surfaces, such as white leather 4.11(c), produce higher scattered speckle patterns. This issue can be addressed by utilizing a laser pointer with a higher output power and minimizing the ambient illumination. Despite this limitation, the obtained results indicate that the model can effectively distinguish between materials with similar visual characteristics, making it a promising tool for practical applications.

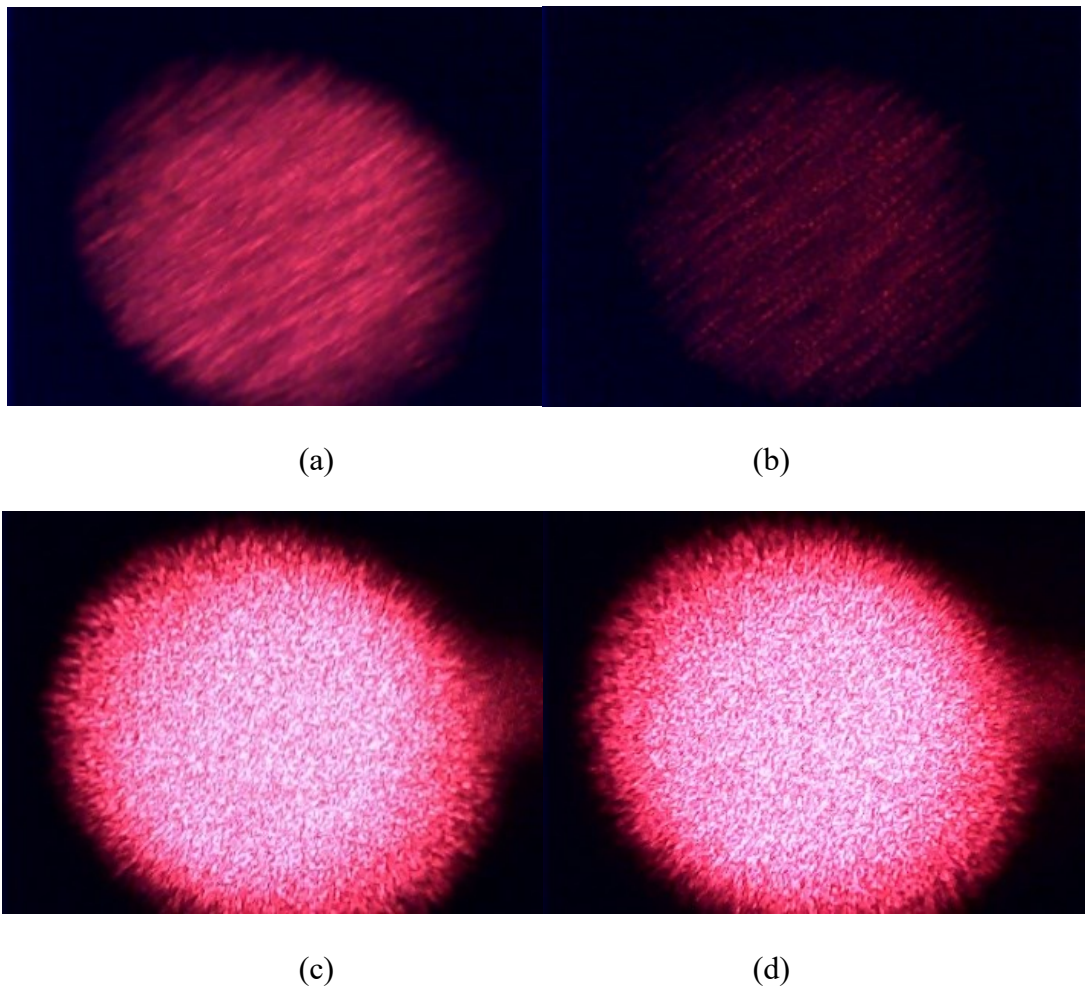

(a)  (b)

(c)  (d)

Figure 4.11 Speckle patterns for (a) black leather, (b) Black MDF, (c) white leather, and (d) the reflective front layer of ceramic.

Table 4.7 provides a comparison between the accuracy of the proposed approach and the model used in Sensicut [25]. The model used in Sensicut utilized a lens-less Raspberry Pi camera and a green laser to generate speckle patterns, which differs from the commonly used laser pointer in most laser cutters. Additionally, they utilized a larger dataset, consisting of 38,232 images, to achieve higher accuracy compared to the proposed approach.

Table 4.7 Proposed Model Comparison.

| Model | Laser pointer | Camera | Number of images | Accuracy |
|---|---|---|---|---|
| SensiCut [25] | Green (515 nm, 5 <w) | Lens-less | 38,232 | 98.01% |
| Proposed approach | Red (650 nm, 5 <w) | USB | 6000 | 98.1% |

## 4.5 Summary of The Chapter

In this chapter, an approach to speckle sensing has been developed, capitalizing on readily accessible technology and resources. The method revolves around the use of an affordable USB camera and a red laser pointer, coupled with a deep learning model, designed for the explicit purpose of classifying materials frequently employed in laser cutting processes. It also presents a newly created dataset, comprising a substantial 6000 images, spanning eight fundamental material categories. These categories encompass a wide spectrum, including Acrylic, MDF, plywood, foam, ceramic, marble, leather, and Thermoplastic Polyurethane (TPU). Notably, within each category, distinct classes were created, accounting for variations in color, the presence or absence of protective layers, surface texture, and other unique material attributes. This innovative approach signifies a departure from prior methods, where a lens-less camera was employed to capture speckle patterns generated by a green laser pointer. The limitations of this earlier technique, stemming from cost and accessibility constraints, emphasize the significance of the approach detailed in this chapter. By employing readily available and affordable components, this method extends the potential application of speckle sensing across various industries, transcending the confines of laser cutting alone.

The creation of a new dataset through the proposed approach has empowered the deep learning model introduced in this chapter to determine whether a laser-cut material has a protective layer or not, and it's even capable of distinguishing between the top and bottom surfaces of these materials. This approach has shown exciting outcomes when it comes to recognizing 20 different surface types using speckle imaging and deep learning techniques. The model proved to be highly accurate in identifying various materials like different types of wood, acrylic sheets, leather, ceramic, marble, and TPU. Nevertheless, there were instances where the model got a bit confused, especially with materials that had transparent protective layers or similar surface properties.

# Chapter 5: ENHANCING ENERGY CONSUMPTION IN LASER CUTTING

## 5.1 Overview and Background

This chapter focuses on enhancing the energy efficiency and safety aspects of laser cutting machines. During the cutting process, the interaction of the laser beam with the workpiece can generate potentially harmful smoke and fumes. To address this issue, this study developed a deep learning model based on a convolutional neural network (CNN) to automatically detect and identify smoke and fumes generated during cutting. It created a dataset of 2500 images captured during laser cutting on different laser cutting materials, in contrast with the previous chapters the camera used in this approach is a focused USB camera, this enabled the ability to capture a wide range of smoke and fume patterns. The smoke detection approach leverages a pre-trained ResNet50 model to extract valuable image features, which it meticulously fine-tuned using the proposed dataset. Through extensive cross-validation and model selection, ResNet50 consistently outperformed other models like VGG16 and Xception. Therefore, the study adopted ResNet50 as the primary convolutional base for the smoke detection system, ensuring its reliability and accuracy in identifying smoke presence.

This chapter also explores the practical implementation of previously proposed approaches within laser cutting machines to enhance their overall energy efficiency. It aims to reduce energy consumption and ensure safety by continuously monitoring potential smoke hazards during cutting operations.

The chapter starts with the smoke detection approach then it demonstrates the proposed strategies for a practical implementation of previously proposed approaches by delving into the intricate relationship between energy consumption, material characteristics, and operational strategies in laser cutting operations. It employs innovative methodologies to enhance efficiency and sustainability while prioritizing safety through smoke detection.

### 5.1.1 Setup and Image Acquisition Process

This section presents a comprehensive dataset for smoke that is generated during laser cutting. Since the smoke and fumes produced during the cutting process are different from those in publicly available datasets, a dataset of 2500 images was generated using a USB camera with a Macro $CO_2$ laser cutting machine to train a deep learning model to detect the smoke. The dataset was split into two sets: a training set and a validation set. The training set consisted of 2000 images, while the validation set had 500 images. The training set was used to train the model, and the validation set was used to evaluate its performance during the training process. Samples from the generated images in the used dataset are presented in Figure 5.1.

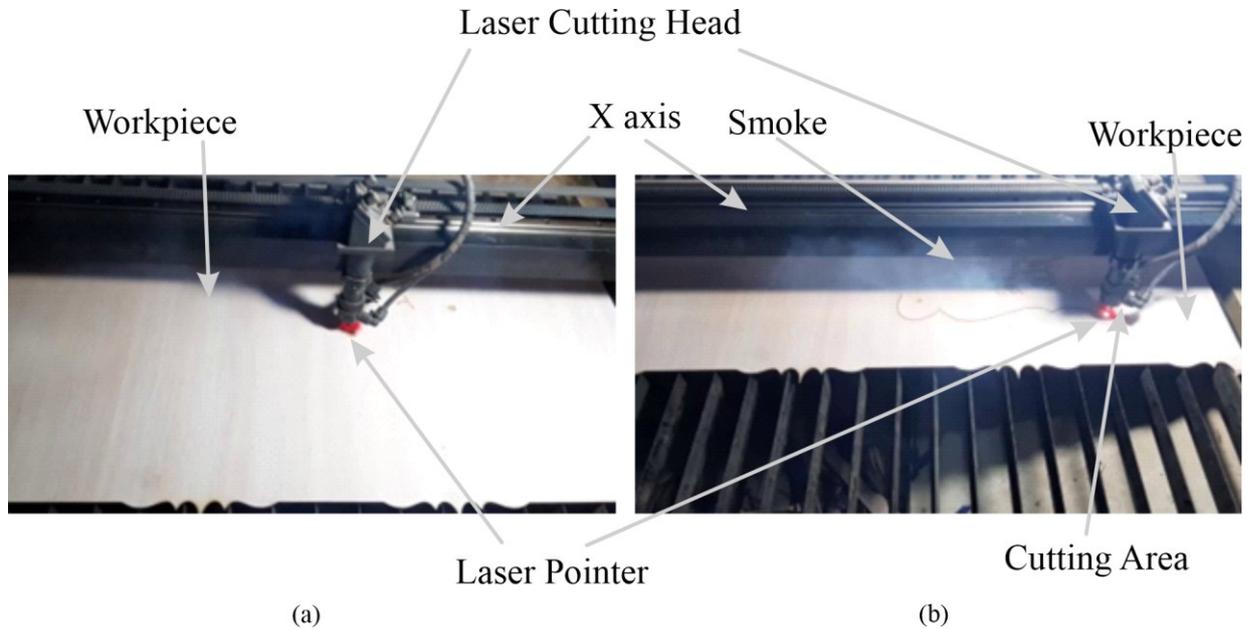

Figure 5.1 Dataset Sample Images – (a) No-Smoke and (b) Smoke Classes.

### 5.1.2 Proposed Smoke Detection Model

In this section, a deep-learning model for smoke detection is presented. The model builds upon the ResNet50 [36] pre-trained model, which serves as the convolutional base for feature extraction from images. To classify these extracted features, fully connected layers (Dense) are employed. The classification task involves sorting images into two distinct classes: one for images displaying generated smoke and another for images where smoke is absent. This results in a binary classification model, as illustrated in Figure 5.2.

The first step involves loading the pre-trained weights of the ResNet50 model, which was trained on the ImageNet dataset. The input image is resized from 1280 X 720 to 224 X 224 using the bilinear interpolation algorithm to match the input shape (batch size, 224, 224, 3). An image data generator is used to augment and generate images in 64 batches. To construct the model, the ResNet50 convolutional base is added as the first layer, followed by a Flatten layer, a fully connected layer with 256 neurons, with a rectified linear unit (ReLU) activation function. Finally, a fully connected output layer with 1 unit as the model just binarily classify, which is activated using a sigmoid activation function.

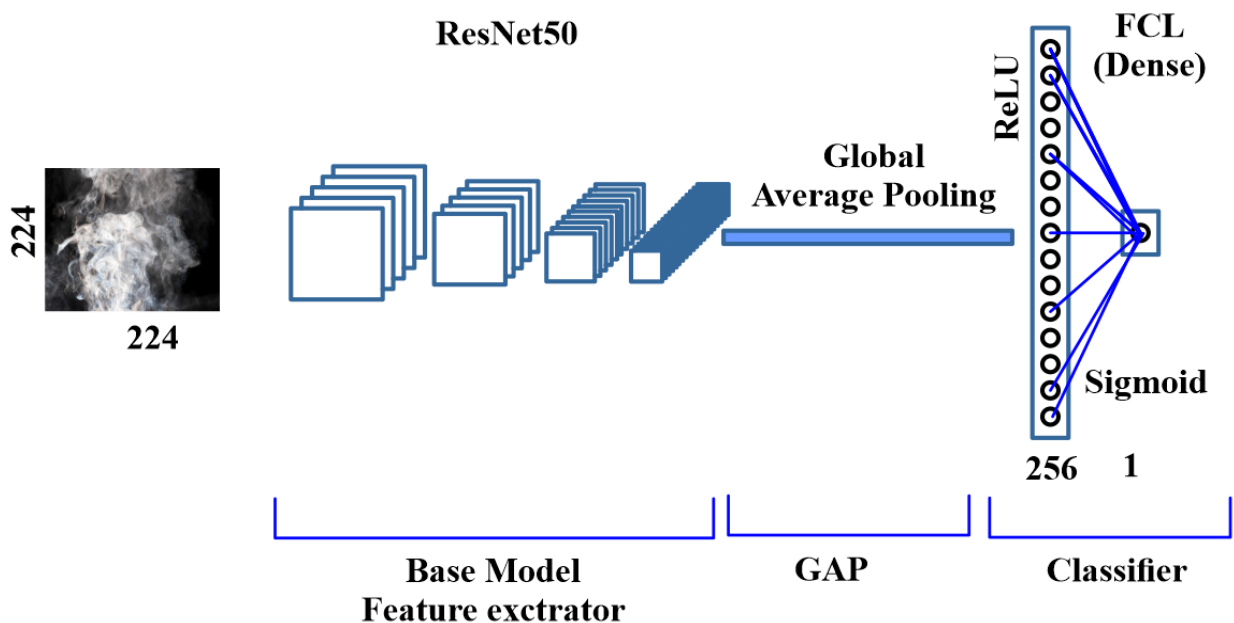

Figure 5.2 Proposed smoke detection model architecture.

Table 5.1 Compiling Parameters.

| Parameter | Type |
|---|---|
| Optimizer | Adam (learning_rate = 0.001) |
| Loss | Binary cross-entropy |
| Metrics | Accuracy |

Table 5.1 displays the parameters of the proposed model, including the optimizer, loss function, and evaluation metric. The ResNet-50 base is disabled from updating its weights during the training process by freezing all layers in the feature extractor. The binary cross-entropy loss function and Adam optimizer with a learning rate of 0.001 are used due to the 20 different classes in the dataset. The model's performance is evaluated using the accuracy metric.

Two callbacks are utilized during training, where the first saves the best weights, while the second stops training early if there is no significant improvement after ten epochs. The model is developed using Jupyter Notebook and Google Colab and is deployed on a machine with an Intel Core i5 - 2.5 GHz processor and 16 GB RAM. The model achieves an inference time of 0.61 seconds to make a prediction.

## 5.2 Validation Results of Smoke Detection Model

The binary classification model for detecting smoke in laser cutting exhibited a significant improvement in accuracy during the initial epochs of training, as illustrated in Fig. 5.3. After 100 epochs of training, the model attained a high accuracy of 100%, which indicates its capability to learn the underlying patterns in the data and generalize effectively. These results suggest that the model has the potential to be utilized in practical applications to detect the smoke in laser cutting machines.

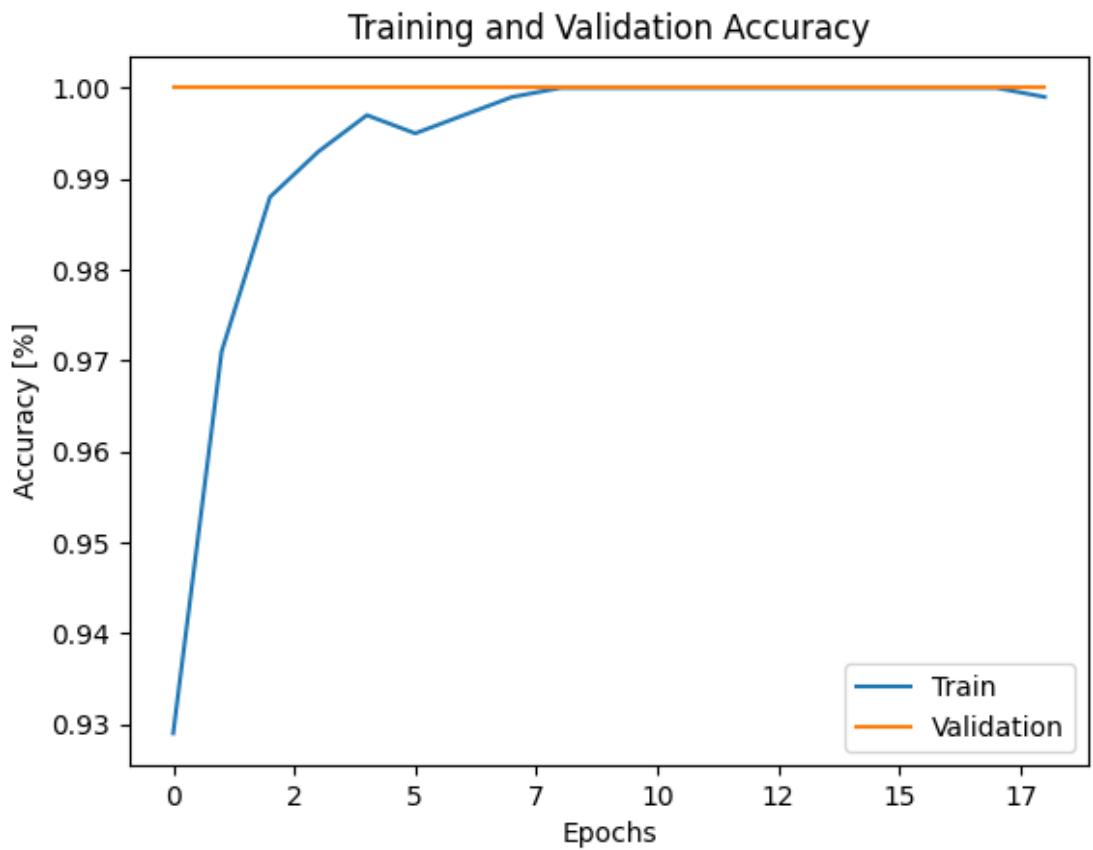

Figure 5.3 Training and validation accuracy of the smoke detection model.

The model's loss function, shown in Fig. 5.4, exhibited a sharp decline during the initial epochs of training, indicating that the model quickly learned the patterns in the data. As training progressed, the loss continued to decrease at a slower rate until it reached a plateau at approximately 0, indicating that the model had converged to a good solution. The low loss value achieved by the model demonstrates its effectiveness in capturing the significant features of the smoke, implying its ability to generalize well to new, unseen data. Taken together, the model's high accuracy and low loss make it a promising tool for practical applications in smoke detection in laser cutting.

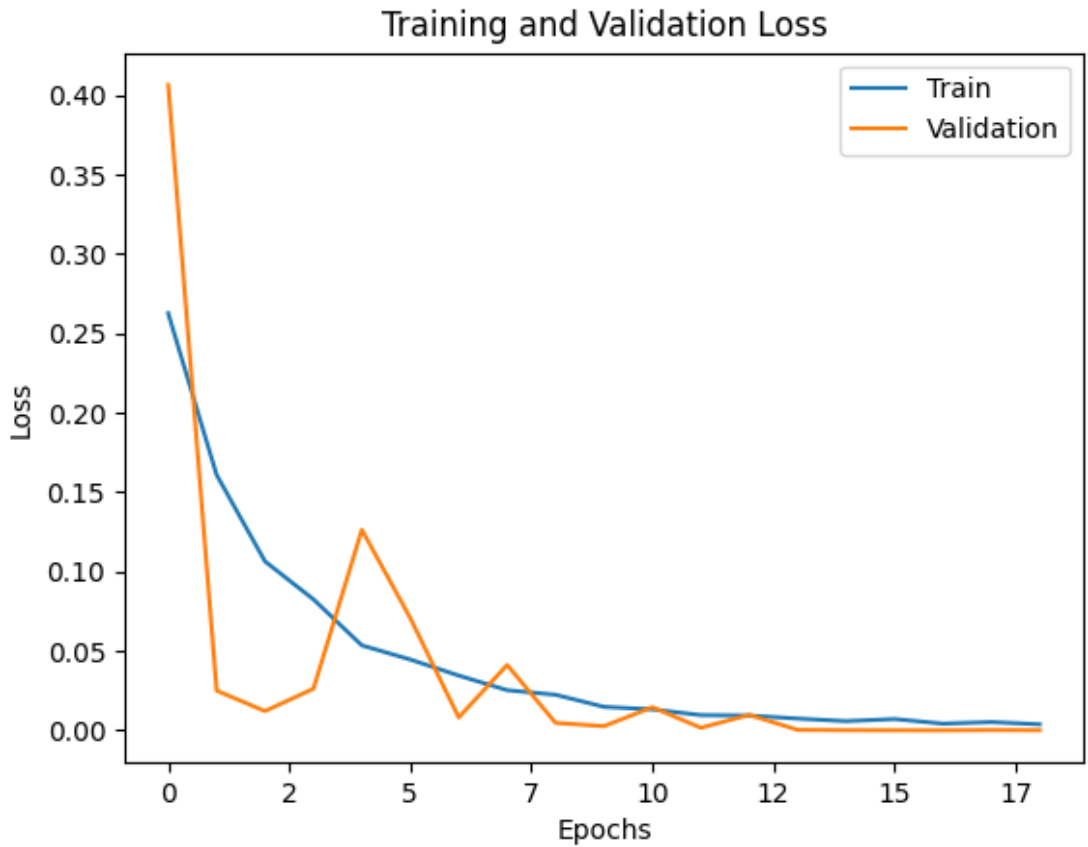

Figure 5.4 Training and validation loss of the smoke detection model.

The smoke detection model's accuracy and efficacy in classifying smoke and non-smoke images has been validated using a new test set of 2500 images from a real time video, demonstrating its reliability. The model's performance was evaluated using a confusion matrix to determine the number of correct and incorrect classifications made by the model and to assess its accuracy and ability to generalize well to new data. After retraining the model on the resulting test set, a confusion matrix has been generated for the predictions and shown Figure 5.5, the model rapidly improved in accuracy, achieving a high level of accuracy in just a few epochs. This suggests that the used dataset provided the model with better learning opportunities, enabling it to learn the underlying patterns in the data more effectively.

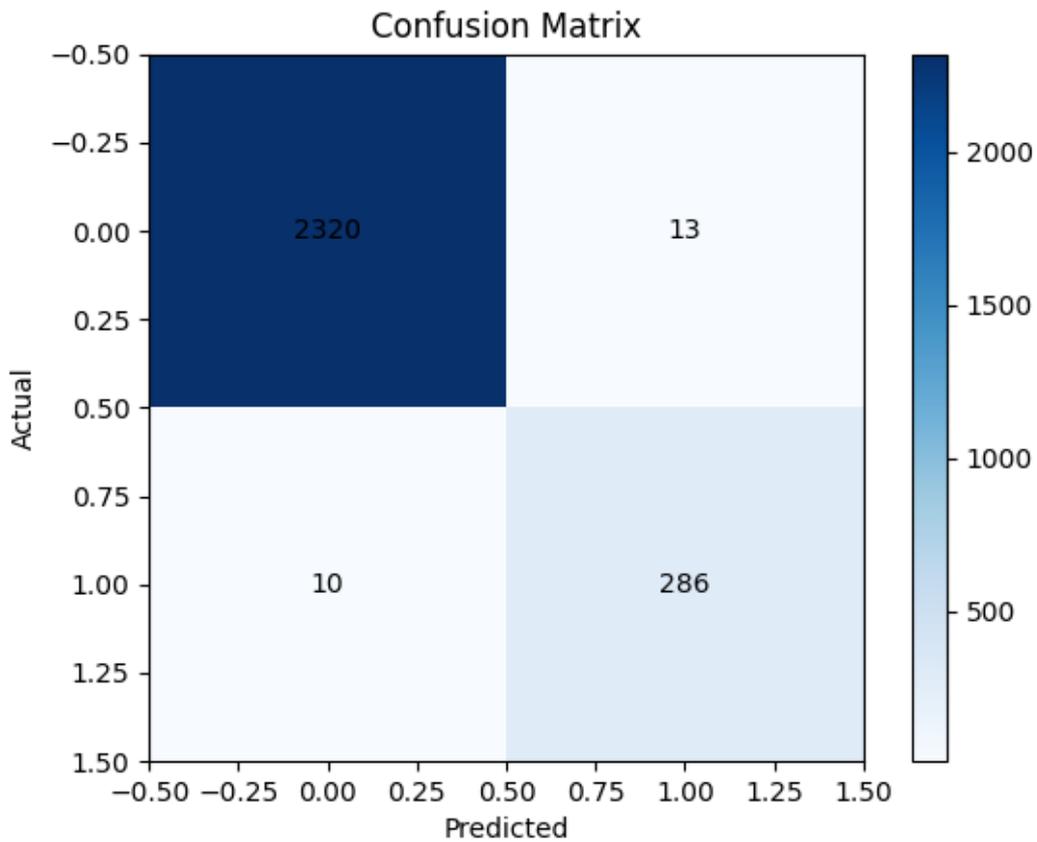

Figure 5.5 Confusion matrix for the smoke detection model.

Furthermore, the model's performance in detecting the smoke in the laser cutting images was evaluated using a classification report of accuracy, precision, recall, and F1-score and shown in Figure 5.6.

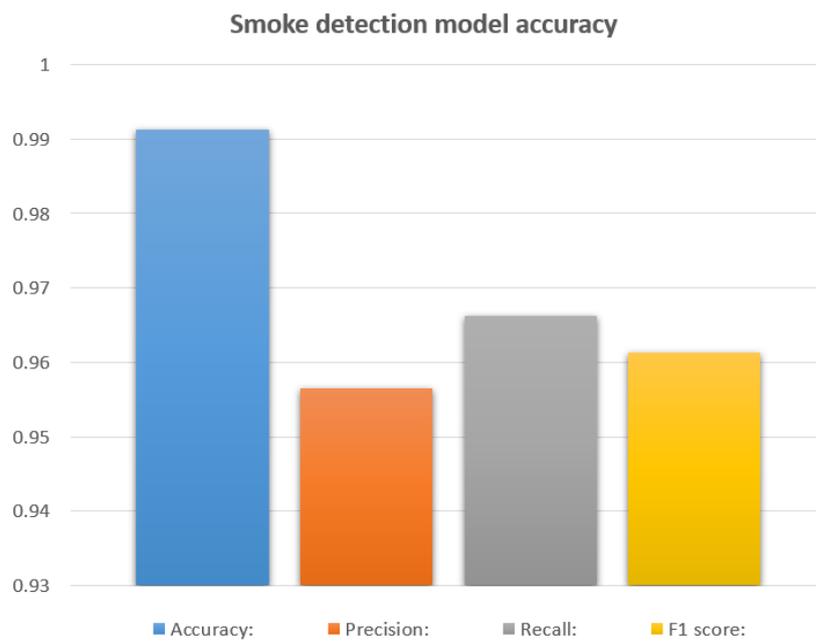

Figure 5.6 Accuracy, precision, recall, and F1-score for the smoke detection model.

The AUC ROC, which stands for "Area Under the Receiver Operating Characteristic Curve," is a metric used to assess the performance of a classification model, in this section it used to evaluate the performance of the model, which depicted the trade-off between the true positive rate and the false positive rate, it shown in Figure 5.7. Overall, the confusion matrix, classification report, and ROC curve demonstrate the efficacy of the model in smoke detection and classification, emphasizing its potential for practical applications.

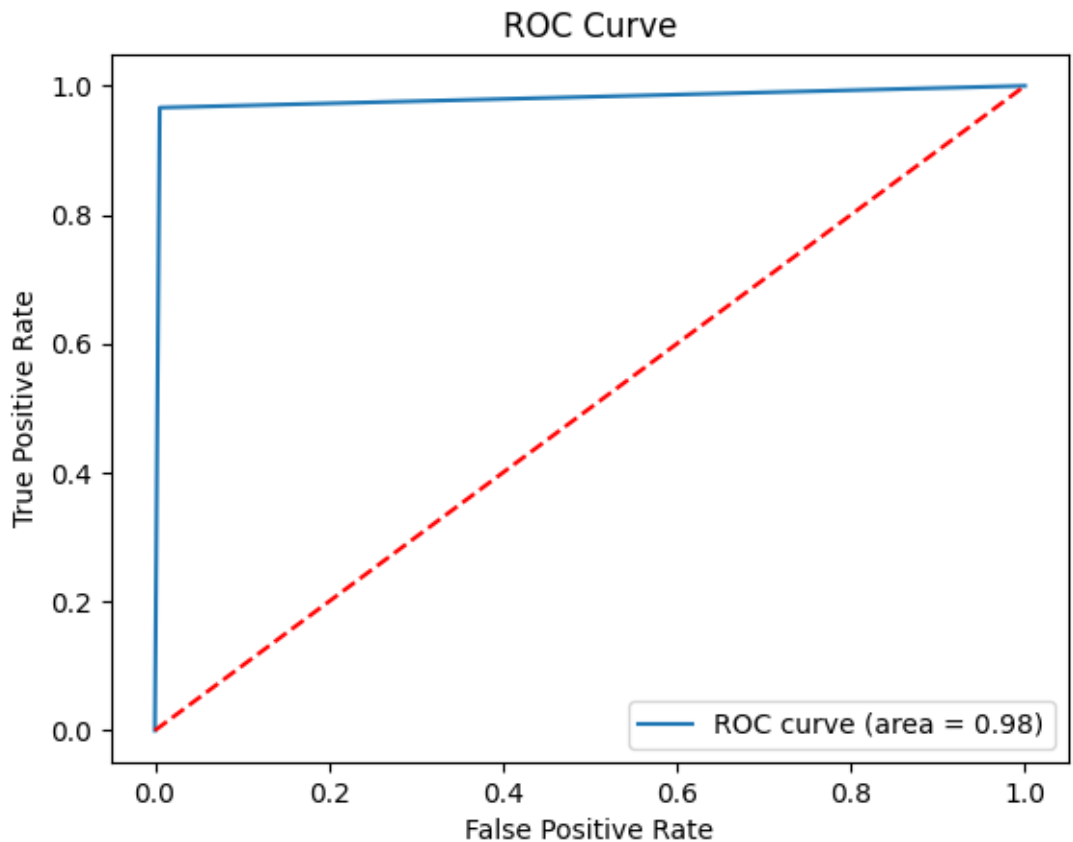

Figure 5.7 AUC ROC Curve for the smoke detection model.

## 5.3   The Proposed Energy Management Approach

The energy management approach introduced in this chapter constitutes a holistic strategy that integrates the findings and models established in prior chapters. It relies on the outputs of the two core deep learning models presented earlier: the material classification model, which accurately identifies the type of material being subjected to laser cutting, and the smoke detection model, which assesses whether smoke is generated during the cutting process. These models serve as the foundation for controlling the exhaust suction pump in a dynamically responsive manner. Figure 5.8 illustrates the fundamental concept and depict the flowchart that elucidates the key components of this innovative energy-saving approach.

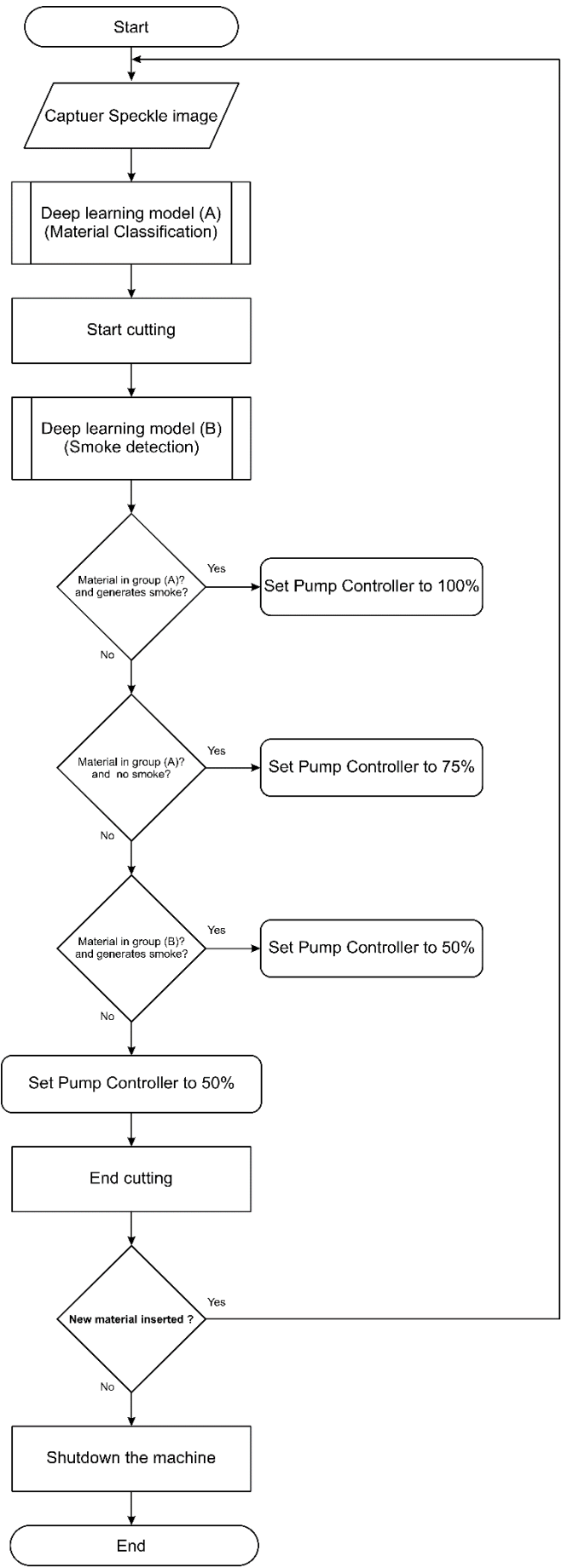

Figure 5.8 Flow chart for the proposed deep learning models for energy saving in laser cutters.

As depicted in the flowchart, once a standard cutting process commences, the material classification approach takes the initial step of identifying the specific material being processed. Subsequently, the smoke detection model comes into play, evaluating whether smoke is generated during the ongoing cutting process. This pivotal step assumes a critical role in the realm of energy management. It empowers the laser cutting machine to dynamically adapt its operating parameters based on the material being cut and the occurrence of smoke generation, ensuring a finely tuned and efficient process until the completion of the cutting operation.

## 5.4 Results of the Proposed Energy Management Approach

### 5.4.1 Calculation of Total Energy Consumption and Savings

To evaluate the efficacy of the adaptive air suction control system in minimizing energy consumption, the total power demand and power savings were quantitatively analyzed. This section outlines the mathematical approach used for calculating these metrics.

#### 5.4.1.1 Baseline Scenario

In the absence of the adaptive control system (without the proposed approach), the baseline power demand of the air suction pump remains constant throughout cutting operations. The baseline power demand is denoted as $P_b$, which is also the maximum power level for the exhaust system and equals 100%.

#### 5.4.1.2 Scenario with Closed-loop Control

With the implementation of the closed-loop air suction control system, the power demand of the air suction pump becomes variable, dynamically adjusting in response to dust and smoke levels. This adaptive control scenario's power demand is represented as $P_a$.

#### 5.4.1.3 Total Power demand

To assess the overall energy consumption, the total power consumed by the air suction pump was calculated for both scenarios.

- Baseline Scenario Total Power demand $\boldsymbol{P'_b}$.
- Adaptive Control Scenario Total Power demand $\boldsymbol{P'_a}$.

*5.4.1.4   Total Energy Savings*

In addition to power savings, the total energy savings were also evaluated by comparing the energy consumption of the two scenarios. The percentage of energy savings encompasses the reductions achieved in both power demand and the time duration of each operation as presented in equation 5.1.

$$E_s\% = \frac{E'_b - E'_a}{E'_b} \cdot 100 \qquad (5.1)$$

Where, $E'_a$ is the total energy consumed after applying the proposed approach for each operation, and $E'_b$ is the total energy consumed by over the time of each operation and could be calculated according to the following equation 5.2.

$$E'_b = P_b \cdot t \qquad (5.2)$$

### 5.4.2   Total Energy Consumption and Savings

In this study, a $CO_2$ Macro laser cutting machine served as the experimental platform for assessing the energy management approach. The machine's power profile encompassed various components, including the energy consumption of the tube cooling system (specifically, a CW5200 Chiller), the power required by the air compressor, the $CO_2$ laser's power source, and the energy demands of the machine's moving systems, which were actuated by stepper motors. However, the primary focus of this research centred on evaluating the power usage of the exhaust and ventilation system, a crucial aspect within the study's scope. Figure 5.9 illustrates an intriguing finding: the energy consumed by the exhaust system exceeded that of the laser source itself. This observation is particularly significant given that this laser cutting machine exclusively processes non-metallic materials and operates with a laser source drawing a mere <120 W of input power, while the exhaust system demands a substantial 0.75 KW to effectively fulfill its role. In the depicted cutting operations, the exhaust system's power $P'_b$ remains constant, without control. The primary objective here is to minimize this power demand. To establish a generalized framework, a percentage will be introduced as a reference point for comparison. This percentage, known as the 100% baseline, serves as a foundational benchmark against which power savings can be measured and evaluated.

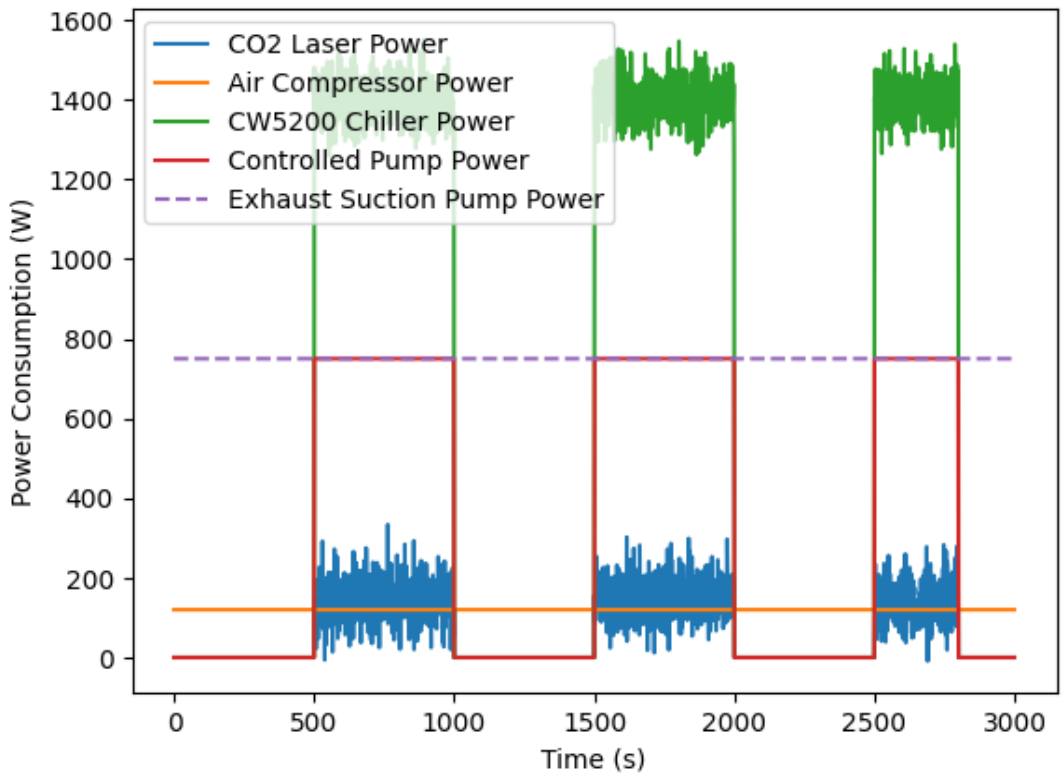

Figure 5.9 Comparison of power demand before and after applying the proposed approach for the same laser cutting process.

Figure 5.10 illustrates the power demand during a laser cutting operation that spans 3000 seconds for the completion of the cutting process. The findings indicate that the conventional laser cutting machine lacks implemented control strategies for its air suction pump. Consequently, the overall power demand throughout the ongoing cutting process, prior to the implementation of the suggested approach, stands at 100%.

Considering the broader energy perspective, it's evident that this power demand translates into a significant energy expenditure over the course of the cutting process. With the introduction of the proposed approach, power demand experiences fluctuations based on the operational mode. The air suction pump is now activated exclusively when smoke or dust are detected, a state contingent on the laser's activity. As a result, the pump only engages during laser operation, aligning its functionality with the laser's presence. This strategic synchronization not only conserves energy but also enhances the efficiency of the cutting process.

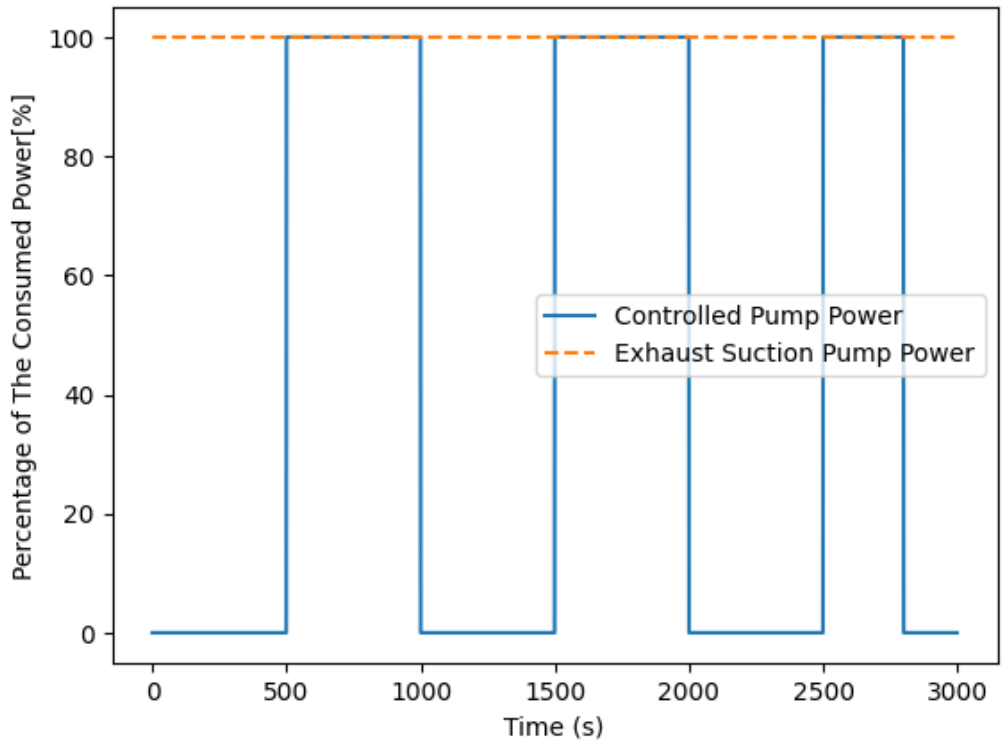

Figure 5.10 Power demand across laser cutting operation before and after applying the proposed approach for the wood materials and without regulating the input power to air suction pump.

Given the dynamic nature of smoke and dust production rates, which inherently vary based on the composition and characteristics of the material being subjected to cutting, a discernible avenue for mitigating overall energy consumption during the cutting procedure presents itself. This phenomenon is expounded upon through a comprehensive evaluation of the energy dynamics, as depicted in Figure 5.11

As mentioned in the flowchart the power of the exhaust pump relies on the material begin cut, there are three different power levels according to the materials categories, the first category consists of materials that generate huge amounts of smoke during the cutting process; including wood, MDF, leather, and metals, the second category comprises most of plastics such as Acrylic, TPU, the last category consists of fabrics, textile materials, and papers.

If the detected material belongs to the category A this means that the pump would work in the maximum capacity 100% input power, the second category B the pump power would set to 75% from its maximum working power, lastly the category C sets the power to 50% from the maximum input power.

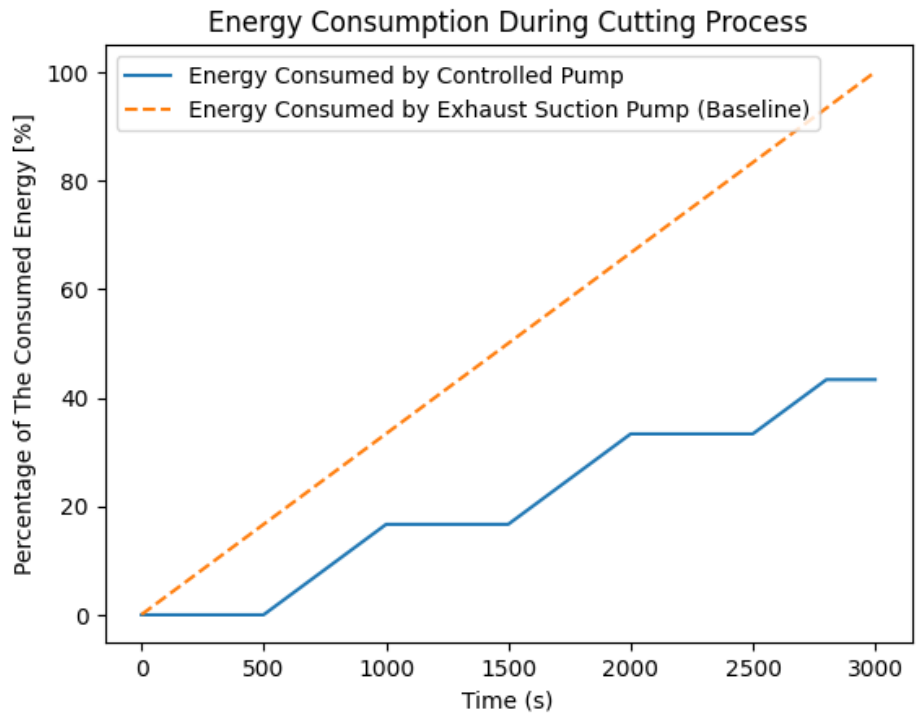

Figure 5.11 Energy consumption across laser cutting operation before and after applying the proposed approach for the wood materials and without regulating the input power to air suction pump.

In light of this understanding, a meticulous control strategy was tactically deployed with the primary aim of orchestrating the air suction pump's activation based exclusively on the operational mode of the laser cutting machine. This initial approach, while displaying an appreciable level of minimization, nonetheless reveals untapped potential for more pronounced energy conservation.

To that end, the research paradigm evolved to encompass the integration of a novel deep learning framework, heralding a paradigm shift in energy-saving endeavors during the laser cutting process. This advanced approach capitalizes on the material-specific recognition facilitated by the proposed deep learning algorithm. The discernment of the material undergoing cutting infuses a heightened layer of intelligence into the energy management framework.

The synthesis of this innovative approach yields substantial dividends in terms of energy efficiency, particularly pronounced when encountered with materials characterized by a reduced propensity for generating voluminous smoke and dust. Notably, materials such as acrylics and select plastics, along with textured variants, exemplify this low smoke and dust production profile. This energized stride towards optimization is cogently showcased through the empirical depiction in Figure 5.12, substantiating the advantageous implications of material-aware energy modulation.

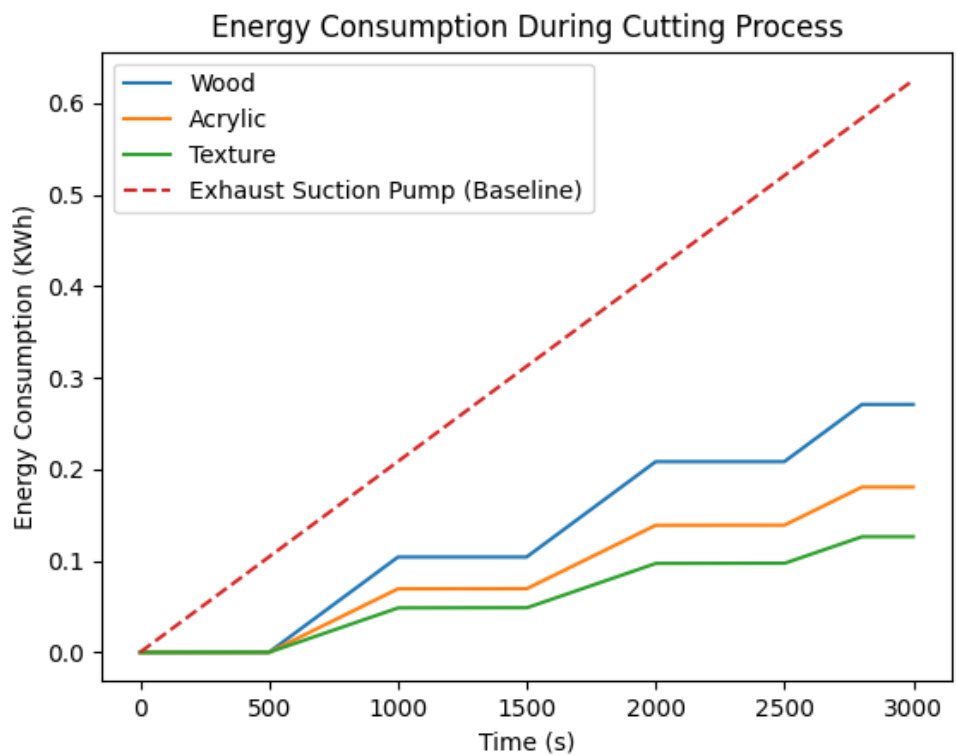

Figure 5.12 Energy consumption across laser cutting operation before and after applying the proposed approach for the wood, Acrylic, and texture materials and the baseline consumption from the air suction pump.

Investigating the energy dynamics, it is notable that the rate of energy consumption during the cutting process can be significantly enhanced through the tailored control strategies. Calculations reveal the tangible benefits in terms of energy savings across distinct cutting scenarios. The wood cutting scenario showcases a commendable energy conservation, yielding an energy savings of approximately 56.66% in comparison to the baseline exhaust suction pump operation. Similarly, the acrylic cutting scenario demonstrates a pronounced energy preservation, resulting in an energy savings of approximately 71.11%. Impressively, the texture cutting scenario further accentuates the efficacy of the approach, yielding an energy savings of around 79.77%. These substantial energy savings underscore the proposition that material-aware control strategies, guided by advanced deep learning algorithms, present an avenue not only for enhancing process efficiency but also for ushering in a more sustainable energy utilization paradigm.

In the pursuit of unraveling the nuanced dynamics of energy consumption and savings, a series of visualizations were harnessed to succinctly encapsulate the insights garnered. As exemplified in Figure 5.13, the energy consumption comparison chart presents a meticulous breakdown of scenarios, shedding light on the varying energy utilization across different materials and operational modes. Notably, the chart reveals the energy consumption disparities among scenarios, where the Texture and Acrylic cutting scenarios stand out with marked differences in their energy utilization profiles, in contrast, the Wood scenario and the baseline 'Exhaust Suction Pump' operation.

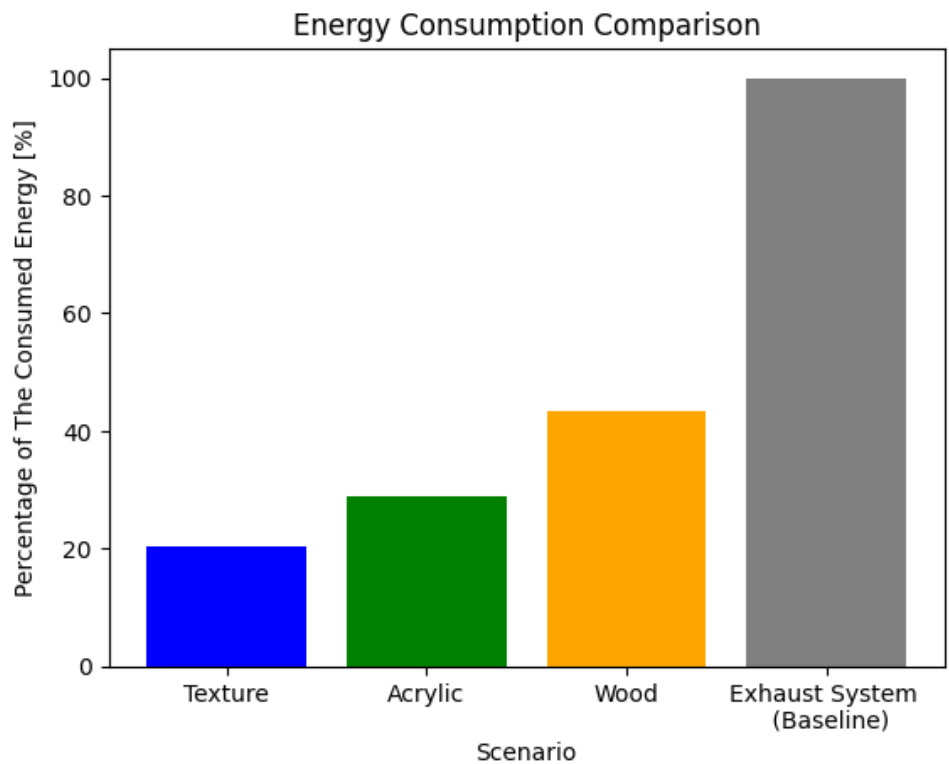

Figure 5.13 Energy consumption when cutting the wood, Acrylic, and texture materials from the air suction pump using the same cutting process.

A nuanced extension of this analysis is observed in Figure 5.14, where a refined perspective emerges through the juxtaposition of energy consumption and savings. The grouped bar chart harmoniously melds these facets, capturing the interplay of energy expenditure and conservation. Striking a harmonious visual balance, this chart offers a comprehensive view, underscoring the tangible energy savings realized by the Texture, Acrylic, and Wood scenarios when juxtaposed against the energy-hungry backdrop of the baseline operation. These visual representations serve as informative signposts in navigating the complex terrain of energy optimization within the realm of laser cutting operations.

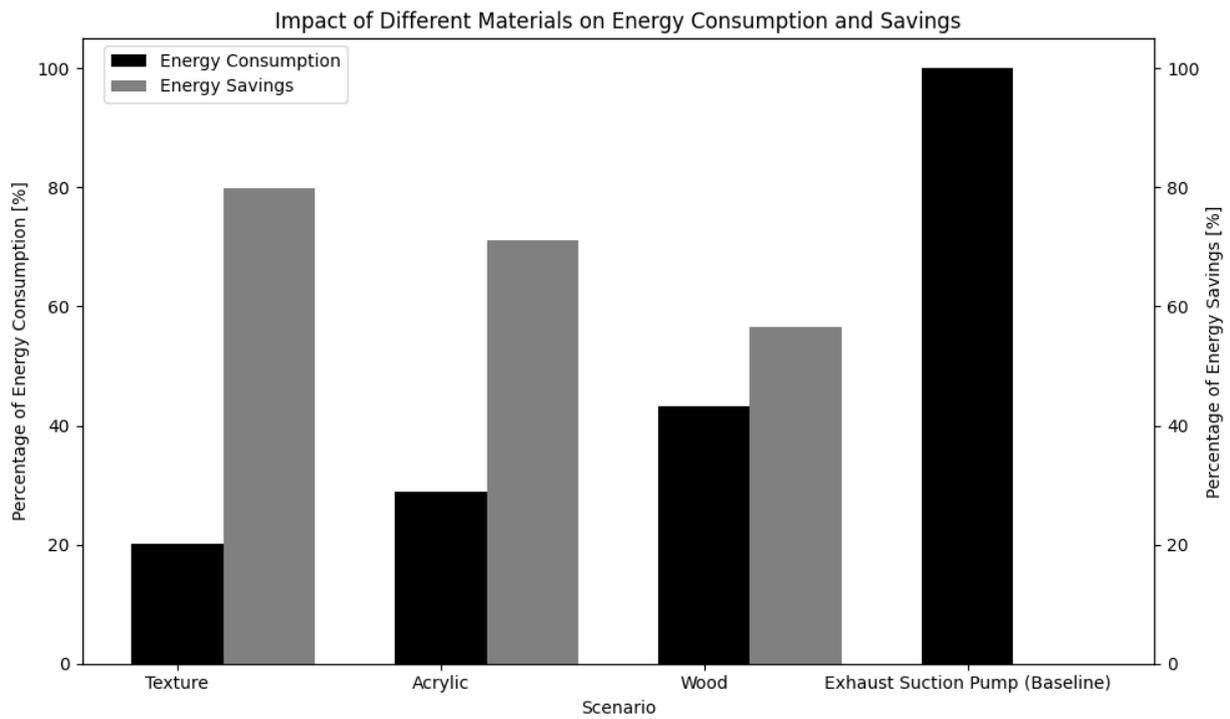

Figure 5.14 Energy savings the selected materials while processing by the same cutting process.

The proposed approach of utilizing deep learning models for material classification and smoke detection, and controlling the exhaust suction pump has shown significant energy savings in laser cutting processes. The real-time energy consumption data before and after implementing the deep learning models demonstrated a reduction in energy consumption, indicating the efficacy of the proposed approach in energy management. Overall, the proposed approach provides a promising solution for efficient and safe laser cutting processes with reduced energy consumption.

## 5.5   Summary of the chapter

In this chapter, a comprehensive exploration unfolds, aiming to enhance both the energy efficiency and safety aspects of laser cutting machines. The primary focus centers on addressing the energy consumption and potential environmental and health risks posed by the generation of smoke and fumes during the cutting process. To tackle this issue, a deep learning model, built upon a convolutional neural network (CNN) architecture, was meticulously developed. This model, employed for detecting and identifying smoke and fumes, was rigorously trained on a substantial dataset comprising 2500 images.

The chapter commences by elaborating on the setup and image acquisition process, underscoring the significance of capturing a diverse range of smoke and fume patterns. The dataset, partitioned into training and validation sets, serves as the foundation for training and evaluating the smoke detection model's performance.

A deep dive into the architecture of the proposed smoke detection model follows, which leverages the capabilities of the pre-trained ResNet50 model for feature extraction. Further, the model's configuration, including layer specifications, optimization parameters, and evaluation metrics, is meticulously detailed.

Subsequent sections delve into the validation results of the smoke detection model, demonstrating its exceptional accuracy and effectiveness in classifying images as smoke or non-smoke. A variety of evaluation metrics, including accuracy, precision, recall, F1-score, and the AUC ROC curve, attest to the model's robust performance.

The narrative then transitions to the heart of the chapter: the proposed energy management approach. This innovative strategy integrates insights and models from previous chapters, utilizing the material classification model to identify the material being cut and the smoke detection model to assess the presence of smoke. The synchronized control of the exhaust suction pump based on these inputs is the core of this energy-saving approach, illustrated through an insightful flowchart.

The subsequent section quantitatively analyzes the results of this approach, focusing on the total power demand and savings. It delineates the methodology for calculating these metrics, introducing the baseline scenario and the closed-loop control scenario, underscoring the role of the closed-loop air suction control system in minimizing energy consumption.

The chapter culminates in a comprehensive analysis of energy consumption and savings across various materials and cutting scenarios. Visualizations and charts clearly depict the energy dynamics, showcasing substantial energy savings achieved through material-aware control strategies guided by advanced deep learning algorithms.

In conclusion, this chapter propels the field of laser cutting towards enhanced energy efficiency and safety. The integration of deep learning models and intelligent control strategies not only reduces energy consumption but also ensures the sustainable utilization of resources, marking a significant step forward in laser cutting technology.

# Chapter 6: CONCLUSION AND FUTURE WORK

## 6.1 Overview

This chapter presents the conclusion and future work of the proposed approach for material classification and smoke detection in laser cutting processes using deep learning models. The proposed approach demonstrated promising results in distinguishing between various types of materials and detecting smoke produced during laser cutting, contributing to efficient and safe laser cutting processes with reduced energy consumption.

An evaluation for the performance of the first model is conducted, throughout the evaluation, the SensiCut dataset is used with the proposed channel splitting model, particularly in terms of inference and training time, performance, and number of parameters. The evaluation process will involve separating 3,000 images from the dataset before training and then testing the model on these images after training to determine whether the model can classify these images with high accuracy. The results have been compared with the original SensiCut model and the confusion matrix between the predicted materials and the actual materials in the dataset has been presented.

In the second part of this thesis, another evaluate to the newly generated dataset has presented by testing the dataset on a deep learning model that can classify the types of materials in the dataset. The dataset divided into two parts: a test set of 300 images and a training set of 6000 images were generated. The model trained on the training set, and its performance evaluated on the test set. Then a confusion matrix between the predicted materials and the actual materials in the dataset has been presented.

Eventually, an evaluation for the proposed smoke detection model in laser cutting. This has been done by testing the model real time on a laser cutting process while cutting some materials that produce smoke and other materials that do not produce smoke. The training and validation accuracies are presented, additionally, a confusion matrix and F1-score, precision, recall and ROC curve are utilized to evaluate this approach.

## 6.2 Contributions to existing knowledge

The proposed approaches contribute to the existing knowledge in the field of laser cutting by utilizing deep-learning models for material classification and smoke detection, while controlling the exhaust suction pump to minimize energy consumption. The material classification model employs speckle sensing with a low-cost USB camera to distinguish between 18 different surface types with high accuracy, enabling precise and efficient control of laser power and speed. The smoke detection model detects smoke produced during laser cutting and triggers the exhaust suction pump to operate only when necessary, reducing energy consumption and increasing safety during the laser cutting process. The proposed approach also provides real-time energy consumption data before and after implementing the deep learning models, demonstrating a significant reduction in energy consumption, indicating the efficacy of the proposed approach in energy management. Overall, the proposed approaches provide a promising solution for efficient and safe laser cutting processes with reduced energy consumption, based on AI techniques.

## 6.3 Future work

The proposed approach can be further improved and extended in several ways. One potential avenue for future work is to explore the use of additional sensors and data sources to enhance the accuracy of the material classification and smoke detection models. For example, the addition of temperature sensors may provide further insight into the laser cutting process and help to optimize energy consumption.

Another area for future work is to investigate the use of reinforcement learning algorithms to optimize the laser cutting process based on real-time feedback from the material classification and smoke detection models. This approach may enable the laser cutting machine to learn and adapt to new materials and cutting scenarios, further increasing efficiency and reducing energy consumption.

Moreover, the proposed approach can be extended to other domains beyond laser cutting, such as material processing and manufacturing, where deep learning models can be utilized to optimize energy consumption and improve safety.

In conclusion, the proposed approach for material classification and smoke detection in laser cutting processes using AI-based deep learning models has demonstrated promising results and provides a promising solution for efficient and safe laser cutting processes with reduced energy consumption. Future work can further refine and extend the proposed approach, contributing to the advancement of the field of laser cutting and material processing.

# Appendix A: Equipment and Components

This appendix provides an overview of the electronic components and equipment utilized in the implementation of the deep learning models and the energy management approach in this study. The specific models and specifications of these items are detailed to ensure clarity and transparency regarding their role in the conducted experiments.

**Camera(A): Raspberry Pi Camera**

This Camera Module is specifically designed for easy plug-and-play compatibility with the Raspberry Pi. It connects to the Raspberry Pi through one of the two small sockets located on the board's upper surface.

It features a built-in IR-CUT camera that automatically switches between day and night modes, ensuring improved image quality in different lighting conditions. This interface utilizes the dedicated CSI interface, which provides a higher bandwidth link for transmitting pixel data from the camera to the processor. This data transfer occurs along the ribbon cable that connects the camera board to the Pi. The camera sensor itself has a native resolution of 5 megapixels, offering still picture resolution at 2592 x 1944 pixels. It also supports video recording at 1080p30, 720p60, and 640x480p60/90.

The camera can capture static images at a resolution of 2592 x 1944 pixels and is also capable of video recording at 1080p@30fps, 720p@60fps, and 640x480p@60/90fps, thanks to its 5MP OV5647 webcam sensor. It is fully compatible with the latest version of Raspbian, the preferred operating system for Raspberry Pi.

**Specifications:**

- CCD size: 1/4 inch.
- Aperture(F): 1.8.
- Focal Length: 3.6 mm (adjustable).
- Field of View: 72 degrees.
- Preferred Sensor Resolution: 1080p.
- Dimension: 1.42inch x1.04inch x0.94inch.
- 4 screw holes which are used for a fixed position.
- Support 3.3V power output.
- Supports connecting infrared LED or flash LED.

**Camera(B): USB camera**

The Kisonli PC-3 Webcam is a versatile and user-friendly webcam solution. It boasts a video resolution of 640 x 480 pixels at a smooth 30 frames per second (30FPS), ensuring clear and fluid video quality. This webcam is equipped with an auto white balance feature, which optimizes the image's color balance for different lighting conditions. What sets it apart is its built-in microphone, enabling convenient audio input during video calls and conferences. It interfaces seamlessly with USB 2.0, ensuring compatibility with a wide range of devices. Its sleek black color adds a touch of sophistication to its design. With a maximum resolution of 640x480 and a CMOS capture sensor, this webcam is a reliable choice for various video-related tasks, making it a valuable addition to your setup.

**Specifications:**

- Capture Sensor: CMOS
- Max. Resolution: 640x480 pixels
- Interface: USB 2.0
- Color: Black
- Model: Kisonli PC-3 HD 480P/30FPS

**Microprocessor: Raspberry Pi Zero**

- Model: Raspberry Pi Zero
- Specifications: 1GHz CPU, 512MB RAM
- Usage: Image acquisition

**Microprocessor: Arduino Uno**

- Model: Arduino Uno
- Specifications: 32KB flash memory, 2KB SRAM, 16M HZ
- Usage: DIY laser cutter

**Motor driver: Ramps 1.4**

- Model: Ramps 1.4
- Specifications: CNC Shield for Arduino Uno
- Usage: DIY laser cutter

**Laser pointer: Red Laser pointer**

- Model: Red Laser pointer
- Specifications: 650 nm – 5 < mW
- Usage: Speckle generation

**Laser pointer: Green Laser pointer**

- Model: Green Laser pointer
- Specifications: 535 nm – 1 < W
- Usage: Speckle generation

These components played integral roles in various aspects of the study, including image generation, data acquisition, and the operation of the DIY laser cutting machine. The detailed specifications provided here ensure a comprehensive understanding of their significance within the experimental setup.

# Appendix B: Source Code and Model Details

This appendix presents the source code and model details employed in implementing the deep learning models discussed throughout this thesis. Python served as the primary programming language to execute various algorithms and construct these models. Each section in this appendix corresponds to a specific deep learning model, showcasing not only the source code but also comprehensive insights into the model architecture and the preprocessing steps applied to input images. This resource aims to provide a deeper understanding of the technical aspects underpinning the methodologies, all developed utilizing TensorFlow and Keras APIs within the Google Colab and Kaggle platforms.

## Pre-processing and Train Test Split for the Sensicut dataset

```python
import numpy as np # linear algebra
import pandas as pd # data processing, CSV file I/O (e.g. pd.read_csv)
import os,shutil

# Select the materials
dir_num = [0,3,4 ,12 ,15 ,16 ,17 ,18 ,19 ,21 ,24 ,27 ,28 ,32 ,33 ,34 ,35
,37 ,38 ,43 ,44 ,46 ,47 ,50 ,51 ,52 ,54 ,55 ,57 ,58]

# Function to count the total number of image inside a directory
def count_files(dir_name):
  counter=0
  for i in os.listdir(dir_name):
    for d in os.listdir(dir_name+'/'+i):
      counter+=1
  print(counter)

count_files('/kaggle/input/lsp-grn/g/train')
```

31299

```python
# List the number of images in side each directory
for n,i in enumerate(os.listdir('/kaggle/input/lsp-grn/g/train')):
    if n in dir_num:
        print(len(os.listdir(f'/kaggle/input/lsp-grn/g/train/{i}')), f':
{i}')
```

512 : 2020.07.20-20.18.48-WhiteAcrylic
512 : 2020.07.19-14.52.59-BirchPlywood
512 : 2020.08.06-18.02.08-LexanWhite
512 : 2020.08.03-16.55.00-WhiteDelrin
512 : 2020.08.12-16.01.10-WhiteStyrene

```
512  : 2020.07.19-18.17.36-WhiteMelamineMDF
512  : 2020.08.04-19.15.28-WhiteFelt
512  : 2020.08.16-19.00.15-GreenCardstock
512  : 2020.07.17-18.15.24-MDFnatural
512  : 2020.08.16-19.58.55-StandardAluminum
512  : 2020.08.17-16.57.45-WalnutHardWood
1025 : 2020.08.27-18.04.05-PETG
512  : 2020.08.15-19.37.06-RedSuede
512  : 2020.07.20-14.36.04-BambooPremiumVeneerMDF
512  : 2020.08.16-21.44.15-CarbonSteel
512  : 2020.09.03-10.54.02-Foamboard
512  : 2020.07.17-15.50.11-MapleHardWood
512  : 2020.08.04-17.07.43-BlackLeather
512  : 2020.08.13-21.01.47-GreenMatboard
512  : 2020.08.16-20.52.29-StainlessSteel
1025 : 2020.08.27-15.46.08-Acetate
512  : 2020.08.06-16.59.03-ABS
512  : 2020.08.12-18.01.48-WhiteCorrugatedCard
512  : 2020.07.17-14.54.51-RedOakHardwood
512  : 2020.08.01-17.33.33-PVCClear
512  : 2020.07.19-17.13.22-BlondeBamboo
512  : 2020.07.31-18.04.34-ExtrudedAcrylicBlack
512  : 2020.07.17-19.16.01-Cork
512  : 2020.07.31-17.11.32-CarbonFiber
512  : 2020.08.05-18.26.21-BlackSilicone
```

```python
# function to create a new train, test, and Val directories
def test_val_dir(dir_name):
    os.mkdir(dir_name)
    for n,i in enumerate(os.listdir('/kaggle/input/lsp-grn/g/train')):
        if n in dir_num:
            os.mkdir('/kaggle/working/{}/{}'.format(dir_name,i))

os.mkdir('g-30')
test_val_dir('g-30/train')
test_val_dir('g-30/test')
test_val_dir('g-30/val')

# Make sure that the directories contain the 30 selected materials

Print(len(os.listdir('/kaggle/working/g-30/val')))
```

```
30
```

```python
# Function to copy the images from the dataset into the new directories
def copyfiles(file_name):
    count = 0
    dir = f'/kaggle/input/lsp-grn/g/{file_name}/'
    for i in os.listdir(f'/kaggle/working/g-30/{file_name}/'):
        for d in os.listdir(dir + i):
            count+=1
            shutil.copy( f'{dir}{i}/{d}' , f'/kaggle/working/g-30/{file_name}/{i}/{d}' )
    print(count)
```

```python
# Copy the images to the new directories and print the number of copied im

copyfiles('train')
copyfiles('val')
copyfiles('test')
```

16386
4160
190

```python
# List the number of images inside each directory in the new dataset
for i in os.listdir('/kaggle/input/lsp-grn/g/train'):
    print(len(os.listdir(f'/kaggle/input/lsp-grn/g/train/{i}')), f': {i}')
```

512 : 2020.07.20-20.18.48-WhiteAcrylic
512 : 2020.07.20-15.37.49-BlackCoatedMDF
512 : 2020.07.21-14.38.16-GreenAcrylic
512 : 2020.07.19-14.52.59-BirchPlywood
512 : 2020.08.06-18.02.08-LexanWhite
512 : 2020.08.05-19.36.47-BlackMatboard
512 : 2020.07.21-15.48.42-GreenTintedAcrylic
512 : 2020.07.17-17.13.59-AmberBamboo
577 : 2020.08.04-14.30.01-BlackDelrin
512 : 2020.08.04-16.06.55-BlackFelt
512 : 2020.09.04-14.49.30-RedPaper
512 : 2020.08.17-15.54.34-WhiteMatteAcrylic
512 : 2020.08.03-16.55.00-WhiteDelrin
512 : 2020.07.21-13.40.08-BlackMatteAcrylic
512 : 2020.08.03-15.59.37-CreamAcrylic
512 : 2020.08.12-16.01.10-WhiteStyrene
512 : 2020.07.19-18.17.36-WhiteMelamineMDF
512 : 2020.08.04-19.15.28-WhiteFelt
512 : 2020.08.16-19.00.15-GreenCardstock

```
512  : 2020.07.17-18.15.24-MDFnatural
512  : 2020.07.20-16.39.41-ClearAcrylic
512  : 2020.08.16-19.58.55-StandardAluminum
512  : 2020.07.19-19.10.02-BlackMelamineMDF
512  : 2020.08.14-18.38.40-OrangeWoolFelt
512  : 2020.08.17-16.57.45-WalnutHardWood
512  : 2020.07.21-12.47.04-BlackAcrylic
512  : 2020.07.31-20.40.46-ExtrudedAcrylicClear
1025 : 2020.08.27-18.04.05-PETG
512  : 2020.08.15-19.37.06-RedSuede
512  : 2020.08.14-20.44.29-RedSyntheticFelt
512  : 2020.08.03-14.07.04-RedAcrylic
512  : 2020.08.17-17.48.19-WhiteCoasterboard
512  : 2020.07.20-14.36.04-BambooPremiumVeneerMDF
512  : 2020.08.16-21.44.15-CarbonSteel
512  : 2020.09.03-10.54.02-Foamboard
512  : 2020.07.17-15.50.11-MapleHardWood
512  : 2020.07.31-16.16.25-Lexan
512  : 2020.08.04-17.07.43-BlackLeather
512  : 2020.08.13-21.01.47-GreenMatboard
512  : 2020.07.20-18.09.13-ClearMatteAcrylic
512  : 2020.08.15-20.30.06-IvoryCardstock
512  : 2020.08.14-17.38.11-BlackCardstock
512  : 2020.08.04-18.01.12-BlackSuede
512  : 2020.08.16-20.52.29-StainlessSteel
1025 : 2020.08.27-15.46.08-Acetate
512  : 2020.07.27-15.50.27-OrangeAcrylic
512  : 2020.08.06-16.59.03-ABS
512  : 2020.08.12-18.01.48-WhiteCorrugatedCard
512  : 2020.08.14-19.47.39-FireWoolFelt
512  : 2020.08.13-19.07.11-BrownCorrugatedCard
512  : 2020.07.17-14.54.51-RedOakHardwood
512  : 2020.08.01-17.33.33-PVCClear
512  : 2020.07.19-17.13.22-BlondeBamboo
512  : 2020.08.13-20.07.46-BrownCardboard
512  : 2020.07.31-18.04.34-ExtrudedAcrylicBlack
512  : 2020.07.17-19.16.01-Cork
512  : 2020.08.15-21.22.13-GreyCardstock
512  : 2020.07.31-17.11.32-CarbonFiber
512  : 2020.08.05-18.26.21-BlackSilicone
```

## Model Creation and Image Pre-processing

```python
from tensorflow.keras.preprocessing.image import ImageDataGenerator

# All images will be rescaled by 1./255
train_datagen = ImageDataGenerator(rescale=1./255,
        zoom_range=[0.8,1.2],
#         rotation_range=20
)
test_datagen = ImageDataGenerator(rescale=1./255)

train_dir = '/kaggle/input/lsp-grn/g/train'
train_generator = train_datagen.flow_from_directory(
        train_dir,
        target_size=(256, 256),
        color_mode='grayscale',
        batch_size=256,
        class_mode='categorical')

validation_dir = '/kaggle/input/lsp-grn/g/val'
validation_generator = test_datagen.flow_from_directory(
        validation_dir,
        target_size=(256, 256),
        color_mode='grayscale',
        batch_size=20,
        class_mode='categorical')
```

Found 16386 images belonging to 30 classes.
Found 4160 images belonging to 30 classes.

```python
from tensorflow.keras import layers
from tensorflow.keras import models

model = models.Sequential()
model.add(layers.Conv2D(32, (3, 3), activation='relu',
                        input_shape=(256, 256, 1)))
model.add(layers.MaxPooling2D((2, 2)))
model.add(layers.Conv2D(64, (3, 3), activation='relu'))
model.add(layers.MaxPooling2D((2, 2)))
model.add(layers.Conv2D(128, (3, 3), activation='relu'))
model.add(layers.MaxPooling2D((2, 2)))
model.add(layers.Conv2D(128, (3, 3), activation='relu'))
model.add(layers.MaxPooling2D((2, 2)))
model.add(layers.Flatten())
model.add(layers.Dense(512, activation='relu'))
model.add(layers.Dense(30, activation='softmax'))
model.summary()
```

```
Model: "sequential"
_________________________________________________________________
 Layer (type)                Output Shape              Param #
=================================================================
 conv2d (Conv2D)             (None, 254, 254, 32)      320

 max_pooling2d (MaxPooling2D  (None, 127, 127, 32)     0
 )

 conv2d_1 (Conv2D)           (None, 125, 125, 64)      18496

 max_pooling2d_1 (MaxPooling  (None, 62, 62, 64)       0
 2D)

 conv2d_2 (Conv2D)           (None, 60, 60, 128)       73856

 max_pooling2d_2 (MaxPooling  (None, 30, 30, 128)      0
 2D)

 conv2d_3 (Conv2D)           (None, 28, 28, 128)       147584

 max_pooling2d_3 (MaxPooling  (None, 14, 14, 128)      0
 2D)

 flatten (Flatten)           (None, 25088)             0

 dense (Dense)               (None, 512)               12845568

 dense_1 (Dense)             (None, 59)                30267

=================================================================
Total params: 13,116,091
Trainable params: 13,116,091
Non-trainable params: 0
_________________________________________________________________
```

```python
# Compile and train the model
from tensorflow.keras import optimizers

model.compile(loss='categorical_crossentropy',
              optimizer=optimizers.Adamax(),
              metrics=['acc'])
historyg = model.fit(
    train_generator,
    steps_per_epoch=120,
    epochs=100,
    validation_data=validation_generator,
    validation_steps=50,
    #callbacks=[callback],
    #verbose=0
)
Epoch 1/100
```

```
120/120 [==============================] - 301s 2s/step - loss: 2.1357 - acc: 0.3028 - val_loss: 1.4919 - val_acc: 0.4730
Epoch 2/100
120/120 [==============================] - 209s 2s/step - loss: 1.3067 - acc: 0.5165 - val_loss: 1.1981 - val_acc: 0.5540
Epoch 3/100
120/120 [==============================] - 206s 2s/step - loss: 1.0190 - acc: 0.6047 - val_loss: 1.0187 - val_acc: 0.6100
Epoch 4/100
120/120 [==============================] - 210s 2s/step - loss: 0.8948 - acc: 0.6512 - val_loss: 0.8031 - val_acc: 0.7440
Epoch 5/100
120/120 [==============================] - 208s 2s/step - loss: 0.7896 - acc: 0.6974 - val_loss: 0.7302 - val_acc: 0.7140
Epoch 6/100
120/120 [==============================] - 205s 2s/step - loss: 0.7096 - acc: 0.7269 - val_loss: 0.7801 - val_acc: 0.7220
Epoch 7/100
120/120 [==============================] - 210s 2s/step - loss: 0.6680 - acc: 0.7448 - val_loss: 0.6291 - val_acc: 0.7510
Epoch 8/100
120/120 [==============================] - 207s 2s/step - loss: 0.6557 - acc: 0.7485 - val_loss: 0.5726 - val_acc: 0.7850
Epoch 9/100
120/120 [==============================] - 207s 2s/step - loss: 0.5677 - acc: 0.7832 - val_loss: 0.5896 - val_acc: 0.7870
Epoch 10/100
120/120 [==============================] - 207s 2s/step - loss: 0.5764 - acc: 0.7774 - val_loss: 0.5219 - val_acc: 0.7930
Epoch 11/100
120/120 [==============================] - 206s 2s/step - loss: 0.5223 - acc: 0.8036 - val_loss: 0.5718 - val_acc: 0.7770
Epoch 12/100
120/120 [==============================] - 208s 2s/step - loss: 0.5084 - acc: 0.8066 - val_loss: 0.6392 - val_acc: 0.7630
Epoch 13/100
120/120 [==============================] - 208s 2s/step - loss: 0.4676 - acc: 0.8198 - val_loss: 0.4552 - val_acc: 0.8370
Epoch 14/100
120/120 [==============================] - 207s 2s/step - loss: 0.4747 - acc: 0.8191 - val_loss: 0.4541 - val_acc: 0.8350
Epoch 15/100
120/120 [==============================] - 204s 2s/step - loss: 0.4496 - acc: 0.8277 - val_loss: 0.3944 - val_acc: 0.8540
Epoch 16/100
120/120 [==============================] - 205s 2s/step - loss: 0.4220 - acc: 0.8382 - val_loss: 0.4371 - val_acc: 0.8410
Epoch 17/100
120/120 [==============================] - 207s 2s/step - loss: 0.4233 - acc: 0.8374 - val_loss: 0.4406 - val_acc: 0.8300
Epoch 18/100
```

```
120/120 [==============================] - 207s 2s/step - loss: 0.3694 - acc: 0.8588 - val_loss: 0.4507 - val_acc: 0.8360
Epoch 19/100
120/120 [==============================] - 205s 2s/step - loss: 0.3686 - acc: 0.8575 - val_loss: 0.3962 - val_acc: 0.8480
Epoch 20/100
120/120 [==============================] - 207s 2s/step - loss: 0.3604 - acc: 0.8645 - val_loss: 0.4181 - val_acc: 0.8290
Epoch 21/100
120/120 [==============================] - 208s 2s/step - loss: 0.3464 - acc: 0.8672 - val_loss: 0.3718 - val_acc: 0.8550
Epoch 22/100
120/120 [==============================] - 206s 2s/step - loss: 0.3558 - acc: 0.8642 - val_loss: 0.3741 - val_acc: 0.8600
Epoch 23/100
120/120 [==============================] - 205s 2s/step - loss: 0.3424 - acc: 0.8677 - val_loss: 0.3818 - val_acc: 0.8580
Epoch 24/100
120/120 [==============================] - 205s 2s/step - loss: 0.3504 - acc: 0.8689 - val_loss: 0.3072 - val_acc: 0.8930
Epoch 25/100
120/120 [==============================] - 201s 2s/step - loss: 0.3083 - acc: 0.8817 - val_loss: 0.3690 - val_acc: 0.8620
Epoch 26/100
120/120 [==============================] - 201s 2s/step - loss: 0.3065 - acc: 0.8810 - val_loss: 0.3182 - val_acc: 0.8790
Epoch 27/100
120/120 [==============================] - 200s 2s/step - loss: 0.2940 - acc: 0.8871 - val_loss: 0.3044 - val_acc: 0.8900
Epoch 28/100
120/120 [==============================] - 202s 2s/step - loss: 0.2801 - acc: 0.8938 - val_loss: 0.3882 - val_acc: 0.8590
Epoch 29/100
120/120 [==============================] - 203s 2s/step - loss: 0.2982 - acc: 0.8864 - val_loss: 0.3788 - val_acc: 0.8580
Epoch 30/100
120/120 [==============================] - 205s 2s/step - loss: 0.2413 - acc: 0.9094 - val_loss: 0.2701 - val_acc: 0.8980
Epoch 31/100
120/120 [==============================] - 208s 2s/step - loss: 0.2210 - acc: 0.9172 - val_loss: 0.3822 - val_acc: 0.8610
Epoch 32/100
120/120 [==============================] - 206s 2s/step - loss: 0.2381 - acc: 0.9079 - val_loss: 0.2621 - val_acc: 0.8860
Epoch 33/100
120/120 [==============================] - 203s 2s/step - loss: 0.2440 - acc: 0.9074 - val_loss: 0.3405 - val_acc: 0.8760
Epoch 34/100
120/120 [==============================] - 206s 2s/step - loss: 0.2324 - acc: 0.9104 - val_loss: 0.3156 - val_acc: 0.8690
Epoch 35/100
```

```
120/120 [==============================] - 202s 2s/step - loss: 0.2155 - acc: 0.9187 - val_loss: 0.2732 - val_acc: 0.8860
Epoch 36/100
120/120 [==============================] - 202s 2s/step - loss: 0.2524 - acc: 0.9044 - val_loss: 0.2449 - val_acc: 0.9100
Epoch 37/100
120/120 [==============================] - 204s 2s/step - loss: 0.2027 - acc: 0.9243 - val_loss: 0.3900 - val_acc: 0.8550
Epoch 38/100
120/120 [==============================] - 204s 2s/step - loss: 0.2152 - acc: 0.9188 - val_loss: 0.2676 - val_acc: 0.8820
Epoch 39/100
120/120 [==============================] - 203s 2s/step - loss: 0.1934 - acc: 0.9256 - val_loss: 0.2316 - val_acc: 0.9190
Epoch 40/100
120/120 [==============================] - 202s 2s/step - loss: 0.1822 - acc: 0.9328 - val_loss: 0.3368 - val_acc: 0.8710
Epoch 41/100
120/120 [==============================] - 201s 2s/step - loss: 0.1842 - acc: 0.9306 - val_loss: 0.2573 - val_acc: 0.9040
Epoch 42/100
120/120 [==============================] - 201s 2s/step - loss: 0.1652 - acc: 0.9380 - val_loss: 0.2381 - val_acc: 0.9130
Epoch 43/100
120/120 [==============================] - 204s 2s/step - loss: 0.1895 - acc: 0.9301 - val_loss: 0.2652 - val_acc: 0.9030
Epoch 44/100
120/120 [==============================] - 205s 2s/step - loss: 0.1695 - acc: 0.9357 - val_loss: 0.2470 - val_acc: 0.9100
Epoch 45/100
120/120 [==============================] - 211s 2s/step - loss: 0.1588 - acc: 0.9410 - val_loss: 0.2343 - val_acc: 0.9230
Epoch 46/100
120/120 [==============================] - 204s 2s/step - loss: 0.1650 - acc: 0.9388 - val_loss: 0.3389 - val_acc: 0.8770
Epoch 47/100
120/120 [==============================] - 207s 2s/step - loss: 0.1409 - acc: 0.9477 - val_loss: 0.2238 - val_acc: 0.9140
Epoch 48/100
120/120 [==============================] - 205s 2s/step - loss: 0.1426 - acc: 0.9461 - val_loss: 0.2735 - val_acc: 0.9030
Epoch 49/100
120/120 [==============================] - 205s 2s/step - loss: 0.1393 - acc: 0.9477 - val_loss: 0.3111 - val_acc: 0.8960
Epoch 50/100
120/120 [==============================] - 209s 2s/step - loss: 0.1332 - acc: 0.9508 - val_loss: 0.2745 - val_acc: 0.8990
Epoch 51/100
120/120 [==============================] - 205s 2s/step - loss: 0.1336 - acc: 0.9505 - val_loss: 0.2664 - val_acc: 0.8940
Epoch 52/100
```

```
120/120 [==============================] - 204s 2s/step - loss: 0.1380 - acc: 0.9487 - val_loss: 0.2009 - val_acc: 0.9320
Epoch 53/100
120/120 [==============================] - 202s 2s/step - loss: 0.1036 - acc: 0.9622 - val_loss: 0.2409 - val_acc: 0.9110
Epoch 54/100
120/120 [==============================] - 206s 2s/step - loss: 0.1082 - acc: 0.9593 - val_loss: 0.2188 - val_acc: 0.9150
Epoch 55/100
120/120 [==============================] - 202s 2s/step - loss: 0.1182 - acc: 0.9562 - val_loss: 0.2967 - val_acc: 0.8870
Epoch 56/100
120/120 [==============================] - 201s 2s/step - loss: 0.1111 - acc: 0.9598 - val_loss: 0.2335 - val_acc: 0.9270
Epoch 57/100
120/120 [==============================] - 204s 2s/step - loss: 0.1153 - acc: 0.9569 - val_loss: 0.2376 - val_acc: 0.9110
Epoch 58/100
120/120 [==============================] - 204s 2s/step - loss: 0.1174 - acc: 0.9567 - val_loss: 0.2981 - val_acc: 0.9050
Epoch 59/100
120/120 [==============================] - 207s 2s/step - loss: 0.0940 - acc: 0.9654 - val_loss: 0.2193 - val_acc: 0.9250
Epoch 60/100
120/120 [==============================] - 207s 2s/step - loss: 0.0857 - acc: 0.9691 - val_loss: 0.1967 - val_acc: 0.9320
Epoch 61/100
120/120 [==============================] - 207s 2s/step - loss: 0.1033 - acc: 0.9625 - val_loss: 0.2649 - val_acc: 0.9200
Epoch 62/100
120/120 [==============================] - 207s 2s/step - loss: 0.0877 - acc: 0.9670 - val_loss: 0.1856 - val_acc: 0.9370
Epoch 63/100
120/120 [==============================] - 208s 2s/step - loss: 0.0902 - acc: 0.9679 - val_loss: 0.2310 - val_acc: 0.9300
Epoch 64/100
120/120 [==============================] - 213s 2s/step - loss: 0.0842 - acc: 0.9690 - val_loss: 0.2229 - val_acc: 0.9240
Epoch 65/100
120/120 [==============================] - 218s 2s/step - loss: 0.0708 - acc: 0.9749 - val_loss: 0.1697 - val_acc: 0.9420
Epoch 66/100
120/120 [==============================] - 232s 2s/step - loss: 0.0737 - acc: 0.9731 - val_loss: 0.2386 - val_acc: 0.9170
Epoch 67/100
120/120 [==============================] - 219s 2s/step - loss: 0.0844 - acc: 0.9697 - val_loss: 0.4065 - val_acc: 0.8720
Epoch 68/100
120/120 [==============================] - 208s 2s/step - loss: 0.1077 - acc: 0.9608 - val_loss: 0.2355 - val_acc: 0.9230
Epoch 69/100
```

```
120/120 [==============================] - 204s 2s/step - loss: 0.0638 - acc: 0.9774 - val_loss: 0.1948 - val_acc: 0.9290
Epoch 70/100
120/120 [==============================] - 207s 2s/step - loss: 0.0544 - acc: 0.9804 - val_loss: 0.1954 - val_acc: 0.9380
Epoch 71/100
120/120 [==============================] - 213s 2s/step - loss: 0.0697 - acc: 0.9749 - val_loss: 0.1836 - val_acc: 0.9440
Epoch 72/100
120/120 [==============================] - 208s 2s/step - loss: 0.0511 - acc: 0.9823 - val_loss: 0.1792 - val_acc: 0.9400
Epoch 73/100
120/120 [==============================] - 209s 2s/step - loss: 0.0520 - acc: 0.9812 - val_loss: 0.1957 - val_acc: 0.9320
Epoch 74/100
120/120 [==============================] - 207s 2s/step - loss: 0.0566 - acc: 0.9789 - val_loss: 0.2505 - val_acc: 0.9170
Epoch 75/100
120/120 [==============================] - 205s 2s/step - loss: 0.0930 - acc: 0.9666 - val_loss: 0.2920 - val_acc: 0.9070
Epoch 76/100
120/120 [==============================] - 206s 2s/step - loss: 0.0660 - acc: 0.9758 - val_loss: 0.2503 - val_acc: 0.9210
Epoch 77/100
120/120 [==============================] - 208s 2s/step - loss: 0.0643 - acc: 0.9780 - val_loss: 0.2140 - val_acc: 0.9320
Epoch 78/100
120/120 [==============================] - 210s 2s/step - loss: 0.0422 - acc: 0.9853 - val_loss: 0.2413 - val_acc: 0.9280
Epoch 79/100
120/120 [==============================] - 210s 2s/step - loss: 0.0425 - acc: 0.9849 - val_loss: 0.2522 - val_acc: 0.9270
Epoch 80/100
120/120 [==============================] - 206s 2s/step - loss: 0.0631 - acc: 0.9777 - val_loss: 0.1756 - val_acc: 0.9480
Epoch 81/100
120/120 [==============================] - 207s 2s/step - loss: 0.0405 - acc: 0.9852 - val_loss: 0.1664 - val_acc: 0.9440
Epoch 82/100
120/120 [==============================] - 216s 2s/step - loss: 0.0411 - acc: 0.9845 - val_loss: 0.2074 - val_acc: 0.9320
Epoch 83/100
120/120 [==============================] - 218s 2s/step - loss: 0.0434 - acc: 0.9835 - val_loss: 0.2431 - val_acc: 0.9400
Epoch 84/100
120/120 [==============================] - 213s 2s/step - loss: 0.0802 - acc: 0.9717 - val_loss: 0.2462 - val_acc: 0.9330
Epoch 85/100
120/120 [==============================] - 209s 2s/step - loss: 0.0402 - acc: 0.9852 - val_loss: 0.1988 - val_acc: 0.9330
Epoch 86/100
```

```
120/120 [==============================] - 214s 2s/step - loss: 0.0371 - acc: 0.9863 - val_loss: 0.2561 - val_acc: 0.9380
Epoch 87/100
120/120 [==============================] - 215s 2s/step - loss: 0.0353 - acc: 0.9871 - val_loss: 0.2422 - val_acc: 0.9340
Epoch 88/100
120/120 [==============================] - 207s 2s/step - loss: 0.0388 - acc: 0.9860 - val_loss: 0.1882 - val_acc: 0.9400
Epoch 89/100
120/120 [==============================] - 206s 2s/step - loss: 0.0659 - acc: 0.9761 - val_loss: 0.2298 - val_acc: 0.9240
Epoch 90/100
120/120 [==============================] - 215s 2s/step - loss: 0.0393 - acc: 0.9852 - val_loss: 0.2113 - val_acc: 0.9360
Epoch 91/100
120/120 [==============================] - 233s 2s/step - loss: 0.0327 - acc: 0.9881 - val_loss: 0.2295 - val_acc: 0.9320
Epoch 92/100
120/120 [==============================] - 212s 2s/step - loss: 0.0418 - acc: 0.9858 - val_loss: 0.4723 - val_acc: 0.8820
Epoch 93/100
120/120 [==============================] - 210s 2s/step - loss: 0.0747 - acc: 0.9733 - val_loss: 0.2514 - val_acc: 0.9420
Epoch 94/100
120/120 [==============================] - 210s 2s/step - loss: 0.0305 - acc: 0.9892 - val_loss: 0.1989 - val_acc: 0.9370
Epoch 95/100
120/120 [==============================] - 213s 2s/step - loss: 0.0281 - acc: 0.9888 - val_loss: 0.1802 - val_acc: 0.9480
Epoch 96/100
120/120 [==============================] - 209s 2s/step - loss: 0.0294 - acc: 0.9895 - val_loss: 0.2724 - val_acc: 0.9210
Epoch 97/100
120/120 [==============================] - 217s 2s/step - loss: 0.0295 - acc: 0.9891 - val_loss: 0.2283 - val_acc: 0.9250
Epoch 98/100
120/120 [==============================] - 216s 2s/step - loss: 0.0280 - acc: 0.9898 - val_loss: 0.2175 - val_acc: 0.9390
Epoch 99/100
120/120 [==============================] - 210s 2s/step - loss: 0.0555 - acc: 0.9797 - val_loss: 0.2581 - val_acc: 0.9290
Epoch 100/100
120/120 [==============================] - 209s 2s/step - loss: 0.0452 - acc: 0.9836 - val_loss: 0.2729 - val_acc: 0.9160
```

```python
model.save(model_g30Adamax.h5')
```

```python
# Plot the Accuracy and Loss curves

import matplotlib.pyplot as plt

acc = historyg.history['acc']
val_acc = historyg.history['val_acc']
loss = historyg.history['loss']
val_loss = historyg.history['val_loss']

epochs = range(len(acc))

plt.plot(epochs, acc, 'ko-', label='Training acc')
plt.plot(epochs, val_acc, color='grey',marker='s',linestyle='None' ,linewidth=5, label='Validation acc')
plt.title('Training and validation accuracy')
plt.legend()

plt.figure()

plt.plot(epochs, loss, 'ko-', label='Training loss')
plt.plot(epochs, val_loss, color='grey',marker='s' ,linestyle='None', linewidth=5, label='Validation loss')
plt.title('Training and validation loss')
plt.legend()

plt.show()
```

## Model Evaluation and Test

```python
import numpy as np
import time
from keras.models import load_model
from tensorflow.keras.preprocessing.image import ImageDataGenerator

# Load the saved model
model = load_model('/kaggle/input/lsp-g30m-adamax/model_g30Adamax.h5')

# Define the generator for loading test images
test_datagen = ImageDataGenerator(rescale=1./255)
test_generator = test_datagen.flow_from_directory(
        '/kaggle/input/lsp-grn/g/test ',
        target_size=(256, 256),
        color_mode='grayscale',
        batch_size=32,
        class_mode='categorical',
        shuffle=False)
```

```python
# Get the number of test samples
num_samples = len(test_generator.filenames)

# Make predictions on the test images and measure the inference time
start_time = time.time()
preds = model.predict_generator(test_generator, steps=num_samples//32+1)
end_time = time.time()

# Calculate the total inference time and the average inference time per sample
total_time= end_time-start_time
avg_time_per_sample =total_time/num_samples

print("Total inference time:", total_time)
print("Average inference time per sample:", avg_time_per_sample)
```

# Appendix C: Thesis-Related Publications

## الملخص

تستخدم ماكينات القطع بالليزر في مختلف الصناعات، ولكن استهلاك الطاقة والتأثير البيئي لها يشكلان تحديات كبيرة. تقترح هذه الدراسة نهجًا يعتمد على التعلم العميق لتقليل استهلاك الطاقة وخفض معدلات التلوث الناتجه عن آلات قطع بالليزر. تم جمع البيانات التجريبية من آلة قطع بالليزر وتحليلها لتحديد فرص توفير الطاقة. بناء علي النتايج الخاصة بالبيانات تم تطوير نموذج مخصص للتعلم العميق يساعد علي تحديد نوع المادة التي تتم قطعها، مما يسهل التحكم في الآلة بناءً على عملية القطع وخصائص المواد والقوة الكهربية التي تحتاجها الماكينة لتقطيع المادة المكتشفه.مما ساعد علي التحكم في مضخة شفط العادم وضاغط الهواء تلقائيًا عند عدم وجود عمليات قيد التنفيذ او مواد لا تحتاج لتشغيلهم، وحيث ان عملية التحكم في فتح وغلق الادوات الملحقة بالماكينة يساعد على عمل خفض لمستويات الطاقة المستهلكه من الماكينة بشكل عام فقد نتج عن هذا النهج خفض في الطاقة الكلية المستهلكه من الماكينة. وللوصول لأعلى قدر من خفض الطاقة، تم وضع نهج آخر يعتمد على التعلم العميق لاكتشاف مستوى الدخان الذي يتم توليده أثناء القطع. تم استخدام هذه المعلومات لضبط قوة المضخة وفقًا لذلك، مما أدى إلى تقليل مزيد من الطاقة. تظهر نتائج هذه الدراسة تقليلًا ملحوظًا بنسبة تصل إلي ٥٠٪ في استهلاك الطاقة من قبل مضخة شفط الهواء طبقا لظروف التشغيل، مما يؤدي إلى تقليل كبير في استهلاك الطاقة العام لآلة قطع الليزر.

محتويات الرسالة:

الفصل الأول: يقدم نظرة عامة على موضوع البحث، مسلطًا الضوء على أهمية قطع الليزر في مختلف الصناعات. ويتم مناقشة دوافع الدراسة، مع التركيز على ضرورة التعامل مع استهلاك الطاقة والتأثير البيئي، وتسليط الضوء على بيان المشكلة، وأهداف البحث، ثم يختتم الفصل بعرض لهيكل الرسالة.

الفصل الثاني: يركز على إجراء مراجعة شاملة للمصادر والمراجع الأدبية. يستكشف الجسم المعرفي والأبحاث المتعلقة بقطع الليزر، وتحسين الطاقة، والتأثير البيئي، وتقنيات التعلم العميق بالإضافة إلى استشعار التشويش في عمليات التصنيع. وايضا يقوم الفصل بتجميع وتحليل الدراسات ذات الصلة، وتحديد الثغرات ووضع الأسس النظرية للبحث.

الفصل الثالث: يتعمق في تطبيق تقنيات التعلم العميق لتصنيف المواد في قطع الليزر باستخدام استشعار التشويش. يتم فحص مختلف الخوارزميات والنماذج المستخدمة لتصنيف المواد، ويتم مناقشة نقاط قوتها وضعفها. يقدم الفصل منهجية وإعداد التجارب لتدريب وتقييم نموذج التعلم العميق المستخدم في الدراسة. ويتم أيضا تقديم وتحليل نتائج النموذج المقترح لتصنيف المواد.

الفصل الرابع: يركز على اكتشاف نوع المواد أثناء عمليات قطع الليزر باستخدام تقنيات التعلم العميق. مع تطوير نموذج التعلم العميق المصمم خصيصًا لاكتشاف نوع المادة المراد تقطيعها ودمجه في نظام قطع الليزر. يصف الفصل منهجية تدريب واختبار النموذج ويقدم نتائج دقة الكشف. يتم مناقشة النتائج بتفصيل. ومقارناتها مع ما تم بالفصل السابق.

الفصل الخامس: يعرض استراتيجيات تحسين الطاقة في قطع الليزر. ويستقصي وسائل لتقليل استهلاك الطاقة عن طريق التحكم في مكونات مثل مضخات شفط الهواء والضواغط. ثم يناقش الفصل استخدام نماذج التعلم العميق لضبط عمليات الآلة تلقائيًا بناءً على ظروف التشغيل، مما يؤدي إلى توفير الطاقة بنسبة قد تصل الي اكثر من ٥٠٪.

الفصل السادس: يناقش الاستنتاجات وملخص لمخرجات الرسالة وأيضا يفتح الباب لأعمال مستقبلية يمكن بنائها علي النهج المقترح.

وفي نهاية الرسالة تم عرض مجموعة من المراجع والملاحق المستخدمة في الرسالة.

# أساليب الذكاء الاصطناعى لتوفير الطاقة فى آلات القطع بالليزر

إعداد

**محمد عبدالله محمد حامد سالم**

رسالة مقدمة للأكاديمية العربية للعلوم والتكنولوجيا والنقل البحرى لاستكمال متطلبات نيل درجة

**ماجستير العلوم**
فى
الهندسة الانظمة الذكية لإدارة الطاقة

إشـــــــراف

| أ.د. حمدي أحمد عاشور | د. أحمد الشناوي |
|---|---|
| أستاذ بقسم الهندسة الكهربية والتحكم | أستاذ مساعد بقسم الهندسة الكهربية والتحكم |
| الأكاديمية العربية للعلوم | الأكاديمية العربية للعلوم |
| والتكنولوجيا و النقل البحري | والتكنولوجيا و النقل البحري |
| الأسكندرية | الأسكندرية |

**2023**